\UseRawInputEncoding
\documentclass[review]{elsarticle}
\usepackage{color}
\usepackage{ulem}

\usepackage{lineno,hyperref}
\modulolinenumbers[5]

\journal{Journal of \LaTeX\ Templates}

\usepackage{algorithm}
\usepackage{algorithmic}
\usepackage{multirow}
\usepackage{xcolor}
\usepackage{amsmath}
\usepackage{graphicx}
\usepackage{float}
\usepackage{longtable,booktabs}
\usepackage{array}
\usepackage{subfigure}
\usepackage{amsthm}
\usepackage{bigstrut,bigdelim,multirow}
\newcommand{\PreserveBackslash}[1]{\let\temp=\\#1\let\\=\temp}
\newcolumntype{C}[1]{>{\PreserveBackslash\centering}p{#1}}
\newcolumntype{R}[1]{>{\PreserveBackslash\raggedleft}p{#1}}
\newcolumntype{L}[1]{>{\PreserveBackslash\raggedright}p{#1}}

\begin{document}

\begin{frontmatter}

\title{A Framework Based on Generational and Environmental Response Strategies for Dynamic Multi-objective Optimization}
\tnotetext[mytitlenote]{Fully documented templates are available in the elsarticle package on \href{http://www.ctan.org/tex-archive/macros/latex/contrib/elsarticle}{CTAN}.}


\author[myfirstaddress]{Qingya Li}
\ead{mrli1991@foxmail.com}
\author[mymainaddress]{Xiangzhi Liu}
\author[mymainaddress]{Fuqiang Wang}
\author[mysecondaryaddress]{Shuai Wang }
\author[mymainaddress]{Peng Zhang}
\author[mymainaddress]{Xiaoming Wu \corref{mycorrespondingauthor}}
\cortext[mycorrespondingauthor]{Corresponding author}
\ead{wuxm@sdas.org}


\address[myfirstaddress]{Guangdong Provincial Key Laboratory of Brain-inspired Intelligent Computation, Department of Computer Science and Engineering,	Southern University of Science and Technology, Shenzhen 518055, China.}
\address[mymainaddress]{Qilu University of Technology (Shandong Academy of Sciences),
	Shandong Computer Science Center (National Supercomputer Center in Jinan), Shandong Provincial Key Laboratory of Computer Networks, Shandong, China.}
\address[mysecondaryaddress]{Heze Branch, Qilu University of Technology(Shandong Academy of Sciences), Biological Engineering Technology Innovation Center of Shandong Province, Shandong, China.}





\begin{abstract}
Due to the dynamics and uncertainty of the \textit{dynamic multi-objective optimization problems} (\textit{DMOPs}), it is  difficult for algorithms to find a satisfactory solution set before the next environmental change, especially for some complex environments. One reason may be that the information in the environmental static stage can not be used well in the traditional  framework. In this paper, \textit{a novel framework based on generational and environmental response strategies} (\textit{FGERS}) is proposed, in which response strategies are run both in the environmental change stage and the environmental static stage to obtain population evolution information of those both stages. Unlike in the traditional framework, response strategies are only run in the environmental change stage. For simplicity, the feed-forward center point strategy was chosen to be the response strategy in the novel dynamic  framework (FGERS-CPS). FGERS-CPS is not only to predict change trend of the optimum solution set in the environmental change stage,  but to predict the evolution trend of the population after several generations in the environmental static stage. Together with the feed-forward center point strategy, a simple memory strategy and adaptive diversity maintenance strategy were used to form the complete FGERS-CPS.  On 13 DMOPs with various characteristics, FGERS-CPS was compared with four classical response strategies in the traditional framework. Experimental results show that FGERS-CPS is effective for DMOPs. 
\end{abstract}


\end{frontmatter}

\linenumbers

\section{Introduction}
\label{intro}
In the real world, there is a kind of problems with multiple objectives, and these objectives always change over time. This kind of problems is called \textit{dynamic multi-objective optimization problems} (\textit{DMOPs}) \cite{sgea}. Since evolutionary algorithms can solve such problems well, they have been widely used in various DMOPs, such as scheduling \cite{cao3} \cite{schedule} \cite{dnsga2},  mission planning \cite{ckps4}, machine learning \cite{clustering} \cite{bsrabil}, wireless network design \cite{ckps9}, and greenhouse control \cite{cao6}. The mathematical form of DMOPs is:
\begin{displaymath}
\left\{\begin{array}{l}
\min_{x\in\Omega} F(x,t)=(f_1(x,t),f_2(x,t),\ldots,f_m(x,t))^T\\
{\rm s.t.} \quad g_i(x,t)\leq 0,i=1,2,\ldots,p,\\
\qquad h_j(x,t)=0,j=1,2,\ldots,q,
\end{array}
\right.
\end{displaymath}
where
$\textbf{x}=(x_1,x_2,\ldots,x_n)$
is the $n$-dimensional decision vector and its domain of definition is $\Omega$. $t$ denotes the time change variable.
$\textbf{F}=(f_1,f_2,\ldots,f_m)$
is the $m$-dimensional objective vector. $g$ represents $p$ inequality constraints, and $h$ denotes the $q$ equality constraints. 

Nowadays, DMOPs are divided into DMOPs with deterministic environmental changes \cite{fda} \cite{dmop} \cite{pps} and DMOPs with less detectable environmental changes \cite{jiang2018less} \cite{sahmoud2019exploiting} \cite{sahmoud2019hybrid}. In this paper, DMOPs used are DMOPs with deterministic environmental changes. In the DMOPs with deterministic environmental changes, the relationship of the environmental change and the number of generations is as follows.
\begin{equation}\label{equ:taot}
t=\frac{1}{n_{T}} \left \lfloor \frac{\tau }{\tau_{T}}  \right \rfloor,
\end{equation}
where \textit{t} and \textit{$\tau$} denote the number of environmental changes and the number of generations, respectively. $n_{T}$ and $\tau_{T}$ show the severity and frequency of the environment change.

In this paper, unless otherwise specified, optimization refers to minimized optimization. In addition, 
in evolutionary algorithms, if it is necessary to compare which of two individuals is better, the definition of Pareto Dominance needs to be introduced. 
\theoremstyle{definition}
\newtheorem {def1}{Definition}
\begin{def1}[Pareto Dominance]
	At the $t$th time step, $p$ and $q$ are any two individuals in the population; $f$ is the objective function; $p$ is said to dominate $q$, denoted by $f\left(p, t\right)\prec f\left(q, t\right)\; iff\; f_{i}\left(p\right)\leq f_{i}\left(q\right)$ $\forall i=\left\{ 1, 2, \ldots, m\right\}$ and $f_{j}\left(p\right)<f_{j}\left(q\right)\exists j\in\left\{ 1, 2, \ldots, m\right\}$.
\end{def1}

The optimal tradeoff solution set represents the best solution set, which is called \textit{Pareto optimal set} (\textit{PS}) in the decision space and \textit{Pareto optimal front} (\textit{PF}) in the objective space. The definitions of PS and PF are as follows.

\newtheorem {def2}[def1]{Definition}
\begin{def2}[Pareto Optimal Set (PS)]
	$x$ is a decision variable; $\Omega$ is the decision space; $F$ is the objective function; therefore, $PS (t) $ is the set composed by all nondominated solutions in the decision space at the $t$th time step and defined mathematically as:
	\begin{displaymath}
	PS(t)=\left\{ x\in\Omega\mid\not\exists x^{\star}\in\Omega,F\left(x^{\star}\right)\prec F\left(x\right)\right\}.
	\end{displaymath}
\end{def2}

\newtheorem {def3}[def1]{Definition}
\begin{def3}[Pareto Optimal Front (PF)]
	$x$ is a decision variable; $F$ is the objective function; thus, $PF(t)$ is the set composed by all the nondominated solutions with respect to the objective space at the $t$th time step and defined mathematically as:
	\begin{displaymath}
	PF(t)=\left\{ y=F\left(x\right)\mid x\in PS(t)\right\}.
	\end{displaymath}
\end{def3}

Due to the dynamics and uncertainty of DMOPs, the traditional methods of solving static multi-objective optimization problems \cite{nsga2} \cite{ckps3} have not been suited. This is because the traditional static algorithms can only bring the population closer to the optimal solutions step by step, and may eventually converge to a local area. However, when the environment changes, the optimal solutions in the new environment may be far away from the optimal solutions in the original environment. At this time, it is often difficult to converge again in a short time. Therefore, static algorithms need to be improved to meet the needs of dynamics  \cite{dnsga2}.
  
To make static algorithms to solve DMOPs, researchers proposed a kind of dynamic framework, which introduces response strategies into static algorithms to  deal specially with environmental dynamics. In this framework, there are two stages: the environmental change stage and the environmental static stage. In the environmental change stage, response strategies are used to handle dynamic changes. And in the  environmental static stage, an optimization algorithm is used to get optimal solutions for problems. Currently, all strategies used to solve DMOPs are in this framework. The common, classic, and state of the art response strategies are as follows.

One of the simplest ways to deal with DMOPs is to increase the diversity of the population \cite{ruan} \cite{pengmeeting}. The principle of the diversity strategy is to introduce some diverse individuals to the population so that the individuals are redistributed in the entire decision space. In this way, when the environment changes, individuals in the vicinity of the optimal solutions in the new environment can quickly converge through evolution. Common diversity strategies include hyper mutation methods \cite{dnsga2}, random immigrants, and other immigrants' methods \cite{ckps31} \cite{ckps32}, the adaptive diversity introduction strategy \cite{liumin}, employing multiple populations and parallel computing \cite{ckps33} \cite{ckps34}. However, the use of the diversity strategy has certain blindness. Because after the introduction of diverse individuals, the individuals still need independent evolution, and there is no guidance for the convergence of the population \cite{ckps}.

Another commonly used method is memory strategies \cite{cao12} \cite{ckps51} \cite{ybravo} \cite{evinek}. The general principle of this method is to memorize the optimal solutions of some previous environments, and then those memorized individuals are added to the population in the new environment. In this way, if the new environment is similar to the original environment where the memory individuals lie, these individuals can quickly guide the population to converge towards the optimal solutions. The memory strategy works well for problems with periodic changes, but it is not ideal for problems that are not periodic.

Another common method to deal with DMOPs is the multi-population strategy \cite{dmop} \cite{liu2014novel} \cite{jin2016reference} \cite{liu2017coevolutionary} \cite{xu2017environment} \cite{liu2020cooperative}. The multi-population method generally divides the population into multiple sub-populations, and then each sub-population handles different sub-problems separately through competition or cooperation. In 2009, Goh et al.  \cite{dmop} proposed a competition-cooperation mechanism to solve DMOPs. Each sub-population represents each subcomponent of DMOPs by competition. And then the winners will cooperate with each other to get better solutions. In 2014, Liu et al. \cite{liu2014novel} proposed a cooperative coevolutionary optimization algorithm using a modified linear regression model to solve DMOPs. The optimization algorithm is based on non-dominated sorting, and the problem is decomposed on the basis of the search process of the decision space. Then, each species subcomponent cooperates to produce better solutions. In 2016, Yang  et al. \cite{jin2016reference} proposed a strategy to divide the population into multiple subpopulations based on reference points. The center point in the new environment is predicted based on the center points of multiple subpopulations belonging to the same reference point in the previous environment. Then, uniform distribution and Gaussian distribution are used to generate the initial population to improve diversity and convergence performance. In 2017, Liu et al. \cite{liu2017coevolutionary} proposed a multi-swarm particle swarm optimization algorithm for DMOPs based on co-evolution technology. The number of swarms is decided according to the number of objective functions, and then the information sharing strategy is used for cooperative evolution. In 2017, Xu et al. \cite{xu2017environment} proposed a cooperative co-evolutionary algorithm based on environmental sensitivity to solve DMOPs. The population is divided into two sub-populations to process variables that are divided into two subcomponents. In addition, differential prediction and Cauchy mutation are used to accelerate the response to environmental changes. In 2020, Liu et al. \cite{liu2020cooperative} proposed a cooperative particle swarm optimization algorithm to solve DMOPs. By a learning strategy, multiple swarms cooperate with each other to get optimum solutions. Moreover, when environment changes, according to the belonging PF subparts, the particles in the previous environment will be relocated by a prediction strategy based on reference points.

The most commonly used strategy is the prediction methods \cite{fps} \cite{pps} \cite{ckps} \cite{spps} \cite{pengpredict} \cite{pmga}. The principle of the prediction method is generally to predict some information after the environmental change by learning the information of the previous environments, and the predicted information can help the population quickly converge towards optimum solutions. In 2006, Hatzakis et al. \cite{fps} proposed a feed-forward prediction strategy. This strategy is to learn the change rule of extreme points  in the historical environments through the prediction model to predict the extreme points of the next environment, and then these points are used to guide the population to converge towards the optimal solutions in the new environment. In 2014, Zhou  et al. \cite{pps} proposed a population prediction strategy. The population is divided into two parts. One part is to use the AR model to learn the center point change rule in the historical environments, thereby predicting the center point in the new environment. And then the predicted center point is used with the center point of the current population to predict the PS shape in the new environment. In 2016, Muruganantham et al. \cite{cao21} proposed a method of using a Kalman filter to predict individuals. This method uses a Kalman filter with MOEA/D to learn the rule of optimal individuals changes in the historical environments, so as to predict the optimal individuals in the next environment. In 2017, Jiang et al. \cite{jiang2017transfer} proposed the  transfer learning-based dynamic multi-objective algorithms, which can use the transfer learning technique to obtain an effective initial population. In 2019, Ruan et al. \cite{ruan2019and}  did a detailed research about when and how to use tranfer knowledge to dynamic multi-objective optimization to obtain the better effect. In 2019, Rong  et al. \cite{rong2019} proposed a multi-model prediction method, which explores different types of changes in DMOPs through the center point, and then different prediction methods are adopted according to different types. In 2019, Liang  et al. \cite{liang2019} proposed a hybrid memory and prediction method that detects whether the new environment is similar to some historical environment. If it is similar, it uses the center point in the current environment and the center point memorized to relocate PS in the new environment.  Otherwise, a differential prediction based on the previous two consecutive environmental center points is used to predict the optimal solutions of the new environment. In 2020, Jiang et al. \cite{jiang2020knee} used estimated knee points by transfer learning to obtain the initial population. In fact, the center point method is widely used in prediction strategies \cite{ckps} \cite{spps} \cite{pps} \cite{zhu2020} \cite{dops} \cite{dss} \cite{rong2018} \cite{dee}. \\
\indent From the literature above, we can see that all the strategies are implemented during the environmental change stage at present. However, when solving DMOPs, there are two stages: the environmental change stage and the  environmental static stage.  In the environmental static stage, no measures are taken in addition to natural evolution. If some effective method is taken to exploit the useful information in this stage, the algorithm will solve DMOPs faster and better.
This paper proposes a novel dynamic convergence-accelerated framework. In the novel framework, response strategies can be designed both in the  environmental change stage and  environmental static stage. And a use example in the novel framework (FGERS-CPS) is proposed. FGERS-CPS not only uses the feed-forward center point method for prediction during the environmental change stage; but also uses this method to predict the convergence trend of the population in the environmental static stage. And then the predicted population combine the current population to make the environmental selection, and the selected individuals carry out the environmental evolution of the next generation. From the experimental results, it is known that FGERS-CPS has a strong convergence ability.

The rest of this article is structured as follows. Section \ref{sec:2} introduces the traditional dynamic framework and  the novel dynamic framework. Section \ref{sec:3} describes FGERS-CPS in detail. Section \ref{sec:4.1} introduces test instances and performance indicators. Section \ref{sec:4} gives
experimental results and analysis. Section \ref{sec:5} does more discussion to further
analyze the advantages and disadvantages of the traditional dynamic framework and the novel dynamic framework. And in the end, Section \ref{sec:6} gives conclusions and future work.
\section{Traditional dynamic framework and  novel dynamic framework}
\label{sec:2}

\subsection{Traditional dynamic convergence-accelerated framework}
\label{sec:2.1}
\begin{figure}
	\centering	
	\begin{minipage}[t]{0.5\linewidth}
		
		\centering
		
		\includegraphics[width=2.3in]{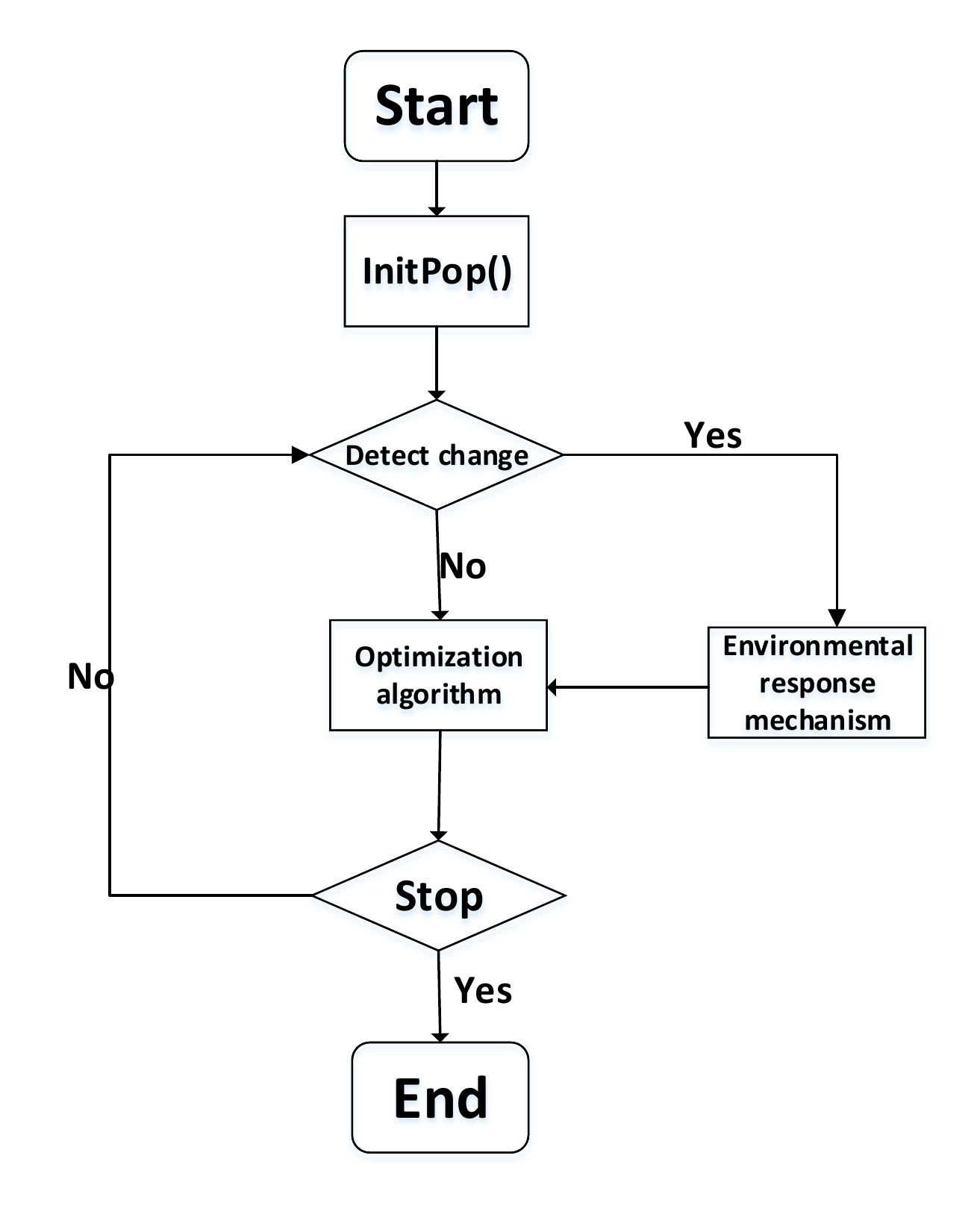}
		
		\caption{Traditional  framework.} 
		\label{fig1}
		
	\end{minipage}%
	\hfill
	\begin{minipage}[t]{0.5\linewidth}
		
		\centering
		
		\includegraphics[width=2.2in]{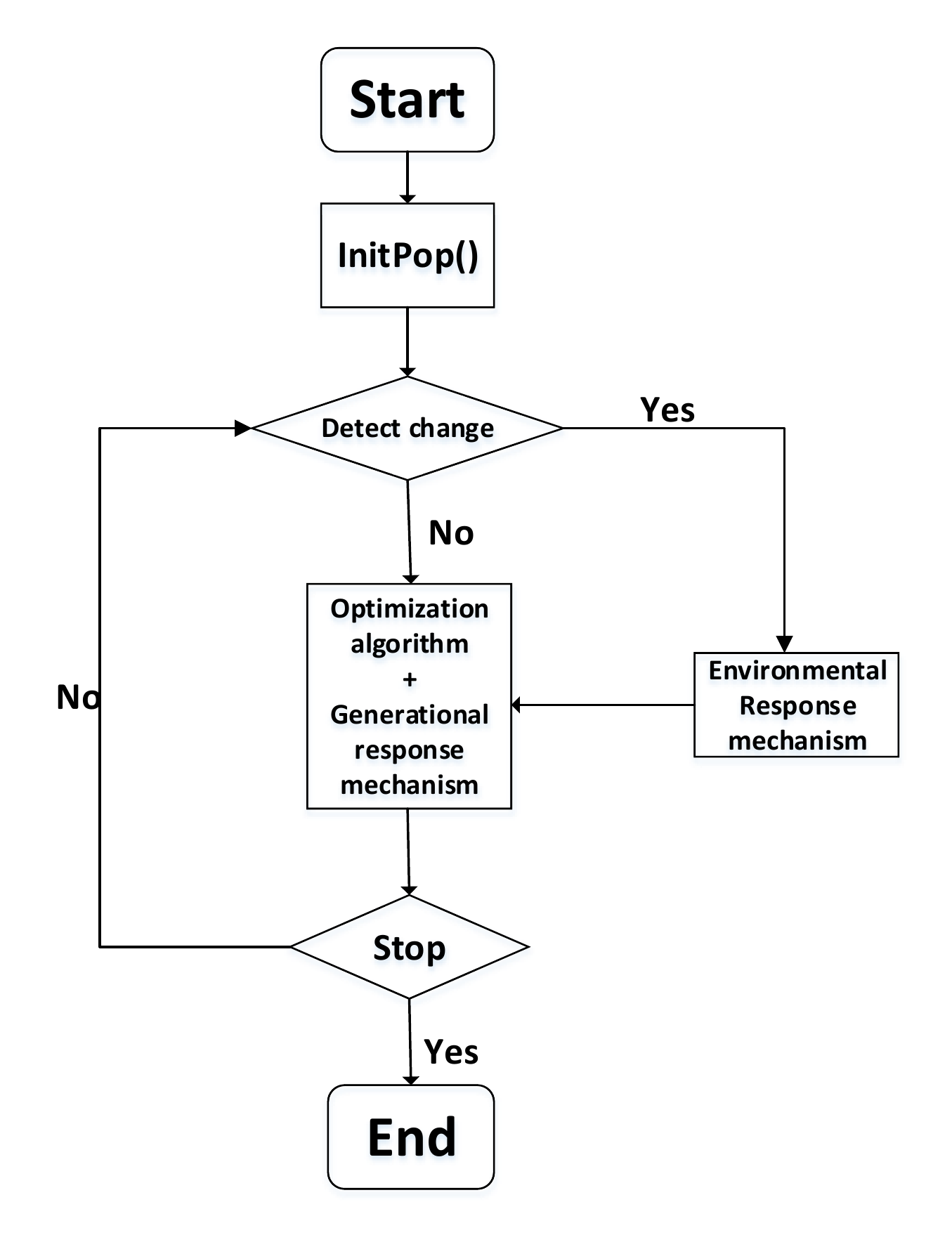}
		
		\caption{Novel framework (adding generational response mechanism).}
		\label{fig2}
		
	\end{minipage}
\end{figure}

Figure \ref{fig1} shows this traditional convergence-accelerated version of the dynamic multi-objective framework, including the following steps:
\begin{enumerate}
	\item Initialize the population.
	\item Detect environmental changes. If no change is detected, go to $Step$ $4$.
	\item Respond the environmental change by the environmental response mechanism.
	\item An optimization algorithm is used to solve problems.
	\item Determine the termination conditions. If not, go to $Step$ $2$; otherwise, it ends and exits.
\end{enumerate}


\subsection{Novel dynamic convergence-accelerated framework}
\label{sec:2.2}


%
%
%
%
%
%
%
%
%

Figure \ref{fig2} shows the novel version of the dynamic multi-objective framework. In addition to the environmental response mechanism, a generational response mechanism is added into the novel framework. Because in some complex dynamic environments (such as the environment with a low change frequency or large change severity), the difficulty of convergence will be greater. If only using environmental response mechanism in such a complex environment, the solutions obtained are not always satisfactory. The generational response mechanism can accelerate the population's convergence by judging the direction of the population's convergence after several generations in the static environment, so as to get the better solutions before the next environmental change.
\section{Proposed FGERS-CPS}
\label{sec:3}
This paper proposes a feed-forward center point prediction strategy under the novel framework. FGERS-CPS includes two aspects: the environmental response strategy and the generational response strategy. When an environmental change is detected, the environmental response strategy is used to respond the environmental change. While in the environmental static stage, the generational response strategy is used to accelerate the convergence of the population before the next environmental change.
\subsection{Environmental response strategy}
\label{sec:3.1}
The environmental response strategy includes three aspects: the prediction strategy based on feed-forward center point; the simple memory strategy and the adaptive diversity maintenance strategy. The purpose of the environmental response strategy is to get the  initial population to accelerate the convergence of the population towards optimal solutions of the environment at the next time step. Here, the initial population is called environmental response population, denoted as $Pop^{t+1}$, which means the population in ($t+1$)th time step. And the population capacity is expressed as $Npop$.
\subsubsection{Prediction strategy based on the feed-forward center point}
\label{sec:3.1.1}
At the environmental change stage, the feed-forward center point strategy \cite{spps}, \cite{ckps}, \cite{rong2019} is widely used.  The feed-forward center point strategy  predicts the direction and distance of environmental evolution through the center points of the non-dominated sets in the final populations in the previous two environments, so as to predict the population which is suitable for survival in the next environment.

The center point of the non-dominated set is used here. The center point in $t$th time step, $C^t$, can be got by Eq. \ref{equ:1}: 
\begin{equation}\label{equ:1}
C_k^t=\frac{1}{|NDSet_k^t|}\sum_{x \in NDSet_k^t}x, \quad \forall k \in \{1,2,\dots,n\}
\end{equation}
where $|NDSet_k^t|$ denotes the cardinality of the non-dominated set and $n$ is the dimensions of decision space. $C_k^t$ shows the center point in $t$th time step and $k$th dimension of decision space. And $NDSet_k^t$ is the non-dominated set in the $k$th dimension of decision space and $t$th time step. And $x$ represents a non-dominated individual at the $t$th time step and $k$th dimension of decision space.

The feed-forward center point strategy in the environmental response mechanism can be shown by Eq. (\ref{equ:2}).
\begin{equation}\label{equ:2}
NDInd_k^{t+1}=NDInd_k^{t}+(C_k^t-C_k^{t-1})+Gauss(0,d),\quad \forall k \in \{1,2,\dots,n\}
\end{equation}
where $C_k^t$ and $C_k^{t-1}$ show the center points in the $k$th dimension of decision space and $t$th and  $(t-1)$th time step, respectively.  $n$ is the dimension of decision space. $NDInd_k^{t}$ represents a non-dominated individual of $k$th dimension of decision space and $t$th time step in the nondominated set $NDSet_k^t$. And $NDInd_k^{t+1}$ denotes the predicted non-dominated individual of the $k$th dimension and $(t+1)$th time step in the predicted non-dominated set $NDSet_k^{t+1}$. $Gauss(0,d)$ is a Gaussian perturbation added in order to avoid falling into a local optimum. $d$ is the variance of disturbance. By Eq. \ref{equ:2}, the nondominated set in the ($t$+1)th time step, $NDSet^{t+1}$, can be got from the nondominated set in $t$th time step, $NDSet^{t}$.

Here, the number of non-dominated individuals in $NDSet^{t}$ is called $Nnd$. As it is known from Eq. (\ref{equ:2}), every individual in $NDSet^{t+1}$ is got from one individual in $NDSet^{t}$. Thus, the number of individuals in $NDSet^{t+1}$ is also $Nnd$.

After getting the  $NDSet^{t+1}$ by Eq. (\ref{equ:2}), every individual in $NDSet^{t+1}$ needs to be checked if the individual has been beyond the boundary range of the decision variables. If so, Algorithm \ref{alg:check} is used to revise the values of the individual in $NDSet^{t+1}$. In Algorithm \ref{alg:check}, $Ind_k$ and $Ind_k^{'}$ show the individuals $Ind$ and $Ind^{'}$ in the $k$th dimension, respectively.

\begin{figure}
	\centering	\includegraphics[width=0.7\textwidth]{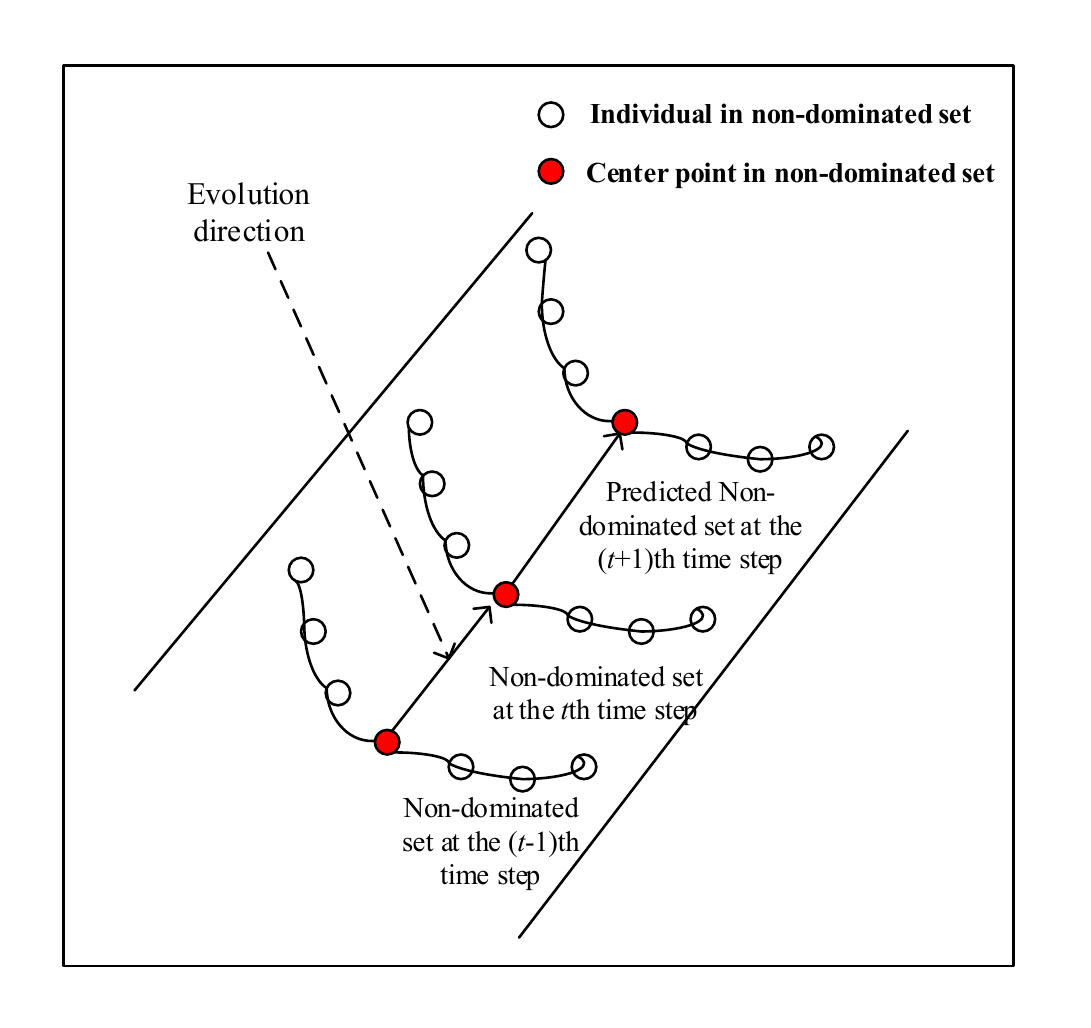}
	\caption{Environmental response mechanism.} \label{fig3}
\end{figure}

As shown in Fig. \ref{fig3}, the red points are the center points of the non-dominated sets. The evolution direction means the change direction of the center point of the non-dominated set at the ($t-1$)th time step to the center point of the non-dominated set at the $t$th step.  The non-dominated set at the $t$th time step moves along to the evolution direction, and then the predicted non-dominated set at the $(t+1)$th time step can be got. 
\begin{algorithm}[tbp]
	\renewcommand{\algorithmicrequire}{\textbf{Input:}}
	\renewcommand{\algorithmicensure}{\textbf{Output:}}
	\caption{Boundary check and processing}
	\label{alg:check}
	\begin{algorithmic}[1]
		\begin{small}
			\REQUIRE{
			$Ind$, one individual in the current population; $Ind^{'}$, the predicted individual from $Ind$; $low$ and $up$, the upper and lower bounds  in the decision vector; $n$, the dimension of the decision space.
			}
			\ENSURE{$Ind^{'}$, the processed  $Ind^{'}$.}
			\STATE    $k$=0;
			\WHILE {$k< n$}
			\IF{$Ind_k^{'} > up_k$}
		    \STATE     $Ind_k^{'} =0.5*(Ind_k+up_{k})$.
			\ENDIF 			
			 \IF { $Ind_k^{'} < low_k$}
			\STATE     $Ind_k^{'}=0.5*(Ind_k+low_{k}).$
			\ENDIF
			\STATE $k=k+1.$
			\ENDWHILE
		 \STATE   Return $Ind^{'}$.
		\end{small}
	\end{algorithmic}
\end{algorithm}
\subsubsection{Memory strategy}
\label{sec:3.1.2}
In the real world, the optimal solutions of some multi-objective problems change periodically. Therefore, this paper introduces a memory strategy. Since the memory strategy is not the focus of this paper, the simplest memory strategy is used here. In the strategy, some  fixed number of optimal individuals in the current environment are directly saved to the initial population $Pop^{t+1}$ after the next environmental change. Here, the number of memory individuals is called $Nmem$.
\subsubsection{Adaptive diversity maintenance strategy}
\label{sec:3.1.3}
The adaptive diversity maintenance strategy is also simple, just introducing random individuals into the population, together with the predicted non-dominated set and the memory set to form a population, as the initial population after an environmental change. This adaptive diversity maintenance strategy is very similar to that in \cite{ckps} \cite{dops}. The set made up by these random individuals is called the diversity set here. The size of the diversity set is called $Ndiv$.

When $Nnd+Nmem>Npop$, let $Nnd$ to be ($Npop-Nmem$) to revise the number of the non-dominated set, $NDSet^{t+1}$. And then  $Nnd$ individuals are selected randomly from $NDSet^{t+1}$ to form new $NDSet^{t+1}$.

In fact, the size of the diversity set is equal to the size of the population minus the size of the predicted non-dominated set and the size of the memory set, i.e., $Ndiv=Npop-Nnd-Nmem$.

In general cases, when a prediction strategy was used, if the problem is very difficult, more diverse individuals are usually needed. This is because when the problem is difficult, the prediction strategy is not very effective, which means the predicted non-dominated set is wrong. More diverse individuals can help find optimum solutions by evolution in time.
On the contrary, when the problem is very easy, less diverse individuals are usually needed. Since the prediction strategy can find the optimum solutions easily, more diverse individuals may affect the convergence speed instead. In the proposed diversity maintenance strategy, the gotten diverse individuals usually meet this rule. So this is a adaptive diversity maintenance strategy.
\subsubsection{Detailed process of Environmental response strategy}
\label{sec:3.1.4}
The detailed environmental response strategy is described in Algorithm \ref{alg:env}. Steps 1 and 2 describe the prediction strategy based on feed-forward center point. Step 3 does boundary check and processing to prevent individuals from exceeding the boundaries. Step 4 denotes the memory strategy. Steps 5, 6, and 7 are the adaptive diversity maintenance strategy. Step 8 gets the environmental response population by combining three sub-populations.
\begin{algorithm}[tbp]
	\caption{Environmental response strategy}
	\label{alg:env}
	\begin{small}
		\begin{description}
			\item[\textbf{\textit{Input}:}] $Nmem$, the capacity of the memory set; $Npop$, the size of the population; $NDSet^{t-1}$ and $NDSet^{t}$, the nondominated set in   ($t-1$)th and $t$th time step, respectively; the low and up boundary, $low$ and $up$.
			\item[\textbf{\textit{Output}:}] The environmental response population, $Pop^{t+1}$.
			\item[\textbf{\textit{Step 1}:}] Calculate the center points $C^t$ and $C^{t-1}$ by Eq. (\ref{equ:1}).
			\item[\textbf{\textit{Step 2}:}] Calculate the non-dominated set at the $(t+1)$th time step, $NDSet^{t+1}$, by Eq. (\ref{equ:2});
			 	\item[\textbf{\textit{Step 3}:}] Use Algorithm \ref{alg:check} to do boundary check and processing for every individual in $NDSet^{t+1}$.
			\item[\textbf{\textit{Step 4}:}] Select randomly $Nmem$ individuals from current population as the memory set, $MemSet^{t+1}$;
			\item[\textbf{\textit{Step 5}:}] if $Nmem+Nnd>Npop$, revise the size of $NDSet^{t+1}$, $Nnd$, to be $Npop-Nmem$. And then  $Nnd$ individuals are selected randomly from $NDSet^{t+1}$ as new $NDSet^{t+1}$.
			\item[\textbf{\textit{Step 6}:}] Calculate the size of the diversity set, $Ndiv$, by 
			 $Ndiv=Npop-Nnd-Nmem$.
			\item[\textbf{\textit{Step 7}:}] Calculate $Ndiv$ diverse individuals, $Divind^{t+1}$, by Eq. (\ref{equ:div}) to form the diversity set, $DivSet^{t+1}$.
			\begin{equation}\label{equ:div}
			Divind_k^{t+1}=rand(low_k,up_k), \quad \forall k \in \{1,2,\dots,n\},
			\end{equation}
			where $rand()$ is the random function and $n$ is the dimensions of decision space. $Divind_k^{t+1}$ denotes $Divind^{t+1}$ in the $k$th dimension of decision space.
			\item[\textbf{\textit{Step 8}:}] Get the environmental response population, $Pop^{t+1}$, by Eq. (\ref{equ:p}).
			\begin{equation}\label{equ:p}
			Pop_k^{t+1}=NDSet_k^{t+1} \cup MemSet_k^{t+1} \cup DivSet_k^{t+1}, \quad \forall k \in \{1,2,\dots,n\}.	
			\end{equation}
			\item[\textbf{\textit{Step 9}:}] Return environmental response population, $Pop^{t+1}$.
		\end{description}
	\end{small}
\end{algorithm}
\subsection{Generational response strategy}
In the environmental static stage, the feed-forward center point strategy is used to predict the convergence trend of the population according to the change direction of  the center points of the populations every several generations, thereby obtaining the predicted population after several generations. Choosing how many generations of population center points to predict is a matter of prediction step size. If the step size is larger, the prediction trend is farther, i.e., it may predict the population after several generations or dozens or more generations. If the step size is small (such as selecting the contact generation for prediction, i.e., the step size is 1), the population evolution trend of the next generation can be predicted. Then, the obtained predicted population and the current population are together for environmental selection and the selected population can be used for the evolution of the next generation. In this paper, for the sake of convenience, the step size is chosen to be 1, i.e., the center points of the populations of consecutive generations are used. Here, the obtained predicted population by the generational response strategy is called the generational response population.

%
%
In the generational response mechanism,  the center point in the $g$th generation, $C^{g}$, can be got by Eq. (\ref{equ:3}).

\begin{equation}\label{equ:3}
	C_k^g=\frac{1}{|NDSet_k^g|}\sum_{x \in NDSet_k^g} x, \quad \forall k \in \{1,2,\dots,n\}
\end{equation}
where $|NDSet_k^g|$ denotes the cardinality of the non-dominated set and $n$ is the dimensions of decision space. $NDSet_k^g$ is the non-dominated set in the $g$th generation. And $x$ represents a non-dominated individual in the $g$th generation and $k$th dimension. 

The prediction process of the population in the $(g+1)$th generation can be expressed by Eq. (\ref{equ:4}).
\begin{equation}\label{equ:4}
	Pop_k^{g+1}=Pop_k^{g}+(C_k^g-C_k^{g-1})+Gauss(0,d), \quad \forall k \in \{1,2,\dots,n\}
\end{equation}
where $C_k^g$ and $C_k^{g-1}$ show the center points in the $k$th dimension and $g$th and  $(g-1)$th generation, respectively. $n$ is the dimensions of decision space. $Pop_k^{g}$ and $Pop_k^{g+1}$ represent the population in $k$th dimension and $g$th and $g+1 $th generation, respectively. $Gauss(0,d)$ is a Gaussian perturbation added in order to avoid falling into a local optimum. $d$ is a variance of disturbance. By Eq. \ref{equ:4}, the population in the $(g+1)$th generation, $Pop^{g+1}$, can be got from the population in the $g$th generation, $Pop^{g}$.

After getting $Pop^{g+1}$, each individual in  $Pop^{g+1}$ needs also to be checked if its boundary has been beyond the boundary range of the decision variables. If so, Algorithm \ref{alg:check} is used to revise the values of the individual.

\begin{figure}
	\centering	\includegraphics[width=0.7\textwidth]{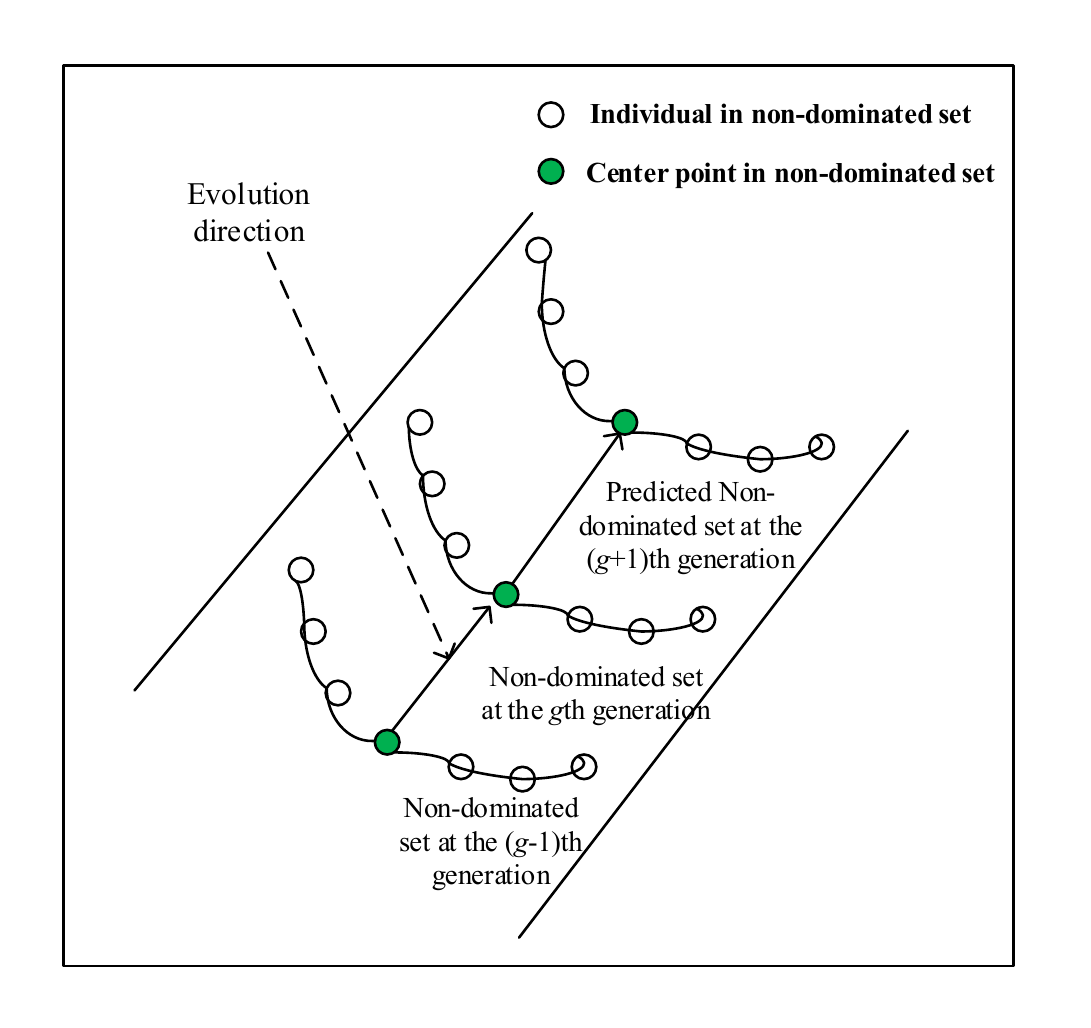}
	\caption{Generational response mechanism (carried out in decision space).} \label{fig4}
\end{figure}

As shown in Fig. \ref{fig4}, the green points are the center points of the non-dominated sets. The evolution direction means the change direction of the center point of the non-dominated set of the $(g-1)$th generation to the center point of the non-dominated set of the $g$th generation.  The population of the $g$th generation moves along to the evolution direction, and then the predicted population of the $(g+1)$th generation can be got. 

The detailed strategy is shown in Algorithm \ref{alg:gen}. Step 3 calculates the prediction population in the $ (g+1) $th generation. Step 4  does boundary check and processing to prevent the prediction population from exceeding the boundaries. Step 5 does environmental selection from mixed population.
\begin{algorithm}[tbp]
	\caption{Generational response strategy}
	\label{alg:gen}
	\begin{small}
		\begin{description}
			\item[\textbf{\textit{Input}:}]  $Pop^{g-1}$ and $Pop^{g}$, the population in the ($g-1$)th and $g$th generation, respectively.
			\item[\textbf{\textit{Output}:}] The generational response population, $Pop^{g+1}$.
			
			\item[\textbf{\textit{Step 1}:}] Calculate the non-dominated sets, $NDSet^{g-1}$ and $NDSet^{g}$;
			\item[\textbf{\textit{Step 2}:}] Calculate the center points, $C^g$ and $C^{g-1}$ by Eq. (\ref{equ:3}).
			\item[\textbf{\textit{Step 3}:}] Calculate the generational response population, $Pop^{g+1}$, by Eq. (\ref{equ:4}).
			\item[\textbf{\textit{Step 4}:}] Do boundary check  and processing for each individual in  $Pop^{g+1}$ by Algorithm \ref{alg:check}.
        	\item[\textbf{\textit{Step 5}:}] Combine $Pop^{g+1}$ and $Pop^{g}$ to do environmental selection \cite{nsga2} to get new $Pop^{g+1}$.
			\item[\textbf{\textit{Step 6}:}] Return generational response population, $Pop^{g+1}$.
		\end{description}
	\end{small}
\end{algorithm}

\subsection{Proposed FGERS-CPS and time complexity analysis}
\subsubsection{Detailed process of FGERS-CPS}
The detailed process of FGERS-CPS is in Algorithm \ref{alg:dcaf}. Step 2  does the environmental change detection. If an environment changes, go to the Step 3 to do environmental response strategy; else, turn to Step 4 to do generational response strategy.
\subsubsection{Time complexity analysis}
Set M, N, D to be objective number, population size, and the number of decision variables, respectively. In the Algorithm \ref{alg:dcaf}, the time complexity of calculating the non-dominated set in Step 2 is $ O(MN^{2}) $.  The time complexity of the environmental response strategy is $O(D) $. In the generational response strategy, the time complexity of calculating the non-dominated set in Step 1 is $ O(MN^{2}) $. And the time complexity of boundary check and processing in Step 4 is $ O(D) $. In addition, the time complexity of the environmental selection in Step 5 is $ O(MNlogN) $. So the time complexity of the generational response strategy is $ O(MN^{2})+O(D) $. Therefore, the time complexity of the proposed FGERS-CPS is $ O(MN^{2})+O(D) $.
\begin{algorithm}[tbp]
	\caption{The feed-forward center point strategy in the novel dynamic convergence-accelerated framework.}
	\label{alg:dcaf}	
	\begin{small}
		\begin{description}
			\item[\textbf{\textit{Initialization}:}] number of time change, $t=0$; generation counter, $g =0$; total generation number, $gmax$. 
			\item[\textbf{\textit{Step 1}:}] Initialize the population, $Pop^{t}$ and $Pop^{g}$.
			\item[\textbf{\textit{Step 2}:}] Detect the environmental change. If change, calculate the non-dominated set, $NDSet^t$, and  $t=t+1$;
			else,  go to \textit{Step 4}.
			\item[\textbf{\textit{Step 3}:}] The environmental response strategy (Algorithm \ref{alg:env}) is used to respond the environmental change.
			\item[\textbf{\textit{Step 4}:}] The generational response strategy (Algorithm \ref{alg:gen})  is run to accelerate the convergence of the population.  
			\item[\textbf{\textit{Step 5}:}] The optimization algorithm is used to get optimal solutions for the problem.
			\item[\textbf{\textit{Step 6}:}] If $g>gmax$, output $Pop^{t}$, and then end; else, $g=g+1$; go to \textit{Step 2}.
		\end{description}
	\end{small}
\end{algorithm}

\section{Test instances and performance indicators}
\label{sec:4.1}
\subsection{Test instances}
\label{sec:4.1.1}

For DMOPs, there are many classification methods. Among them, the most common method is proposed by \cite{fda}. In \cite{fda}, DMOPs are classified into four types according to changes of PF and PS. They are as follows.
\begin{itemize}
	\item Type 1: PS changes over time, but PF doesn't change.
	\item Type 2: Both PS and PF change over time.
	\item Type 3: PF changes, but PS doesn't change.
	\item Type 4: PF and PS don't change over time.
\end{itemize}
In this paper, some frequently used benchmark DMOPs are used, such as FDA test suite \cite{fda}, dMOP test suite \cite{dmop}, and F5-F10 \cite{pps}. Among them, decision variables of the problems in FDA and dMOP test suites are liner correlation. F5-F10 are more complex problems and their decision variables are nonlinear correlation. Besides, FDA4 and F8 are problems with three objectives. The rest are the problems with two objectives. The detailed information about benchmark DMOPs is listed in Table \ref{tab:test}.
\begin{table*}
	\caption{The detailed information about benchmark DMOPs.}
	\centering
	\label{tab:test}
	\scriptsize
	\begin{tabular}{C{1.5cm}L{1.2cm}C{2.2cm}C{5.0cm}}
		\hline\noalign{\smallskip}
		Problems & Type & Objective number &Relationship between decistion variables \\
		\noalign{\smallskip}\hline\noalign{\smallskip}
		
		FDA1    &Type 1           &2	&Linear\\
		FDA2    &Type 3           &2    &Linear\\
		FDA3    &Type 2           &2    &Linear  \\
		FDA4    &Type 1           &3	&Linear \\	
		dMOP1   &Type 3    &2      &Linear\\
		dMOP2   &Type 2    &2      &Linear\\
		dMOP3   &Type 1    &2      &Linear\\
		F5      &Type 2    &2      &Non-linear\\
		F6      &Type 2    &2      &Non-linear\\
		F7      &Type 2    &2      &Non-linear\\
		F8      &Type 2    &3      &Non-linear\\
		F9      &Type 2    &2      &Non-linear\\
		F10     &Type 2    &2      &Non-linear\\
		\noalign{\smallskip}\hline
	\end{tabular}
\end{table*}
\subsection{Performance indicators}
\label{sec:4.1.2}
Two commonly used indicators, MIGD \cite{sgea} \cite{pps} and MHVD \cite{sgea} \cite{zhoumeeting} \cite{zhang2019}, were adopted to calculate the performance of these five strategies. MIGD is an improved version of IGD \cite{cao6}.
Here, let {$PF_{t}$} be a true set of uniformly distributed  Pareto optimal points of PF at the $t$th
time step and let {$P_t$} be an approximation set of PF at the $t$th time step. The mathematical form of IGD is described in Eq. (\ref{equ:5}).
\begin{equation}\label{equ:5}
IGD(PF_t, P_t)=\frac{\sum_{v \in PF_t } d(v,P_t)}{|PF_t|},
\end{equation}
where
$d(v,P_t)=min_{u \in P_t}|| F(v)-F(u)||$ is the distance between $v$ and $P_t$. $|PF_t|$ is the cardinality of $PF_t$.

MIGD, like IGD, is a comprehensive indicator that can assess not only the convergence performance of the algorithm but also the diversity. The smaller the MIGD values, the better the performance of the algorithm. The mathematical form of MIGD is described in Eq. (\ref{equ:6}).
\begin{equation}\label{equ:6}
MIGD=\frac{1}{|T|}\sum_{t \in T}IGD(PF_t, P_t),
\end{equation}
where $T$ represents a set of discrete time points in a run and $|T|$ is the cardinality of $T$.

Similarly, the Hypervolume Difference(HVD) \cite{sgea} \cite{zhoumeeting} is an improved version of HV \cite{hv0} \cite{tian}.  HVD measures the gap between the hypervolumes of the got PF and  the true PF. $PF_t$  and  $P_t$ are same as those in Eq. (\ref{equ:5}). The mathematical form of HVD is described in Eq. (\ref{equ:7}).
\begin{equation}\label{equ:7}
HVD(PF_t,P_t)=HV(PF_t)-HV(P_t),
\end{equation}
where $HV(S)$ denotes the hypervolume of a set $S$.

MHVD is got by modifying HVD like MIGD to IGD. MHVD shows the average of the HVD values in some time steps over a run. The mathematical form of MHVD is described in Eq. (\ref{equ:8}).
\begin{equation}\label{equ:8}
MHVD=\frac{1}{|T|}\sum_{t \in T}HVD(PF_t, P_t),
\end{equation}
where $T$ is a set of discrete time points over a run and $|T|$ is the cardinality of $T$. The reference point for computing hypervolume is $(Z_1^t+0.5, Z_2^t+0.5, \dots, Z_M^t+0.5)$, where $Z_j^t$ is the maximum value of the $j$th objective of the true PF at the $t$th time step; $M$ is the number of objectives. MHVD is also a comprehensive indicator. The smaller the MHVD values, the better the performance of the algorithm.\\

\section{Experimental results and analysis}
\label{sec:4}
\subsection{Parameter settings}
\label{sec:4.2}
Four classical strategies were compared with FGERS-CPS. The four strategies are as follows: (1) \textit{randomly initialize strategy} (\textit{RIS}) \cite{pps} \cite{ckps}; (2) \textit{the feed-forward prediction strategy} (\textit{FPS}) \cite{fps}; (3) \textit{population prediction strategy} (\textit{PPS}) \cite{pps}; (4) \textit{the prediction strategy based on special points} (\textit{SPPS}) \cite{spps}, which predicts the PF by some special points. The optimization algorithm uses RM-MEDA \cite{rm} in these five strategies. Besides, the Gaussian perturbation $d$ is 0.1. The data of RIS, FPS, PPS, and SPPS in Tables \ref{tab:2} and \ref{tab:3} are got from \cite{spps}\\
\indent The other parameter settings are as follows.
\begin{itemize}
	\item \textit{The problem parameters}: The environmental change severity $n_t$ is 10. The dimensions of decison space of the test problems are 20.
	\item \textit{The population size}: The population size, $Npop$, for all test problems is 100 and the size of the memory set, $Nmem$, is 10.
	\item \textit{Stopping criterion and the number of runs}: The total run generation number is 2500 and the environmental change frequency $\tau_t$ is 25, i.e., the number of environmental changes is 100. Run each algorithm 20 times for each test problem independently.
	\item \textit{The parameters in FPS}:  The objective number is $m$. The number of predicted individuals is $3(m+1)$. Seventy percent of the other individuals in the population are inherited from the best individuals in the current environment. And the other 30 percent of individuals are randomly generated in the decision space \cite{spps}.
	\item \textit{The parameters in PPS}: The parameter $p$ is 3; the length $M$ of the historical information sequence in the AR($p$) model is 23 \cite{spps}.
	\item \textit{The parameters in SPPS}: The Gaussian perturbation $d$ is 0.1.
	\item \textit{Environmental change detection}: There are some detection methods with strong power \cite{sahmoud2019exploiting} \cite{sahmoud2019hybrid} \cite{sahmoud2018type}, which can detect complex environmental changes. However, for convenience,  five percent of the individuals are selected to be reevaluated to detect environmental changes. If the individual's objective values are found to be inconsistent with the original values, it proves that the environment has changed \cite{pps}. Since the detection method is not the focus of this paper, the focus is to propose a novel dynamic framework here. So, the environmental changes of DMOPs used are deterministic, which means no noise in DMOPs. In addition, we assume when environment changes, the all fitness landscape  will change. Therefore, in this situation, this detection method always works.
\end{itemize}
\begin{table*}
	\scriptsize
	\caption{Mean and Standard Deviation of  MIGD values of five strategies on FDA, dMOP and F test suites. The values in bold face denote to have the best effect on these five strategies. \ddag and \dag indicate FGERS-CPS is significantly better than and equivalently to the corresponding strategy, respectively.}
	\centering
	\label{tab:2}
	\noindent\makebox[\textwidth][c]
	{
		\begin{tabular}{L{1cm}L{2cm}L{2cm}L{2cm}L{2.2cm}L{2cm}}
			\hline\noalign{\smallskip}
			Problems &RIS              &FPS                &PPS                 &SPPS       &FGERS-CPS    \\
			\noalign{\smallskip}\hline\noalign{\smallskip}
			FDA1	&1.3155(0.0303)	&0.0516(0.0086)\ddag	&0.0528(0.0091)\ddag	&0.0258(0.0048)\ddag	&\textbf{0.0109(0.0001)}\\
			FDA2	&0.0500(0.0008)\ddag	&\textbf{0.0085(0.0007)}\dag	&0.0097(0.0008)\ddag	&0.0089(0.0004)\ddag	&\textbf{0.0085(0.0006)}\\
			FDA3	&1.7564(0.0655)\ddag	&0.0645(0.0093)\ddag	&0.0941(0.0158)\ddag	&0.0383(0.0076)\ddag	&\textbf{0.0125(0.0007)}\\
			FDA4	&0.4566(0.0092)\ddag	&0.1414(0.0034)	&0.1307(0.0020)	&\textbf{0.1114(0.0030)}	&0.1465(0.0006)\\
			dMOP1	&0.6386(0.0143)\ddag	&\textbf{0.0072(0.0012)}	&0.0379(0.0515)\ddag	&0.0097(0.0021)\dag	&0.0090(0.0001)\\
			dMOP2	&1.6968(0.0541)\ddag	&0.0622(0.0079)\ddag	&0.0607(0.0102)\ddag	&0.0279(0.0040)\ddag	&\textbf{0.0129(0.0002)}\\
			dMOP3	&1.3215(0.0375)\ddag	&0.0523(0.0065)\ddag	&0.0527(0.0108)\ddag	&0.0272(0.0041)\ddag	&\textbf{0.0108(0.0002)}\\
			F5	    &1.1439(0.0418)\ddag	&0.1852(0.0819)\ddag	&0.2323(0.0773)\ddag	&0.0350(0.0079)\ddag	&\textbf{0.0201(0.0014)}\\		   
			F6	    &0.5399(0.0122)\ddag	&0.0548(0.0168)\ddag	&0.0751(0.0424)\ddag	&0.0206(0.0023)\ddag	&\textbf{0.0154(0.0005)}\\
			F7	    &0.6165(0.0153)\ddag	&0.1273(0.0234)\ddag	&0.1006(0.0402)\ddag	&0.0203(0.0030)\ddag	&\textbf{0.015(0.0007)}\\
			F8	    &0.9083(0.0248)\ddag	&\textbf{0.1418(0.0036)}	&0.1455(0.0046)	&0.1423(0.0048)	&0.1688(0.0041)\\
			F9	    &1.1923(0.0325)\ddag	&0.3542(0.0675)\ddag	&0.6186(0.1948)\ddag	&\textbf{0.1078 (0.0189)}	&0.2070(0.0904)\\
			F10	    &1.0691(0.0472)\ddag	&0.4280(0.0531)\ddag	&0.5097(0.0998)\ddag	&0.0720 (0.0205)\ddag	&\textbf{0.0625(0.0073)}\\
			\noalign{}\hline
		\end{tabular}
	}
\end{table*}
\subsection{Performance evaluations}
\label{sec:4.3}
Tables \ref{tab:2} and \ref{tab:3} show the mean and standard deviation of MIGD and MHVD values, respectively.  The best values of the five algorithms are marked in bold. The Wilcoxon rank sum test \cite{wilcoxon} was run to indicate significance at a significant level of 0.05 between different results.

\begin{table*}
	\scriptsize
	\caption{Mean and Standard Deviation of  MHVD values of five strategies on FDA, dMOP and F test suites. The values in boldface denote to have the best effect in these five strategies. \ddag and \dag indicate FGERS-CPS is significantly better than and equivalently to the corresponding strategy, respectively.}
	\centering
	\label{tab:3}
	\noindent\makebox[\textwidth][c]
	{
		\begin{tabular}{L{1cm}L{2cm}L{2cm}L{2cm}L{2cm}L{2cm}}
			\hline\noalign{\smallskip}
			Problems &RIS              &FPS                &PPS                 &SPPS       &FGERS-CPS    \\
			\noalign{\smallskip}\hline\noalign{\smallskip}
			FDA1	&1.2328(0.0107)\ddag	&0.0968(0.0105)\ddag	&0.0948(0.0145)\ddag	&0.0422(0.0030)\ddag	&\textbf{0.0244(0.0003)}\\
			FDA2	&0.0714(0.0013)\ddag	&0.0320(0.0008)\dag	&0.0325(0.0007)\ddag	&0.0324(0.0006)\ddag	&\textbf{0.0311(0.0007)}\\
			FDA3	&1.9361(0.0177)\ddag	&0.7761(0.0183)\ddag	&0.8420(0.0275)\ddag	&0.6592(0.0056)\ddag	&\textbf{0.6071(0.0014)}\\
			FDA4	&1.3189(0.0178)\ddag	&0.2449(0.0047)	&0.2383(0.0051)	&\textbf{0.2166(0.0032)}	&0.4209(0.0042)\\
			dMOP1	&1.1531(0.0182)\ddag	&0.1501(0.0010)\ddag	&0.1688(0.0311)\ddag	&\textbf{0.1463(0.0013)}	&0.1495(0.0003)\\
			dMOP2	&1.2672(0.0113)\ddag	&0.2156(0.0131)\ddag	&0.2190(0.0154)\ddag	&0.1658(0.0021)\ddag	&\textbf{0.1530(0.0004)}\\
			dMOP3	&1.2282(0.0095)\ddag	&0.0972(0.0085)\ddag	&0.0947(0.0147)\ddag	&0.0434(0.0024)\ddag	&\textbf{0.0244(0.0004)}\\
			F5	    &1.3849(0.0142)\ddag	&0.4745(0.0707)\ddag	&0.4845(0.0535)\ddag	&0.2821(0.0116)\ddag	&\textbf{0.2590(0.0039)}\\
			F6	    &0.9995(0.0153)\ddag	&0.3050(0.0325)\ddag	&0.3436(0.0487)\ddag	&0.2668(0.0056)\ddag	&\textbf{0.2540(0.0016)}\\
			F7	    &1.0988(0.0152)\ddag	&0.4448(0.0439)\ddag	&0.3625(0.0288)\ddag	&0.2642(0.0065)\ddag	&\textbf{0.2530(0.0017)}\\
			F8	    &2.6165(0.0143)\ddag	&\textbf{0.2569(0.0117)}	&0.3563(0.0353)	&0.2877(0.0073)	&0.4721(0.0126)\\
			F9	    &1.3930(0.0140)\ddag	&0.6393(0.0872)\ddag	&0.6481(0.0728)\ddag	&\textbf{0.3444(0.0121)}	&0.4550(0.0895)\\
			F10	    &1.3534(0.0188)\ddag	&0.8905(0.0706)\ddag	&0.7476(0.0559)\ddag	&0.3352(0.0252)\ddag	&\textbf{0.3198(0.0081)}\\
			\noalign{}\hline
		\end{tabular}
	}
\end{table*}

 The experimental results are shown in Table \ref{tab:2}, in which five algorithms were compared on 13 benchmark problems for the MIGD indicator. The figures in each cell denote the mean and standard deviation. As can be seen from Table \ref{tab:2}, on FDA1-FDA3 and dMOP2-dMOP3, FGERS-CPS is better than the other strategies. This shows that FGERS-CPS has better performance on the two-objective problems with linear correlation between decision variables. On dMOP1, FGERS-CPS is slightly worse than FPS but better than RIS, FPS, PPS and SPPS. This is  because dMOP1 is a Type 3 problem with fixed PS, and 70 percent of the individuals in the predicted population of FPS inherit the optimal solutions in the current environment. Therefore, FPS is better than FGERS-CPS on dMOP1. On the 3-dimensional benchmark problem FDA4, FGERS-CPS is worse than SPPS. This should be attributed to the guidance of multiple special points in SPPS so that SPPS can converge faster. 
 
 FGERS-CPS is significantly better than the other strategies on the benchmark problems F5-F7 with the nonlinear relationship between decision variables. This shows that FGERS-CPS also has a good effect on the more complex nonlinear benchmark problems. On the three-dimensional problem F8, the effect of FGERS-CPS is worse than FPS, PPS, and SPPS. This denotes that the effect of FGERS-CPS on three-dimensional problems needs to be improved, which is a direction of future research. F9 is a more complicated problem than F5-F7. On F9, when the environment changes, PS will change slightly most of the time, but sometimes, its PS will jump from one area to another area. On F9, FGERS-CPS is slightly worse than SPPS, but it is significantly better than RIS, FPS and PPS. This may be due to the guiding effect of multiple special points in SPPS that makes SPPS have a faster convergence speed than FGERS-CPS. F10 is almost the most complicated test problem here. On F10, the shapes of two consecutive PFs are different. As can be seen from Table \ref{tab:2}, FGERS-CPS is significantly better than the other strategies on F10, which shows that FGERS-CPS is also quite competitive on complex problems.
 
 Table \ref{tab:3} shows the mean and standard deviation of the MHVD values of the five strategies on 13 test problems. From Table \ref{tab:3}, we can see that the comparison effect of these strategies is mostly consistent with Table \ref{tab:2}. One of the inconsistencies is on dMOP1. In Table \ref{tab:3}, the MHVD values of SPPS are better than FPS, which is different in Table \ref{tab:2}. The reason for this phenomenon may be caused by the different characteristics of MHVD and MIGD indicators. Compared with the objective function values got by FPS, MHVD is more inclined to those obtained by SPPS.
 \begin{figure*}[htbp]
 	\begin{center}
 		\footnotesize
 		\begin{tabular}{ccccc}
 			\includegraphics[scale=0.2]{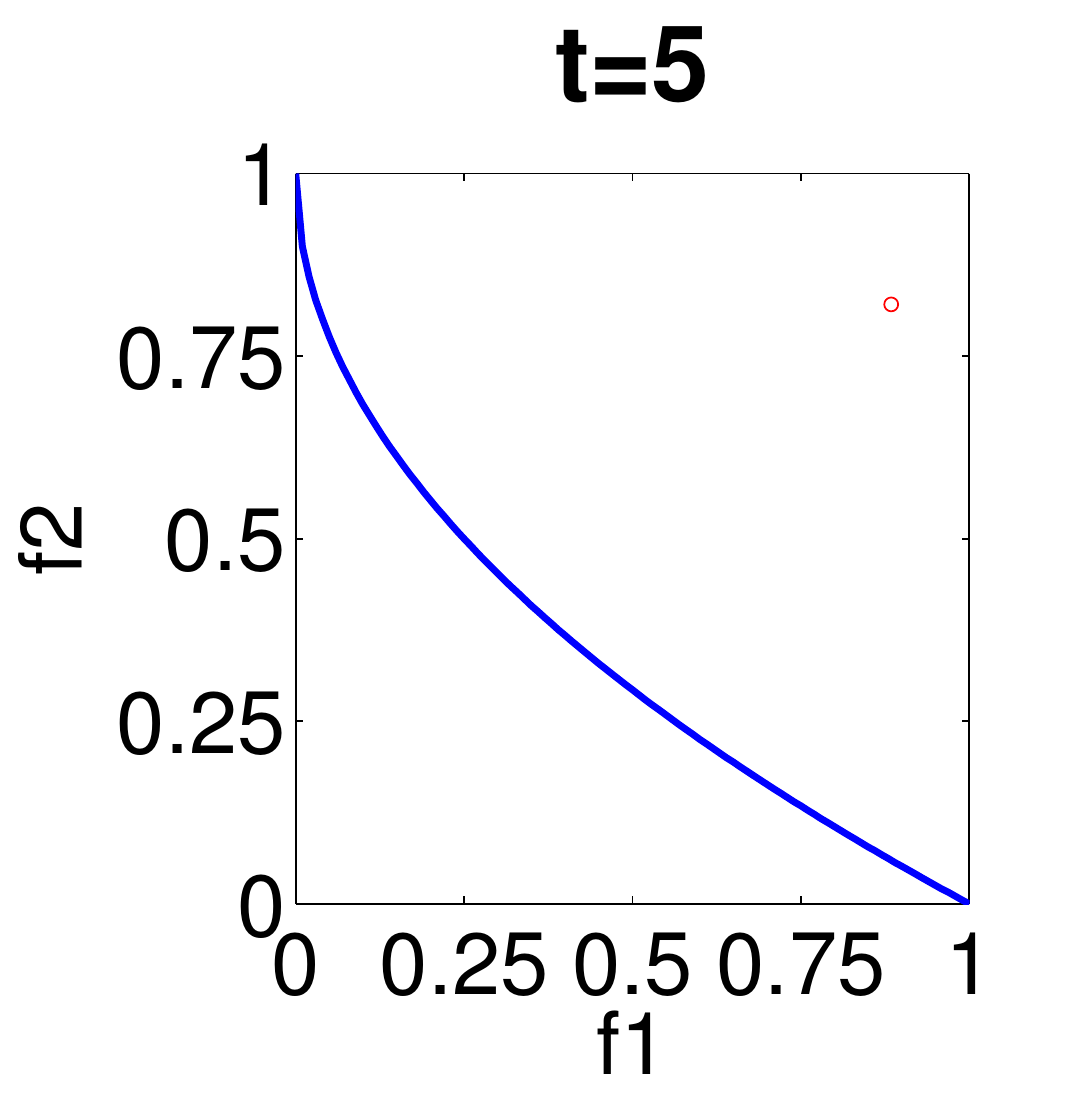}
 			\includegraphics[scale=0.2]{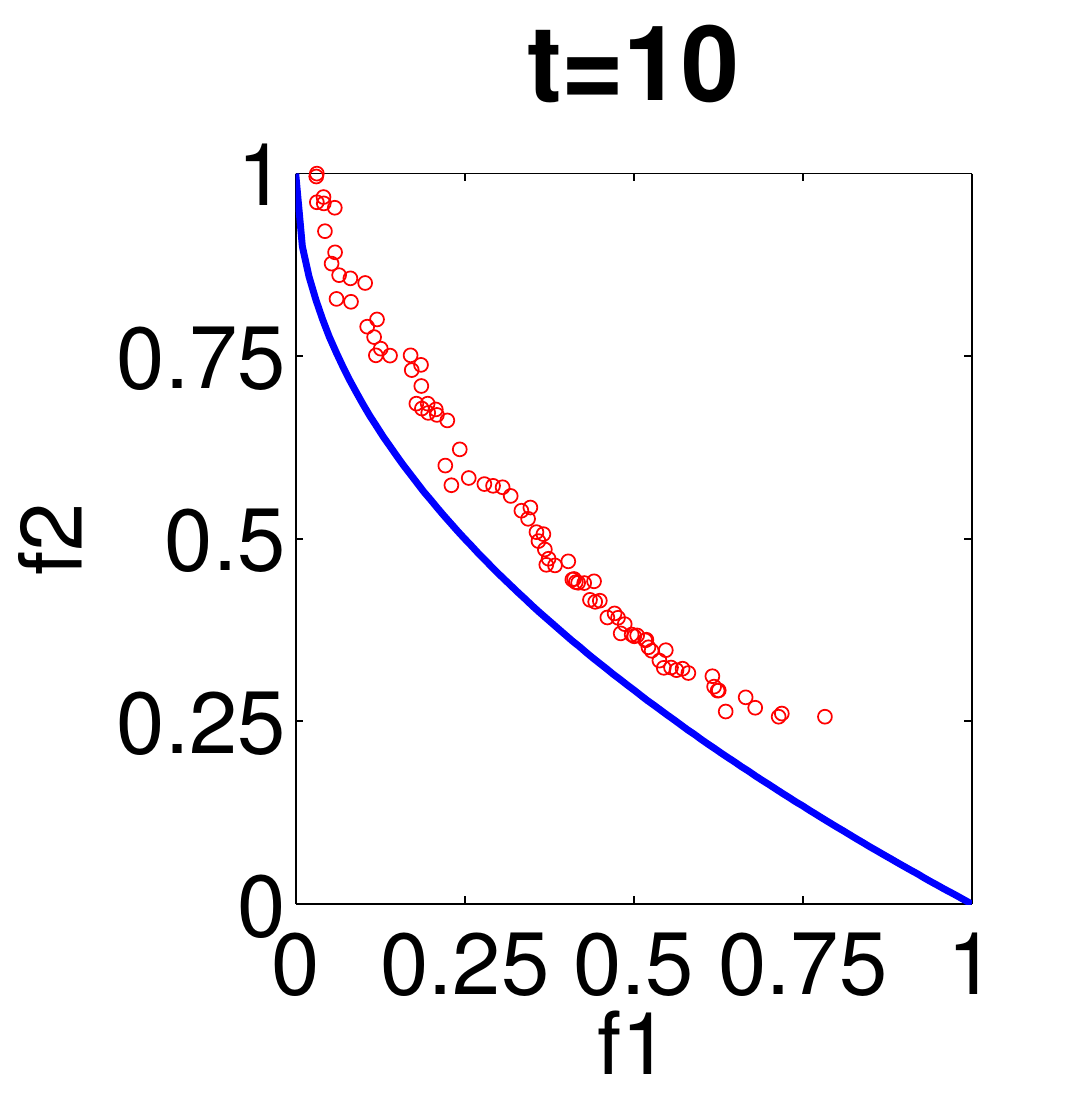}
 			\includegraphics[scale=0.2]{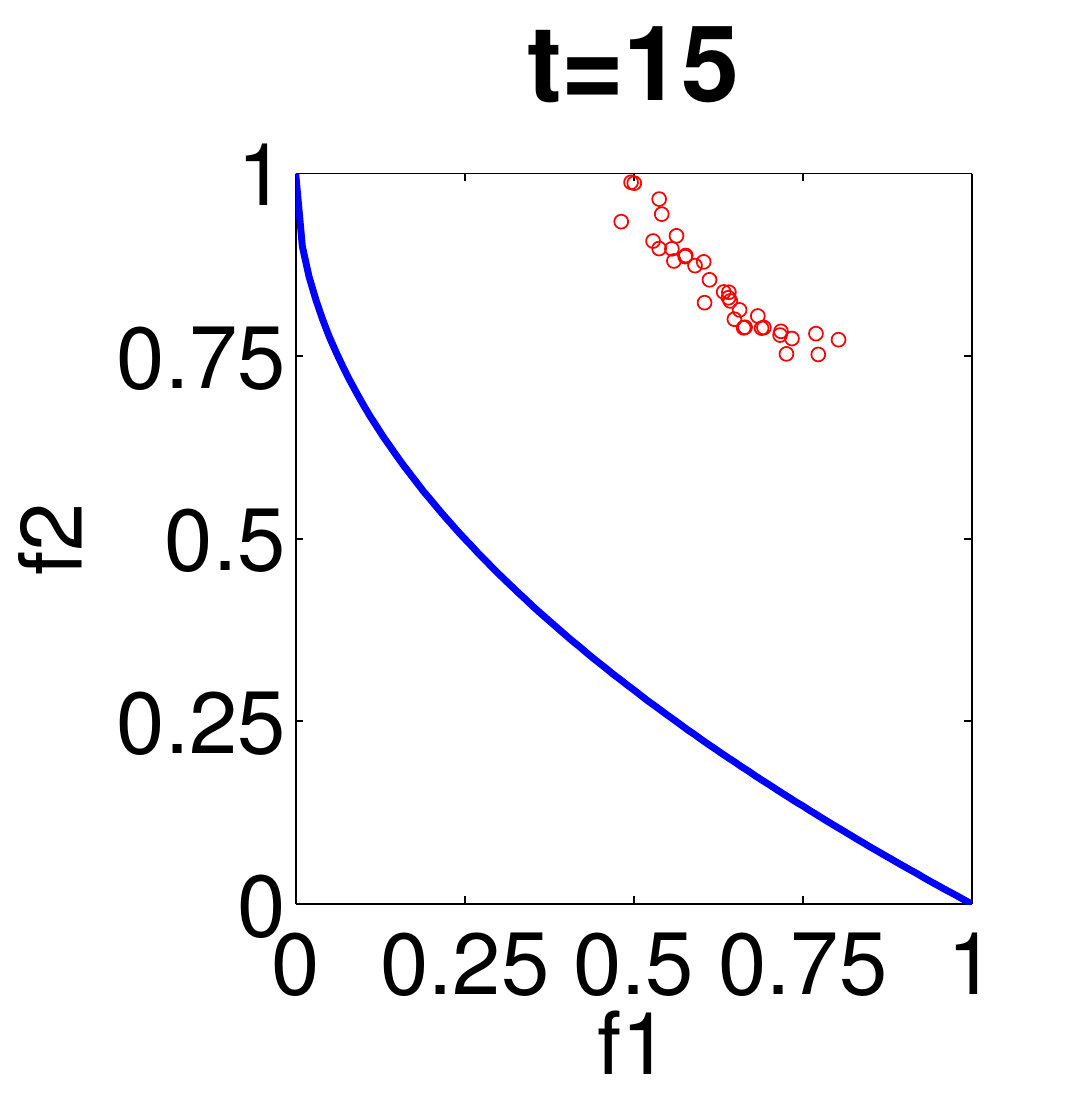}
 			\includegraphics[scale=0.2]{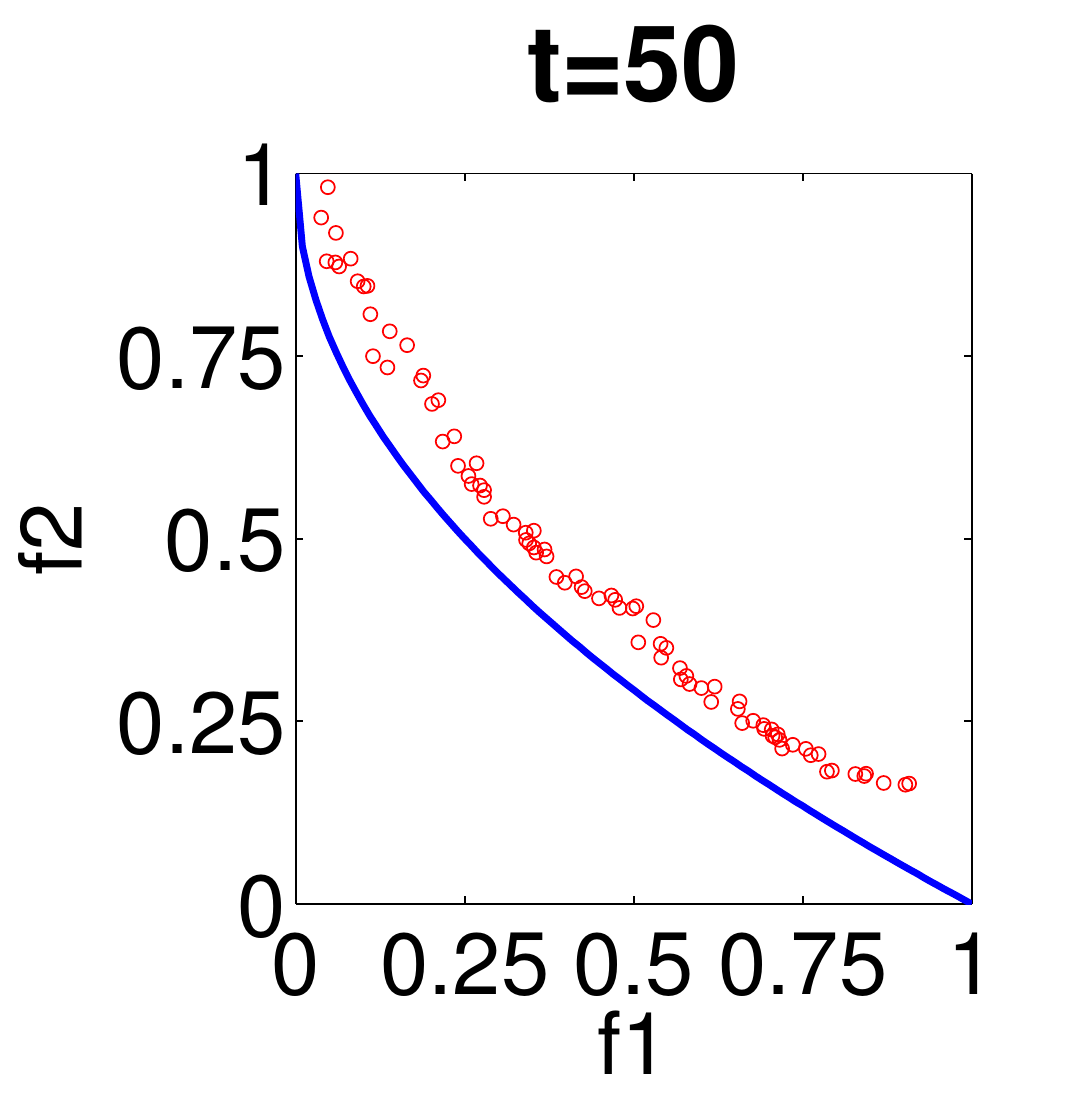}
 			\includegraphics[scale=0.2]{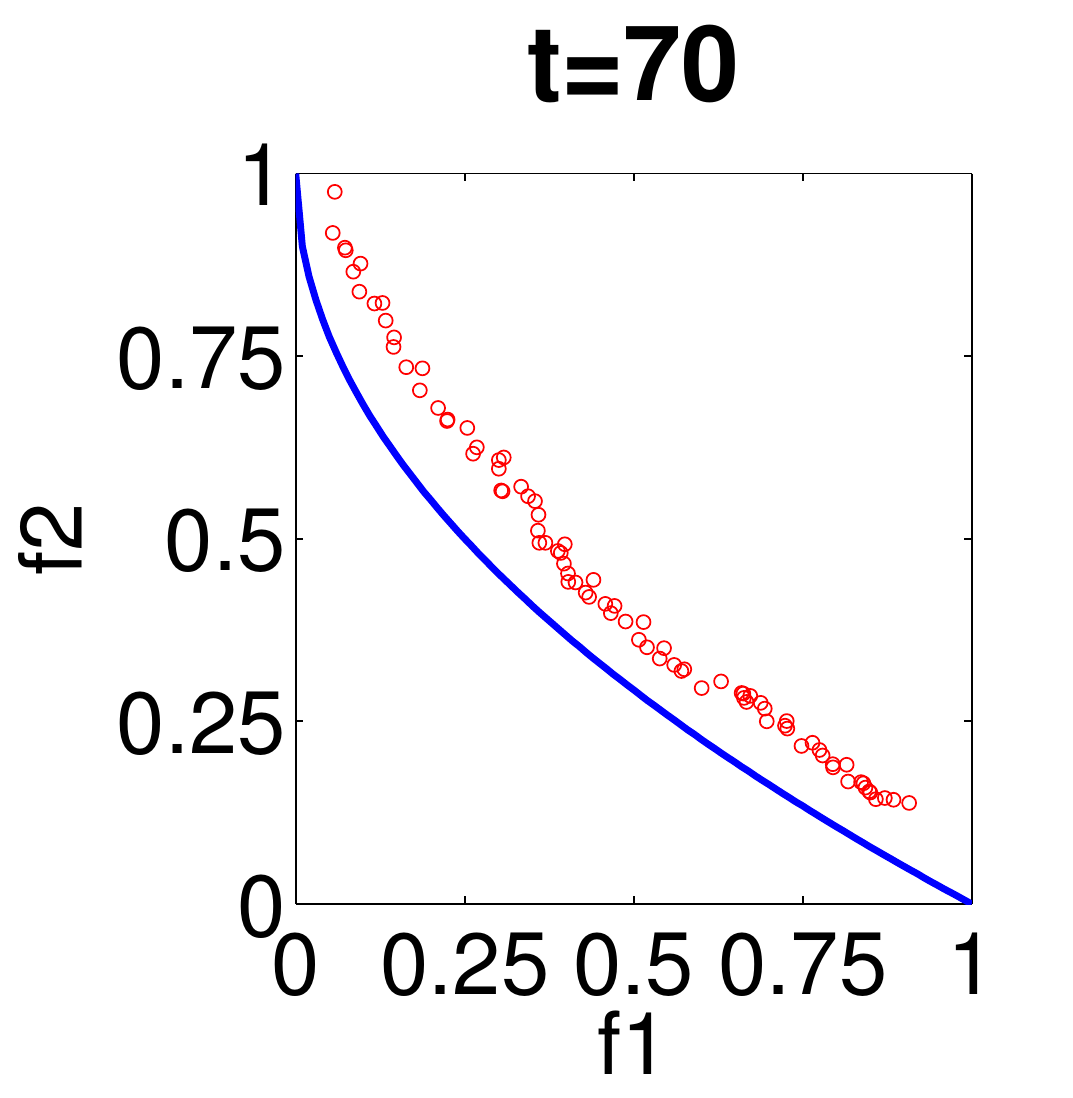}
 			\includegraphics[scale=0.2]{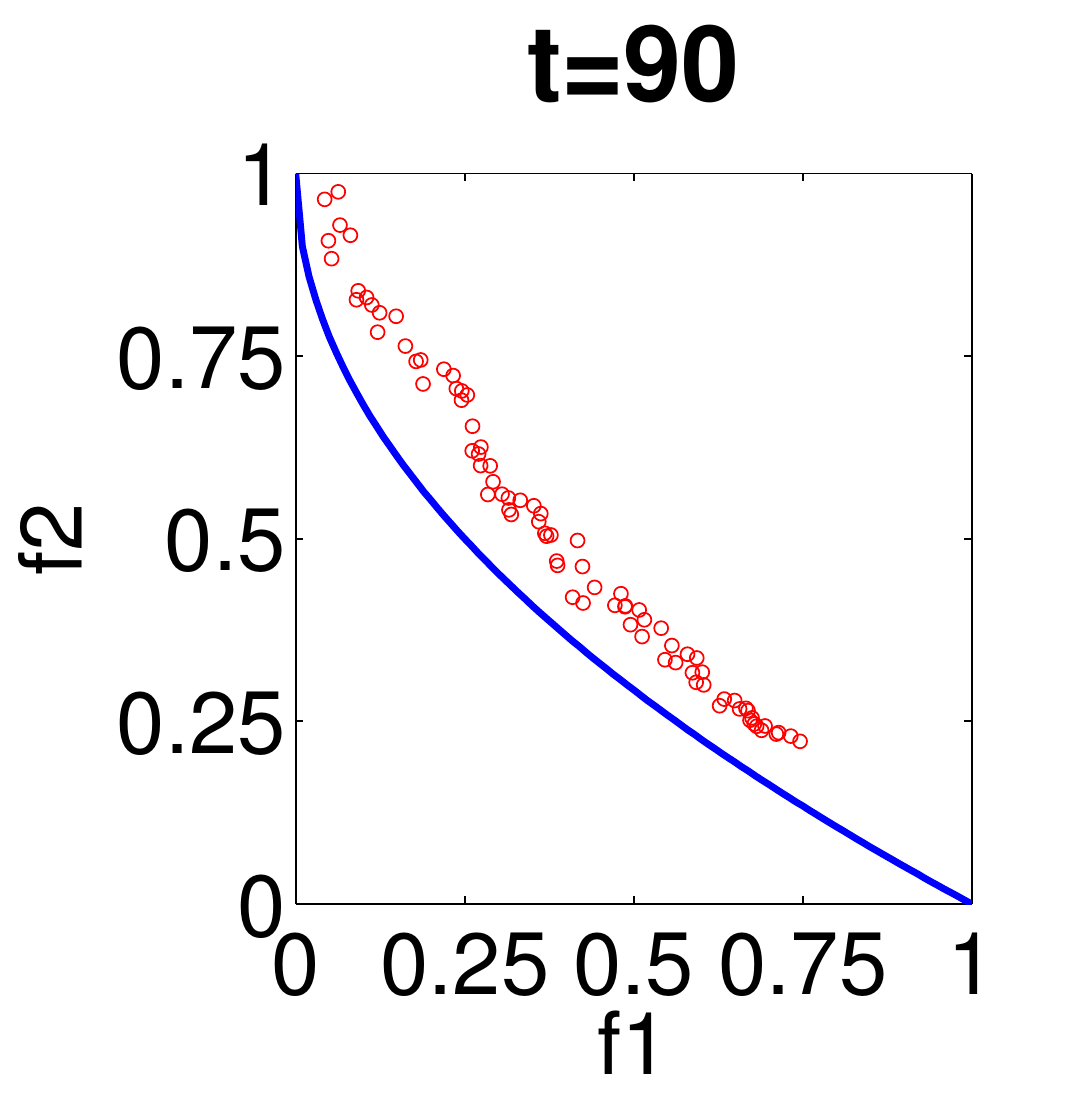}\\
 			\textbf{	(a) RIS}\\
 			\includegraphics[scale=0.2]{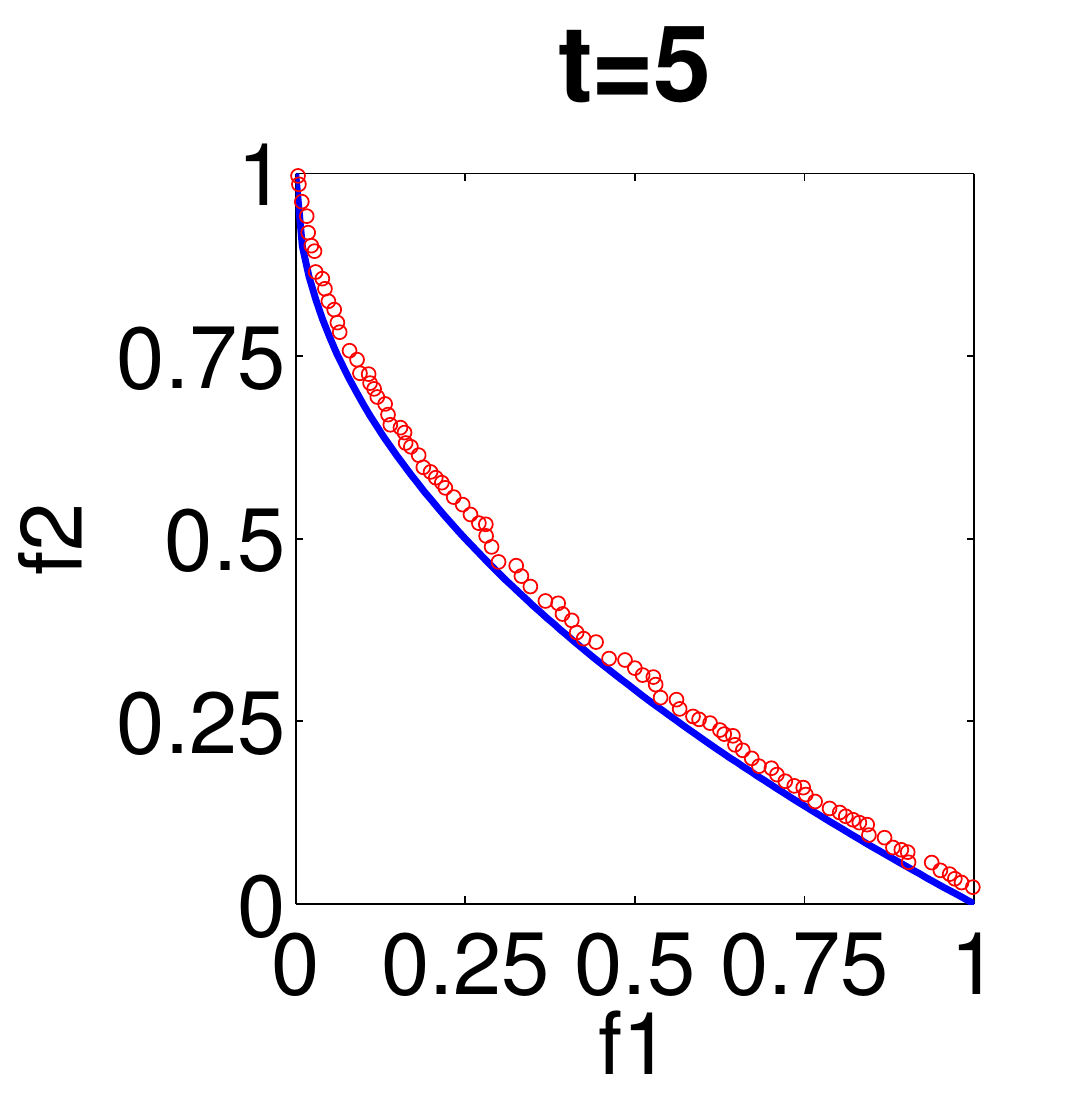}
 			\includegraphics[scale=0.2]{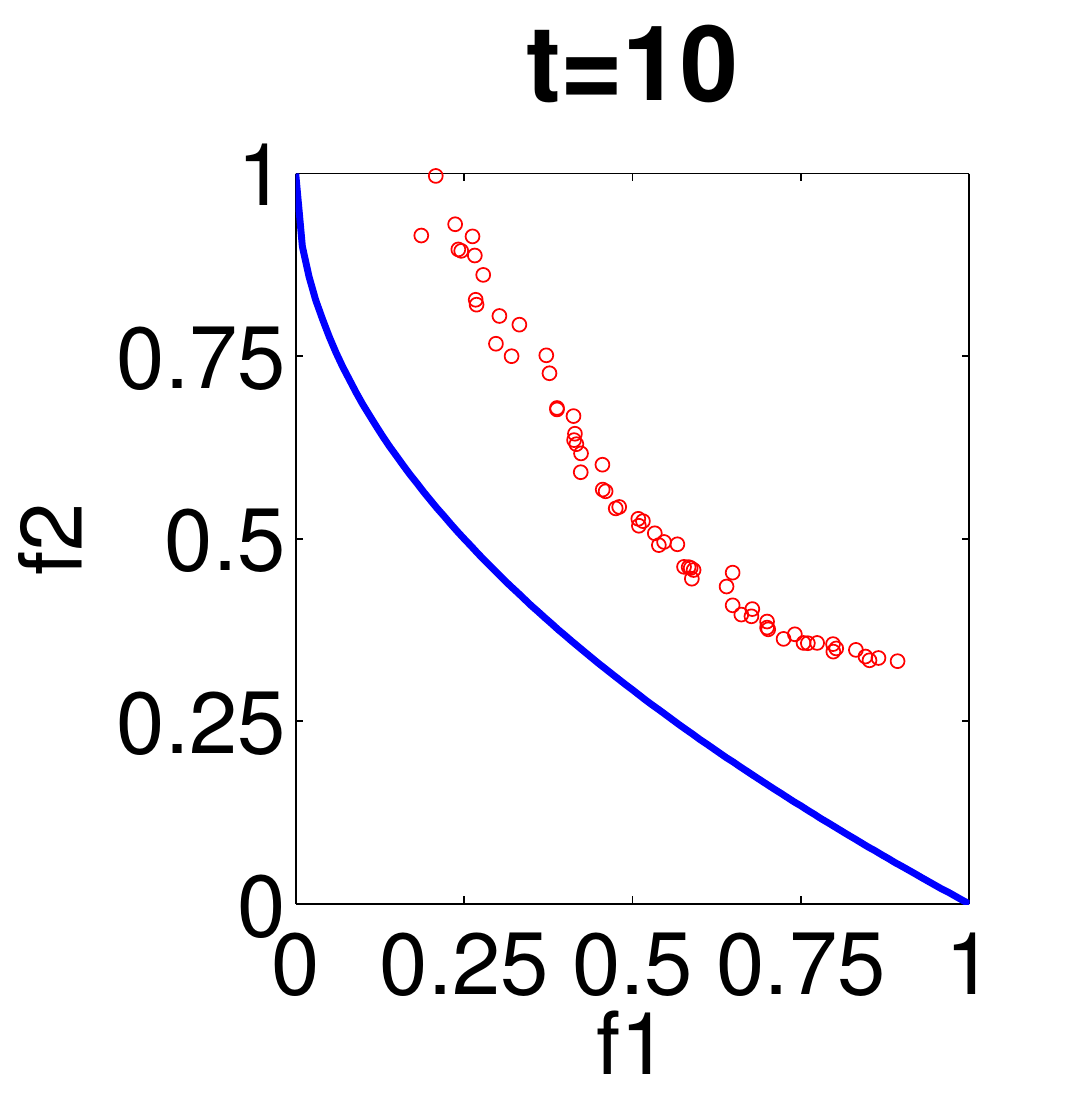}
 			\includegraphics[scale=0.2]{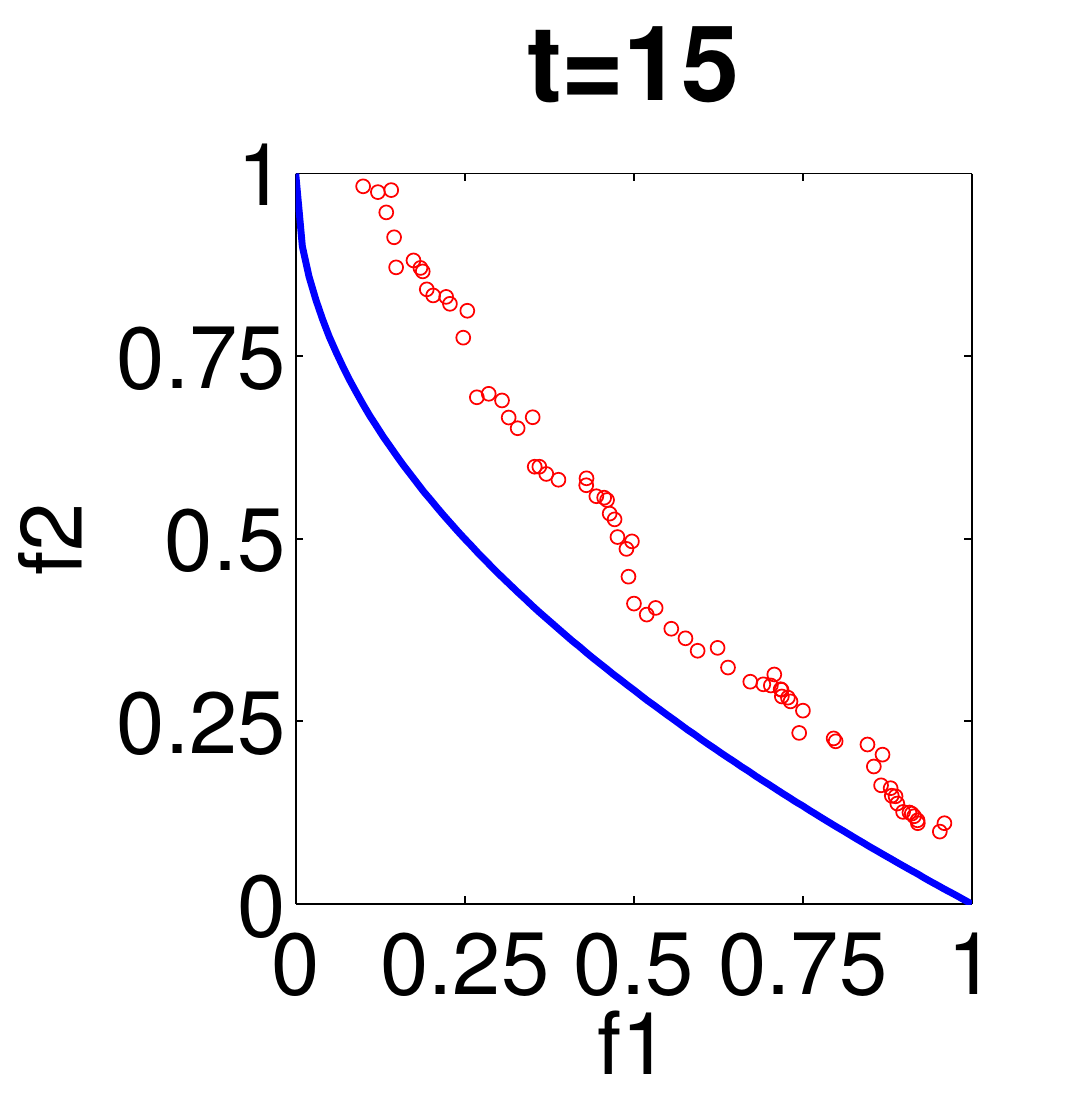}
 			\includegraphics[scale=0.2]{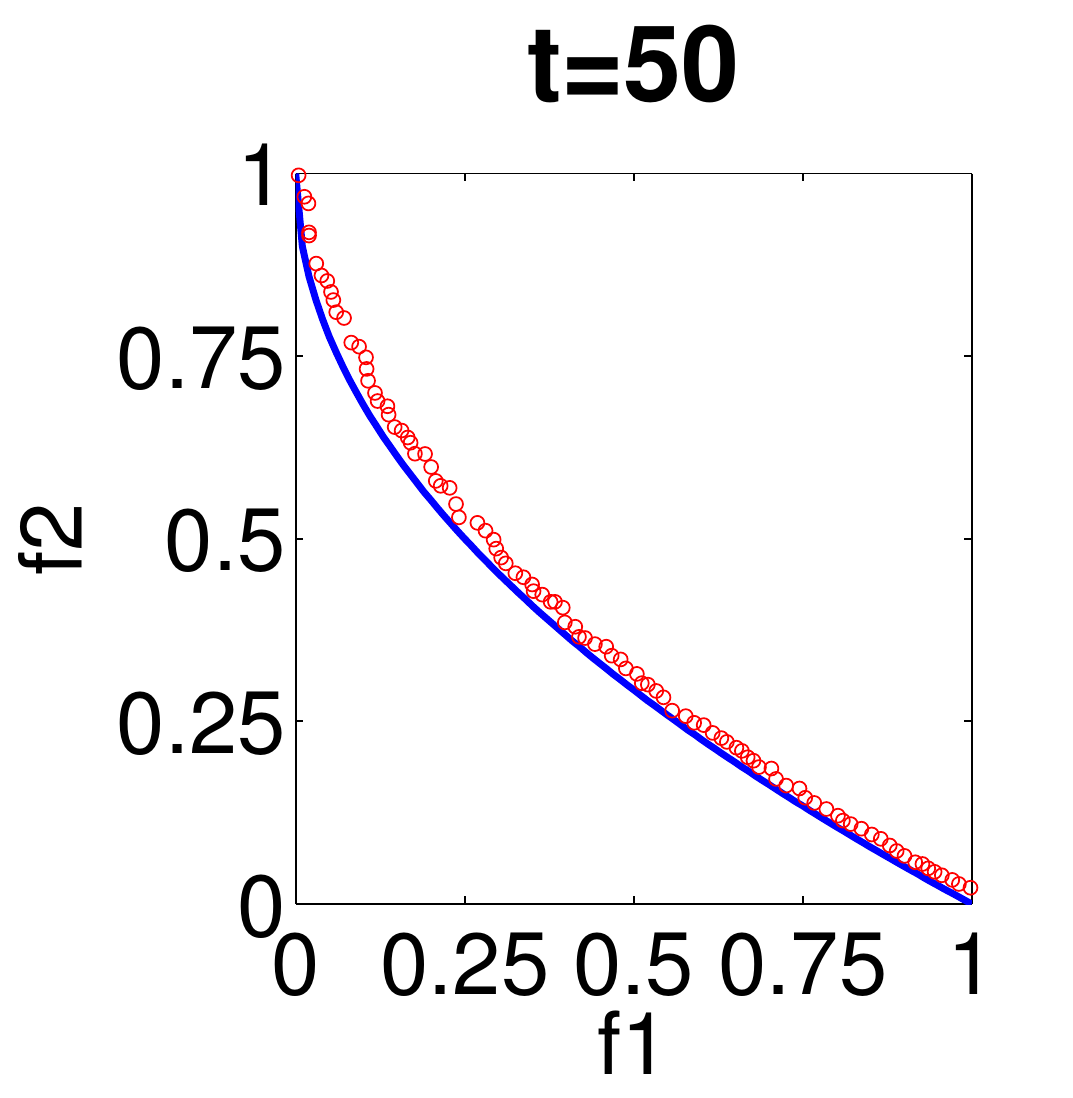}
 			\includegraphics[scale=0.2]{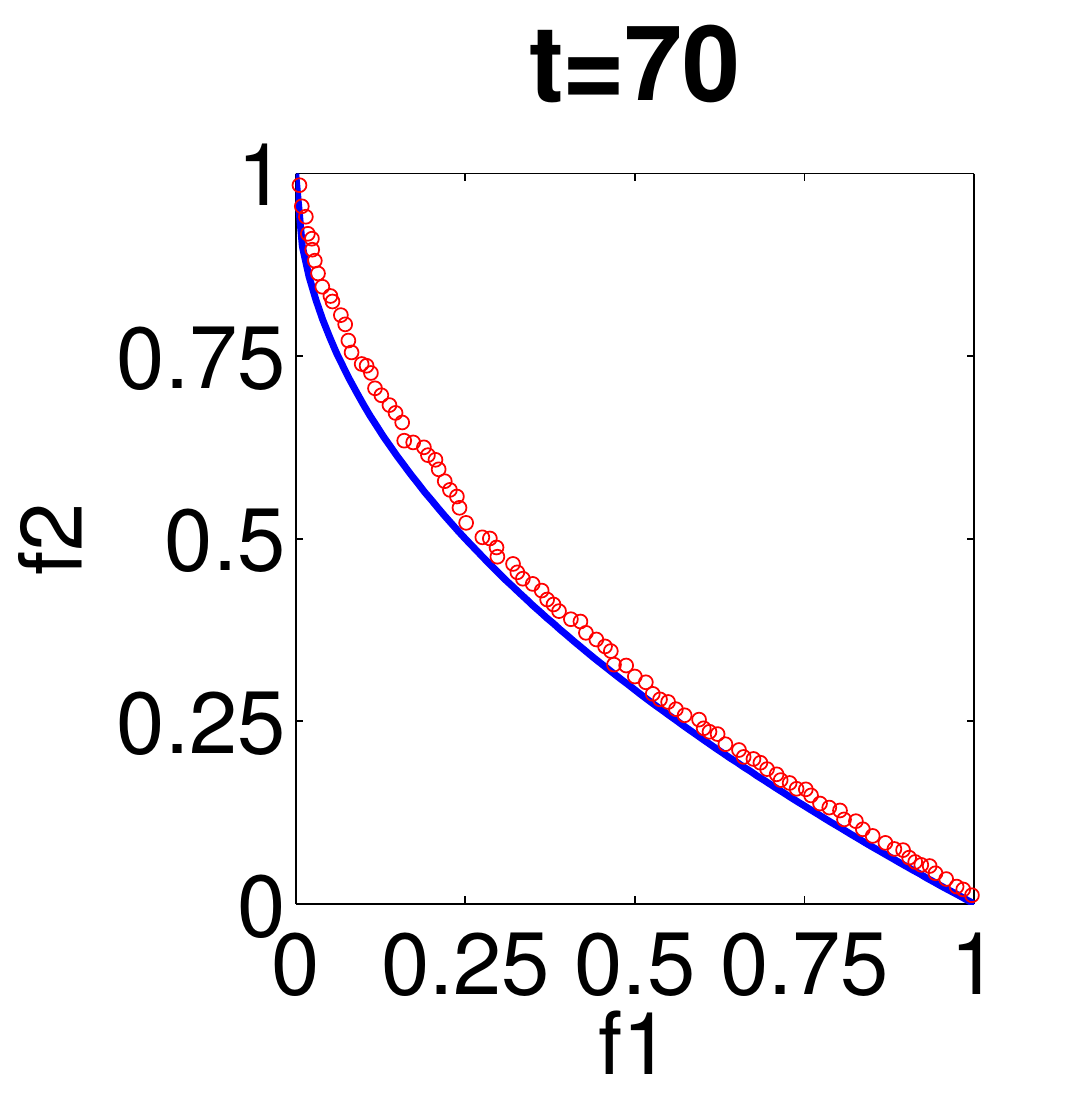}
 			\includegraphics[scale=0.2]{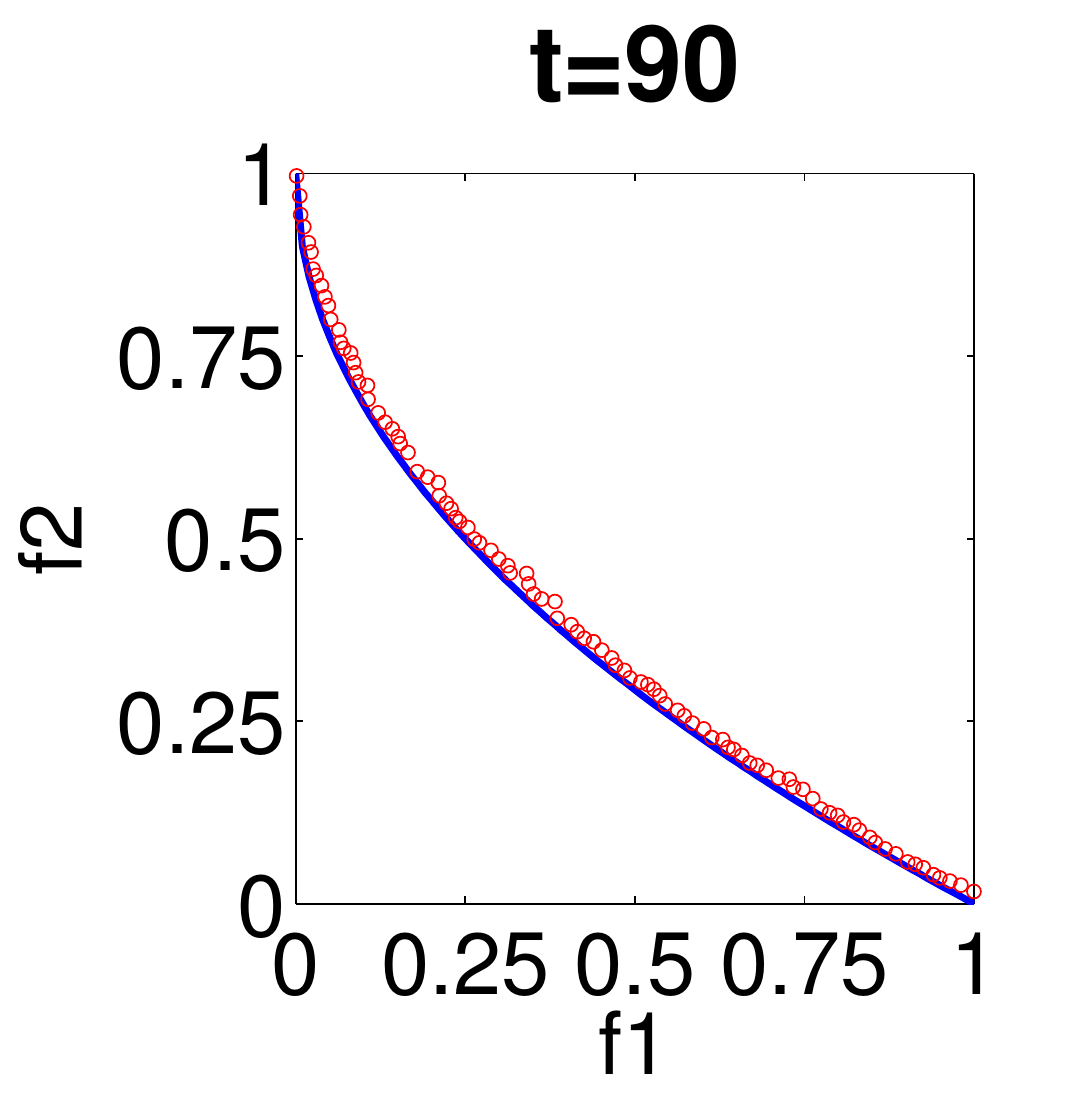}\\
 			\textbf{(b) FPS}\\
 			\includegraphics[scale=0.2]{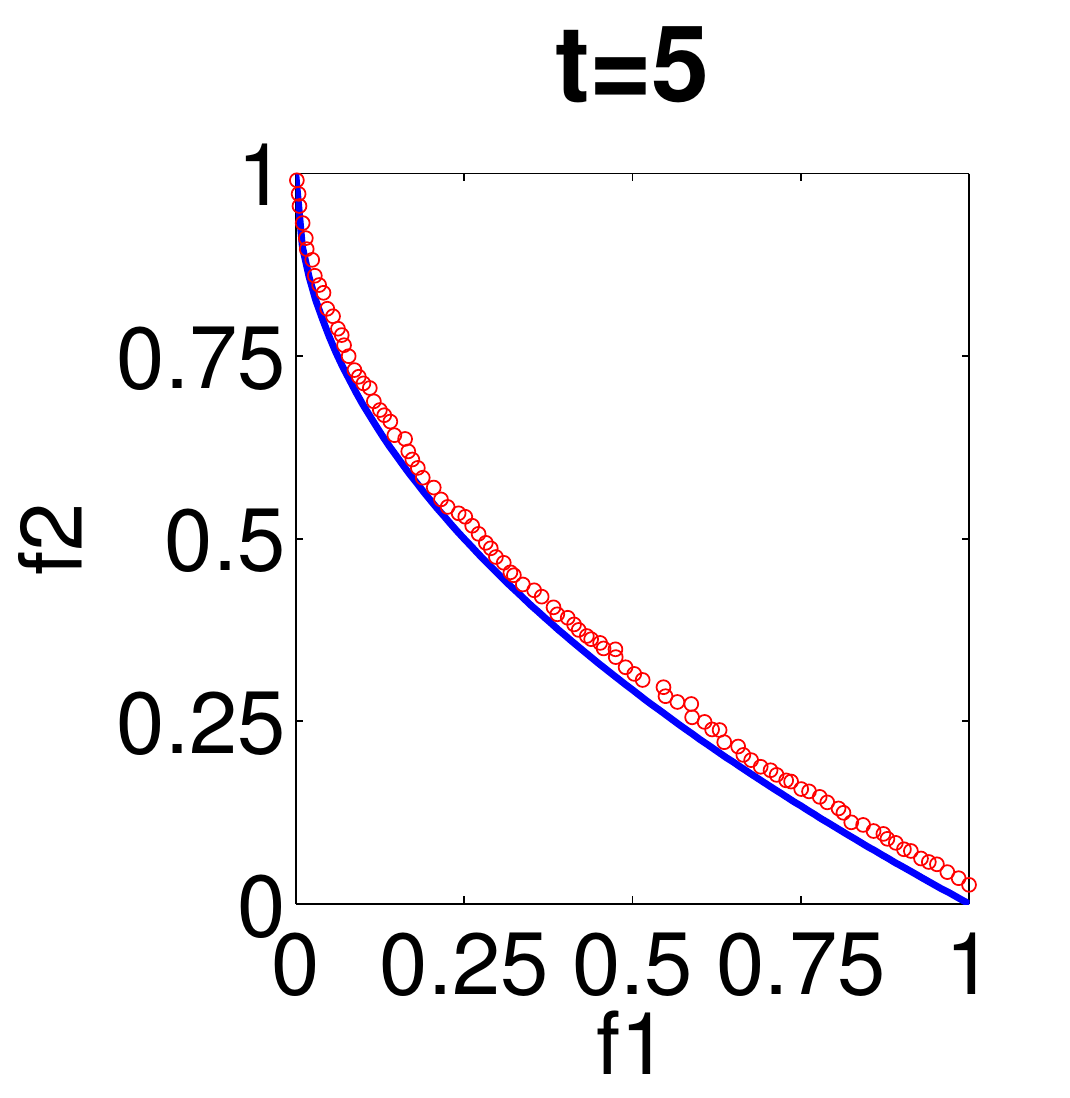}
 			\includegraphics[scale=0.2]{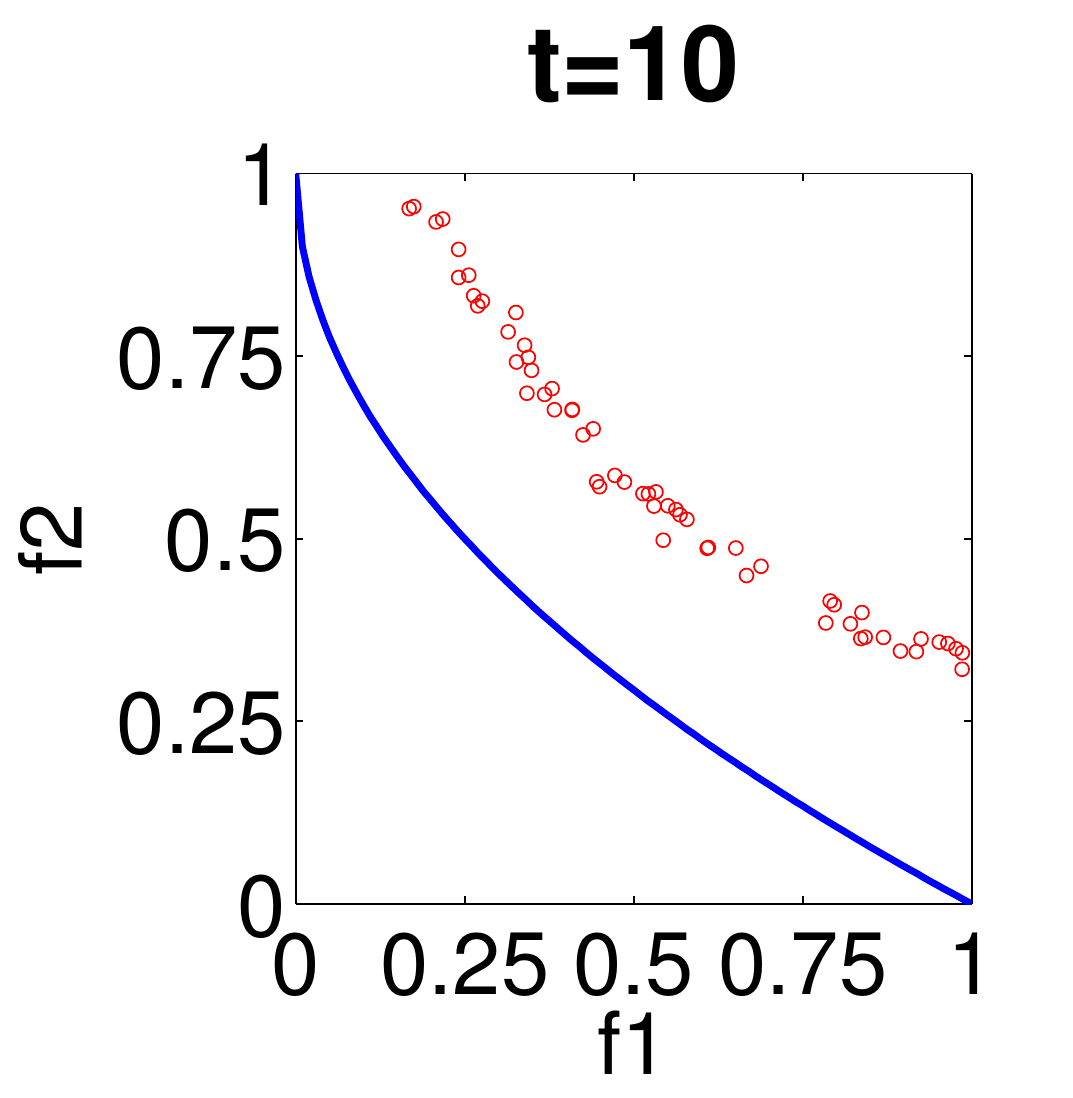}
 			\includegraphics[scale=0.2]{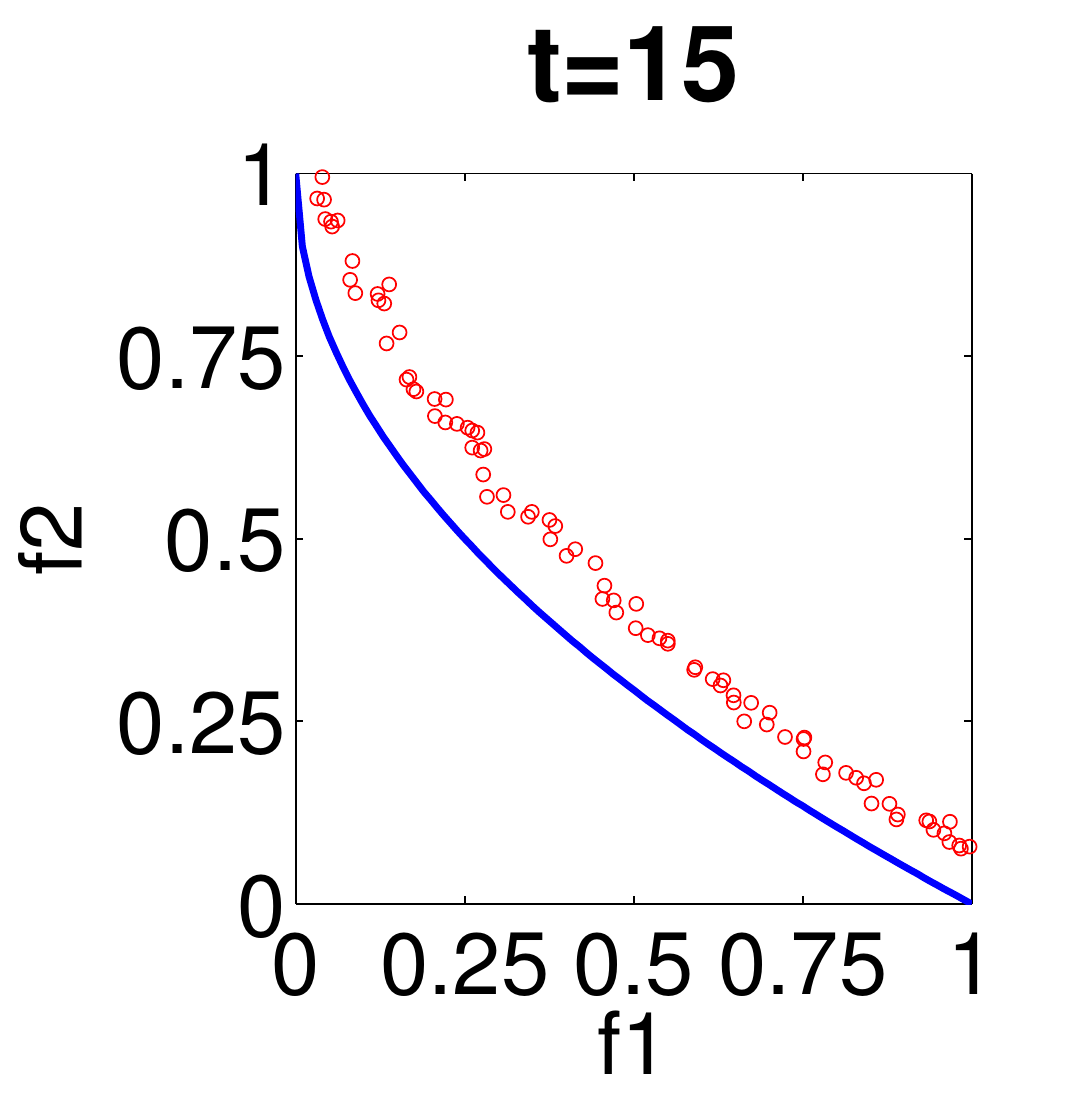}
 			\includegraphics[scale=0.2]{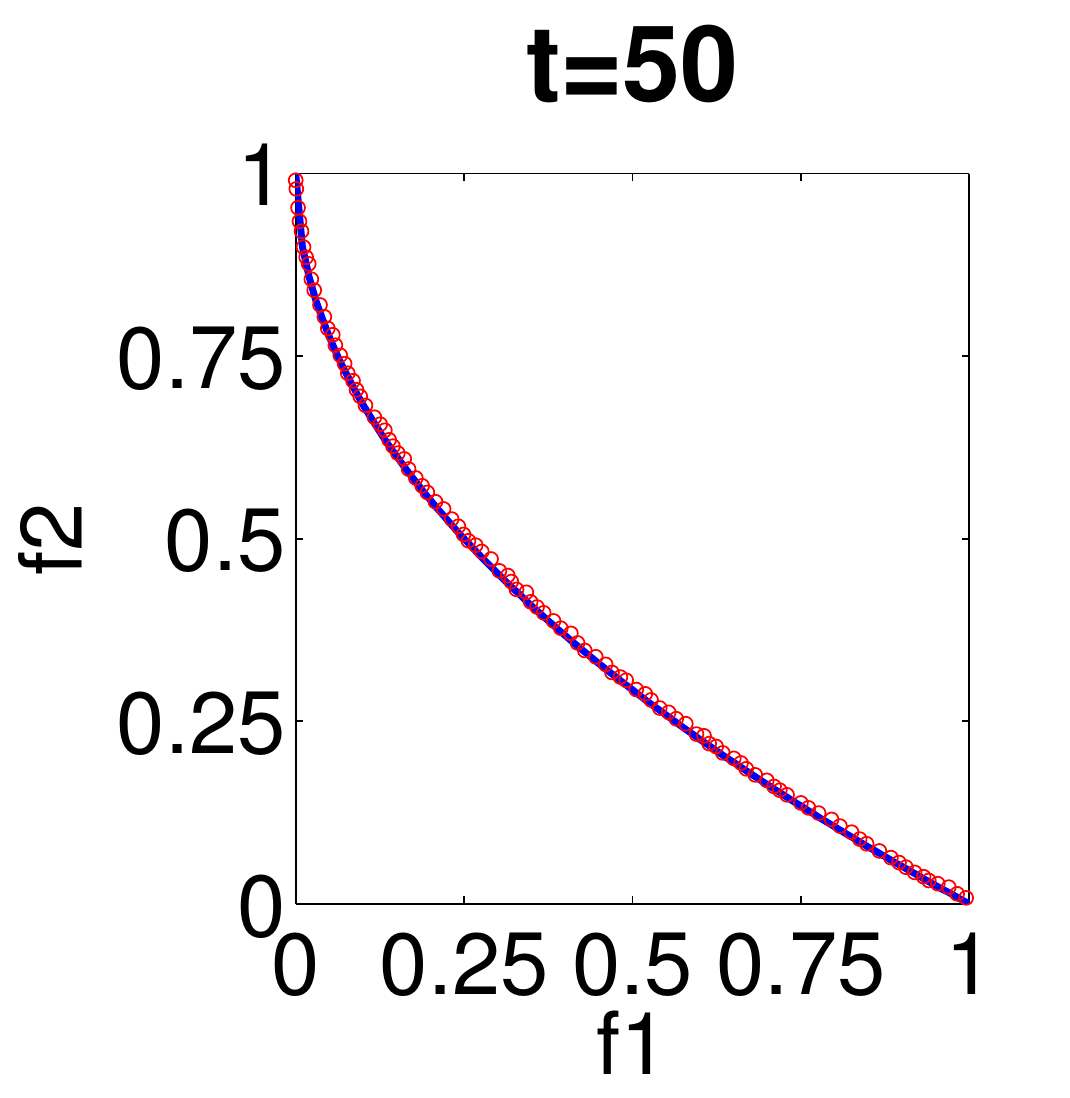}
 			\includegraphics[scale=0.2]{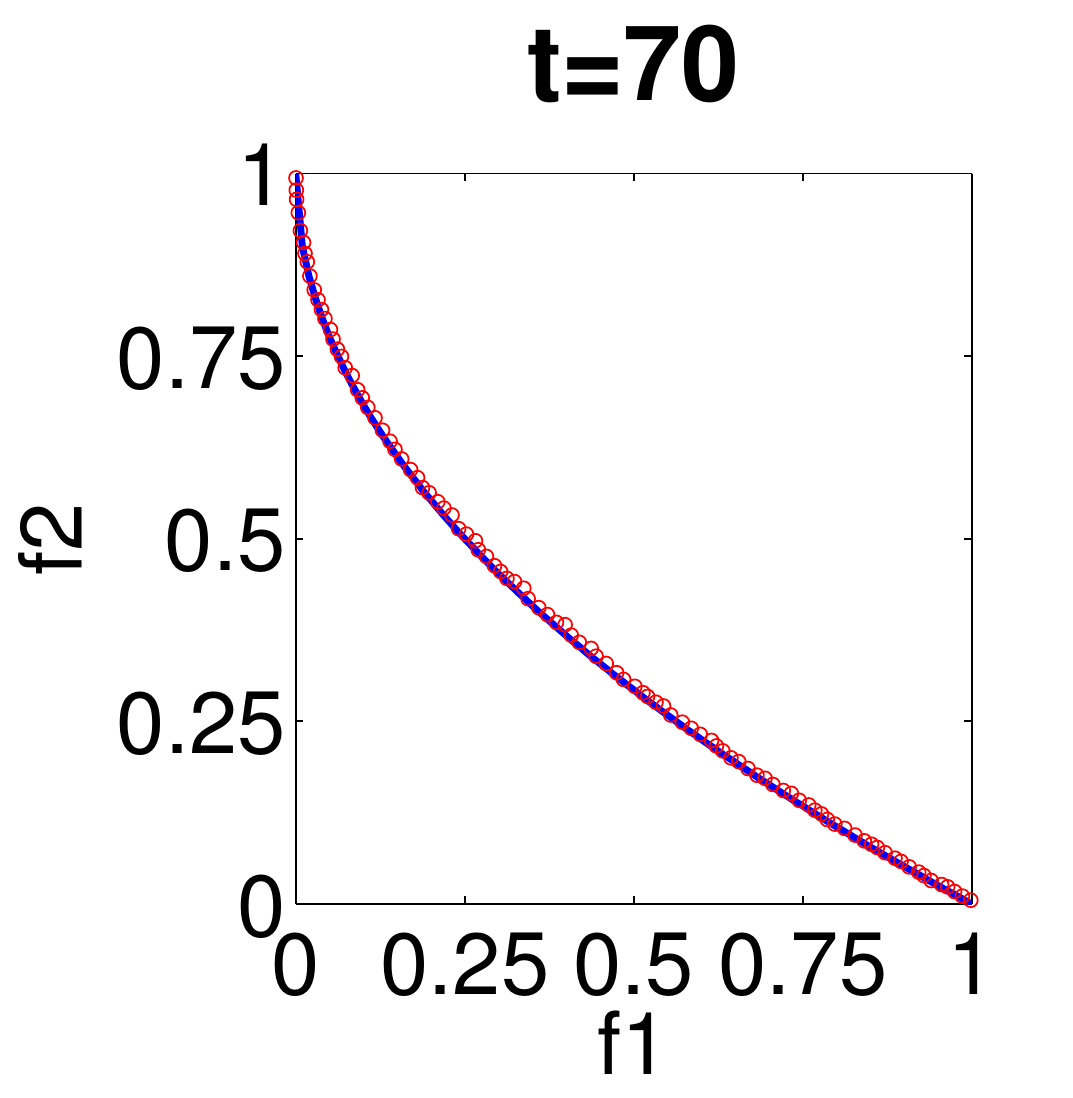}
 			\includegraphics[scale=0.2]{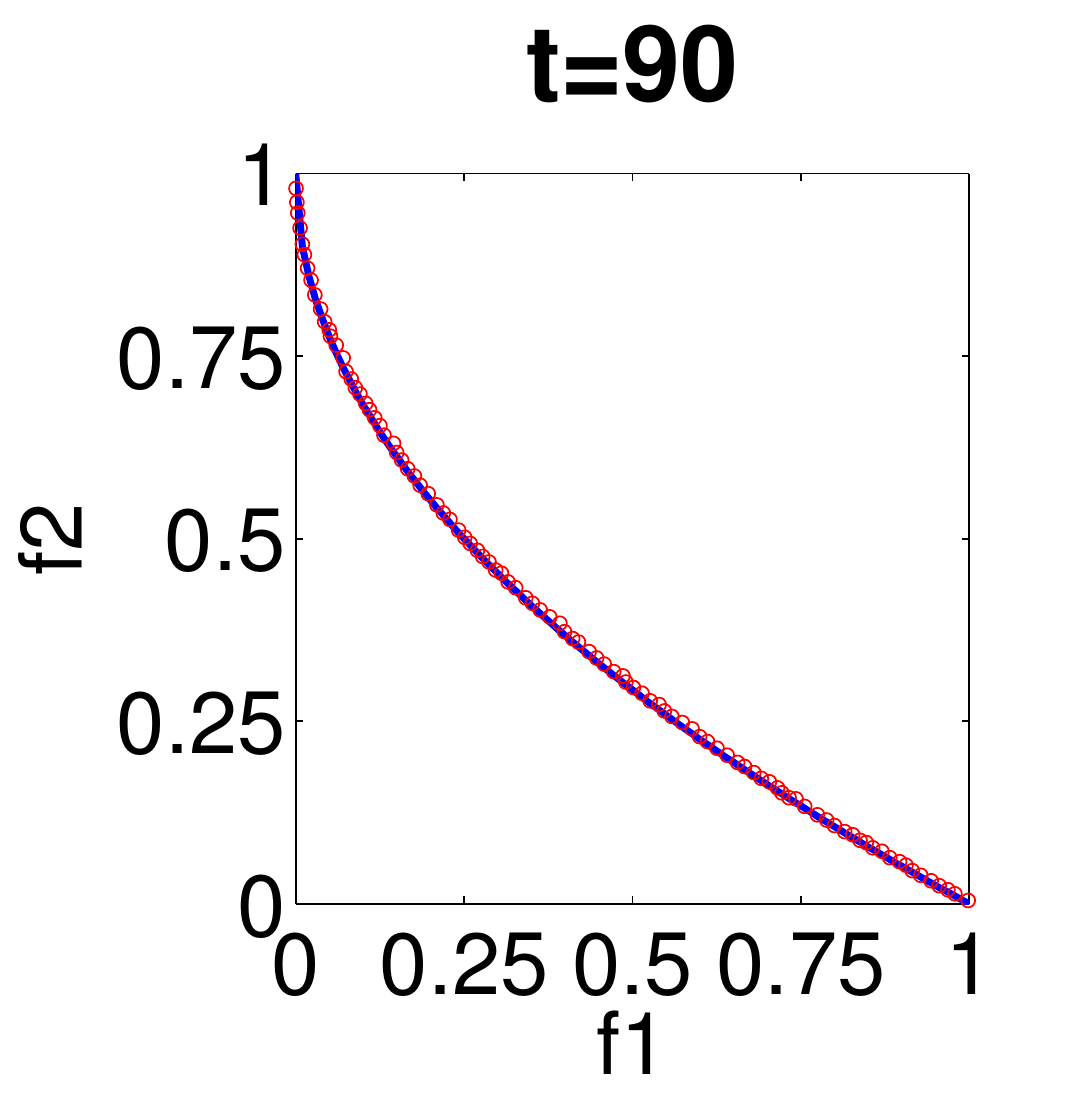}\\
 			\textbf{	(c) PPS}\\
 			\includegraphics[scale=0.2]{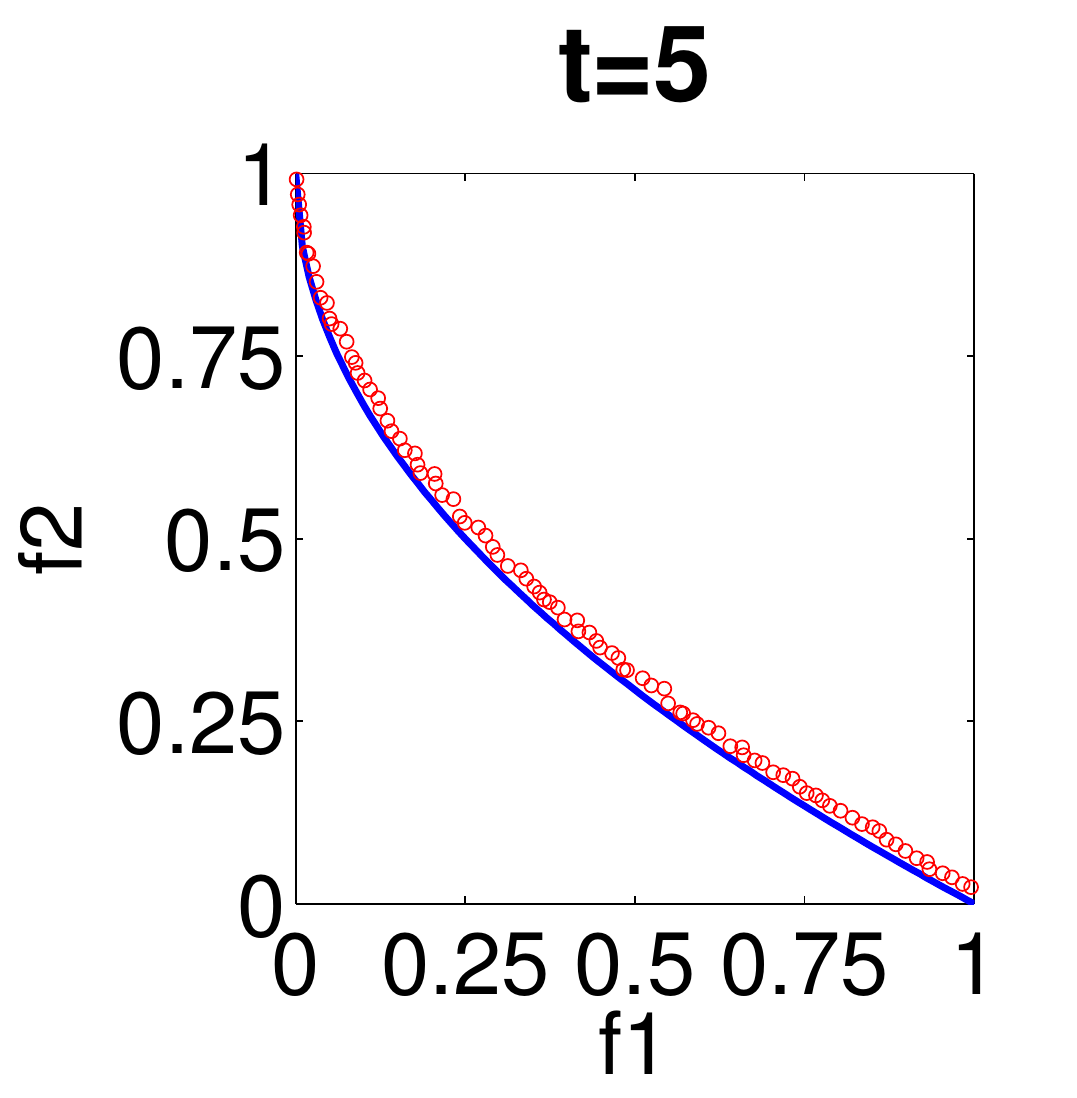}
 			\includegraphics[scale=0.2]{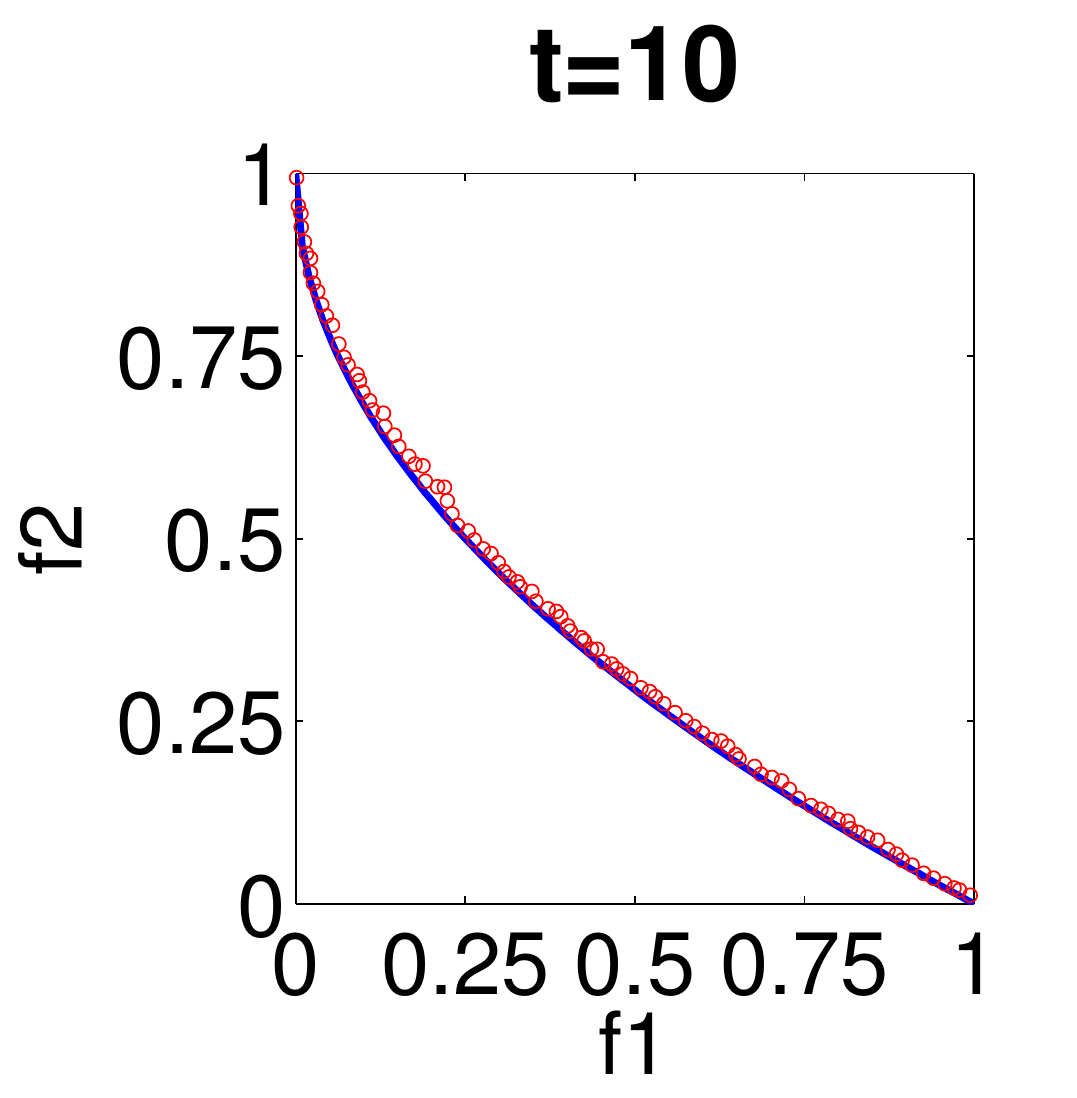}
 			\includegraphics[scale=0.2]{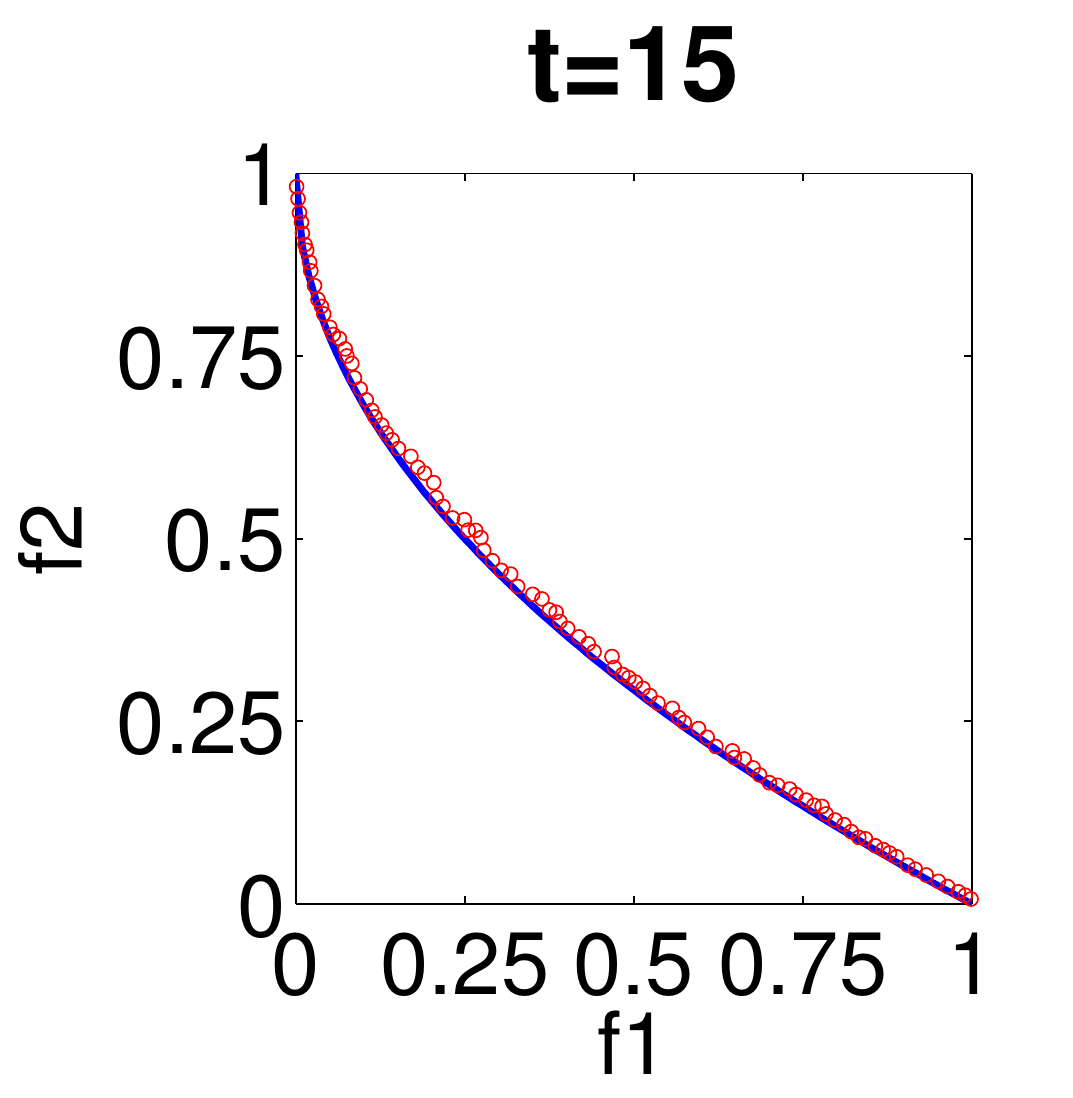}
 			\includegraphics[scale=0.2]{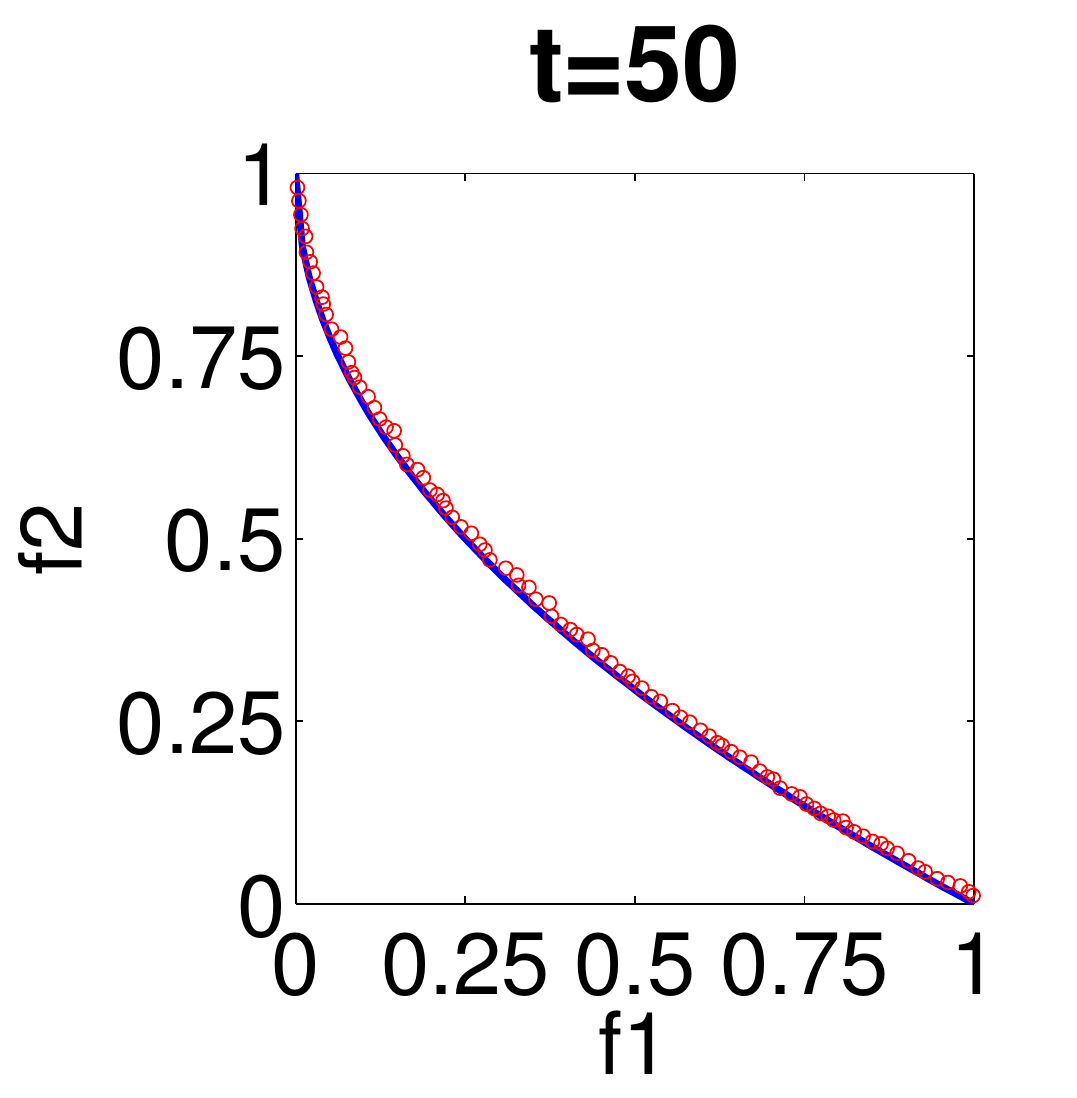}
 			\includegraphics[scale=0.2]{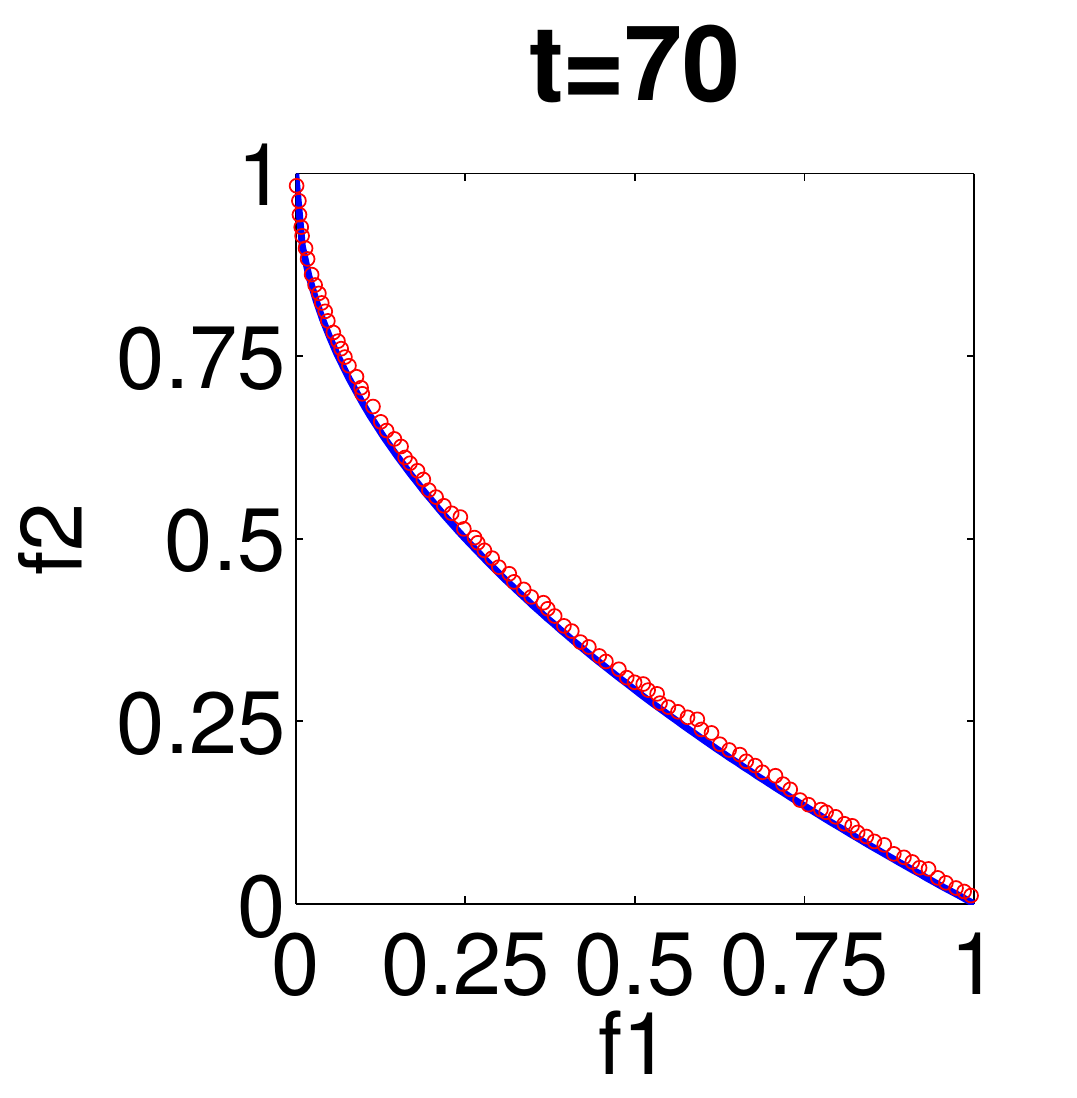}
 			\includegraphics[scale=0.2]{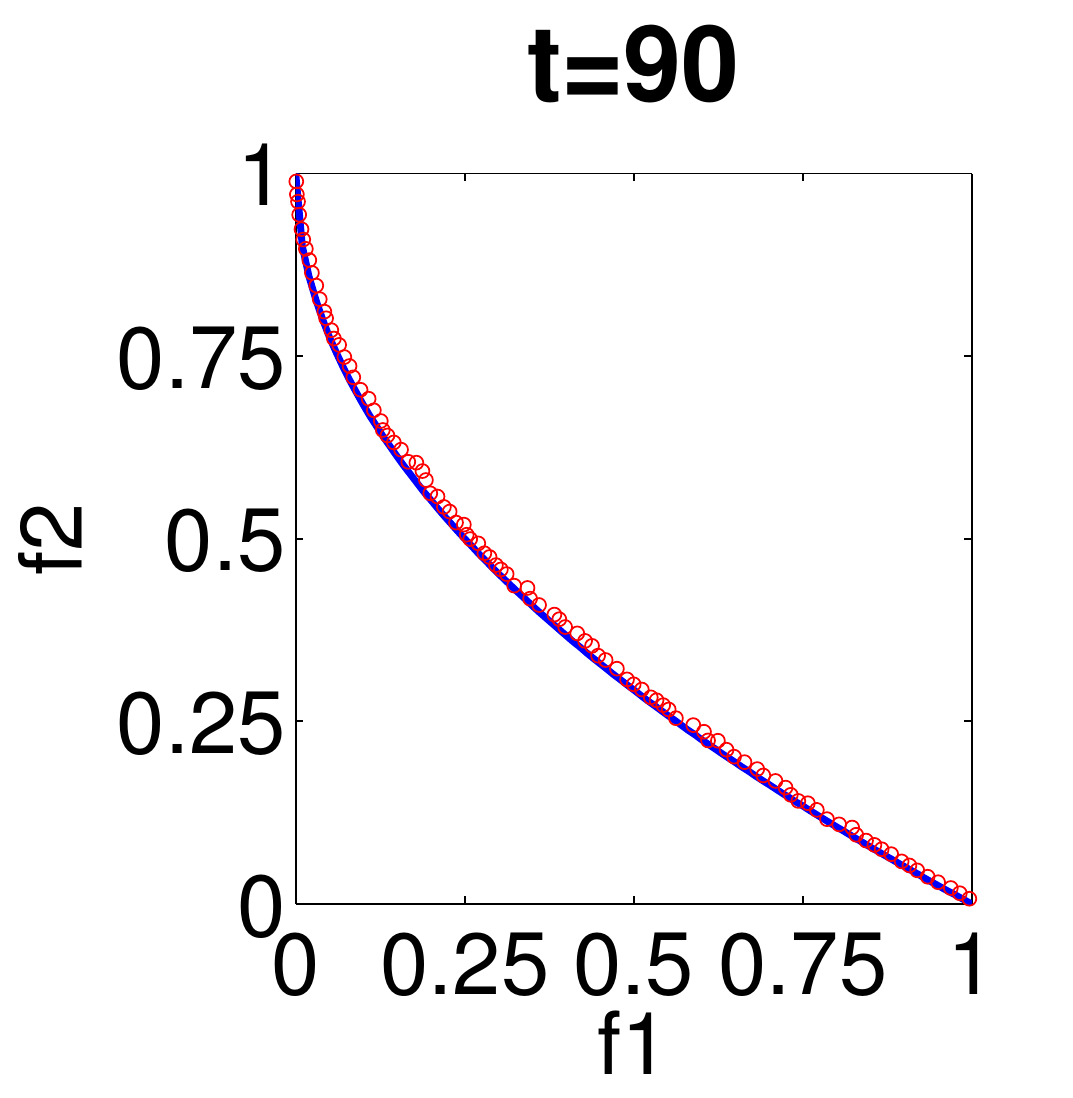}\\
 			\textbf{	(d) SPPS}\\
 			\includegraphics[scale=0.2]{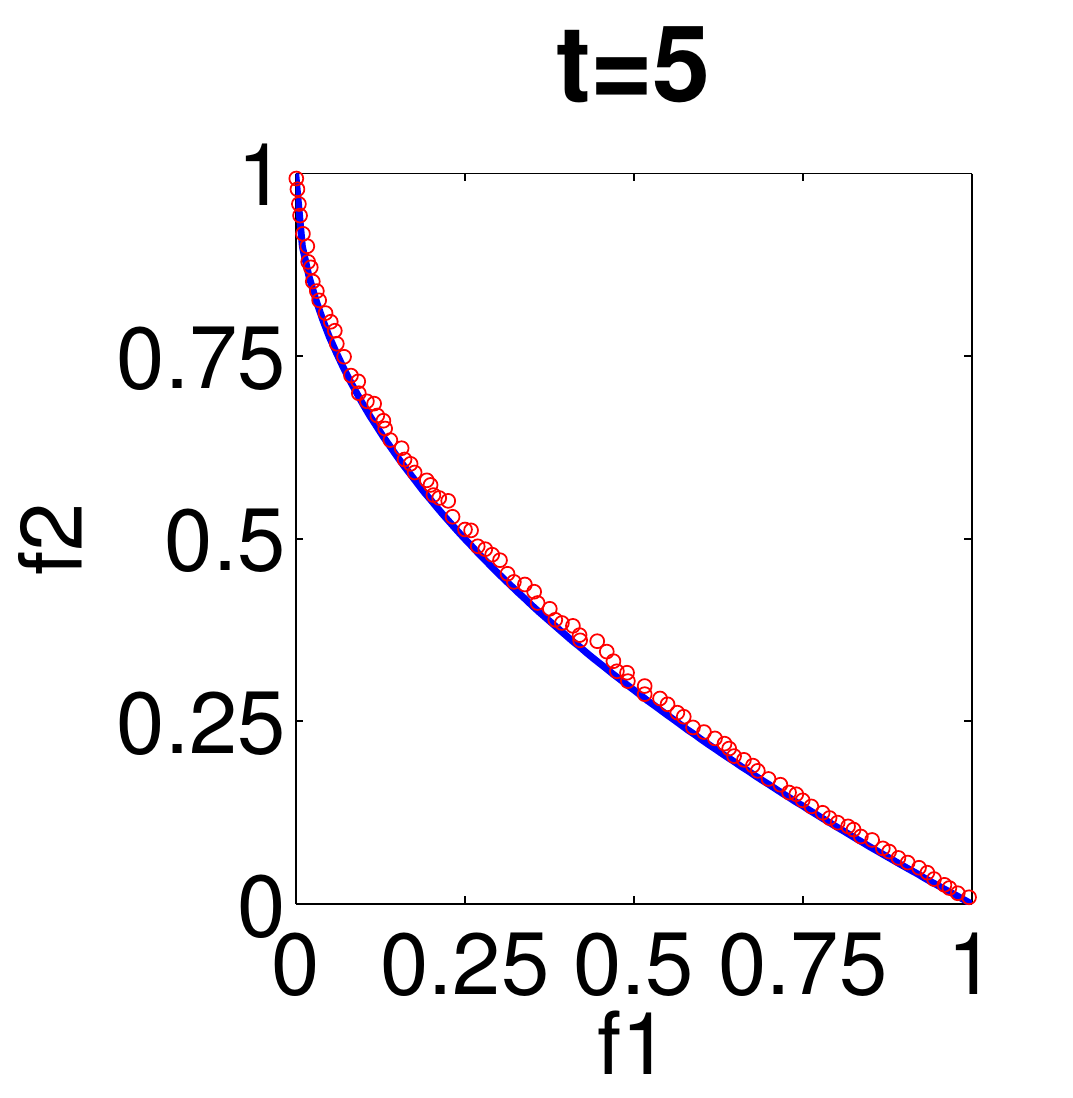}
 			\includegraphics[scale=0.2]{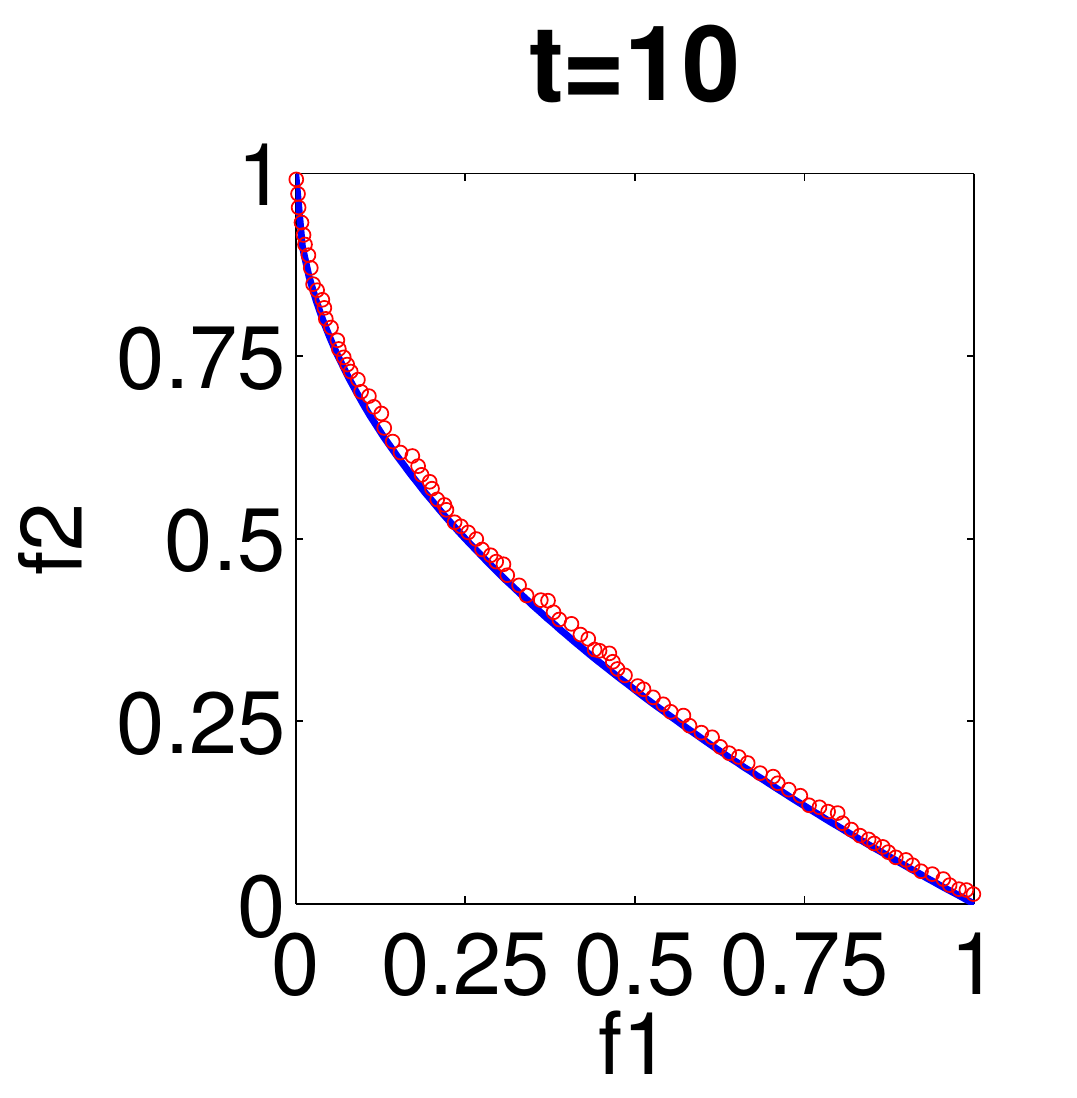}
 			\includegraphics[scale=0.2]{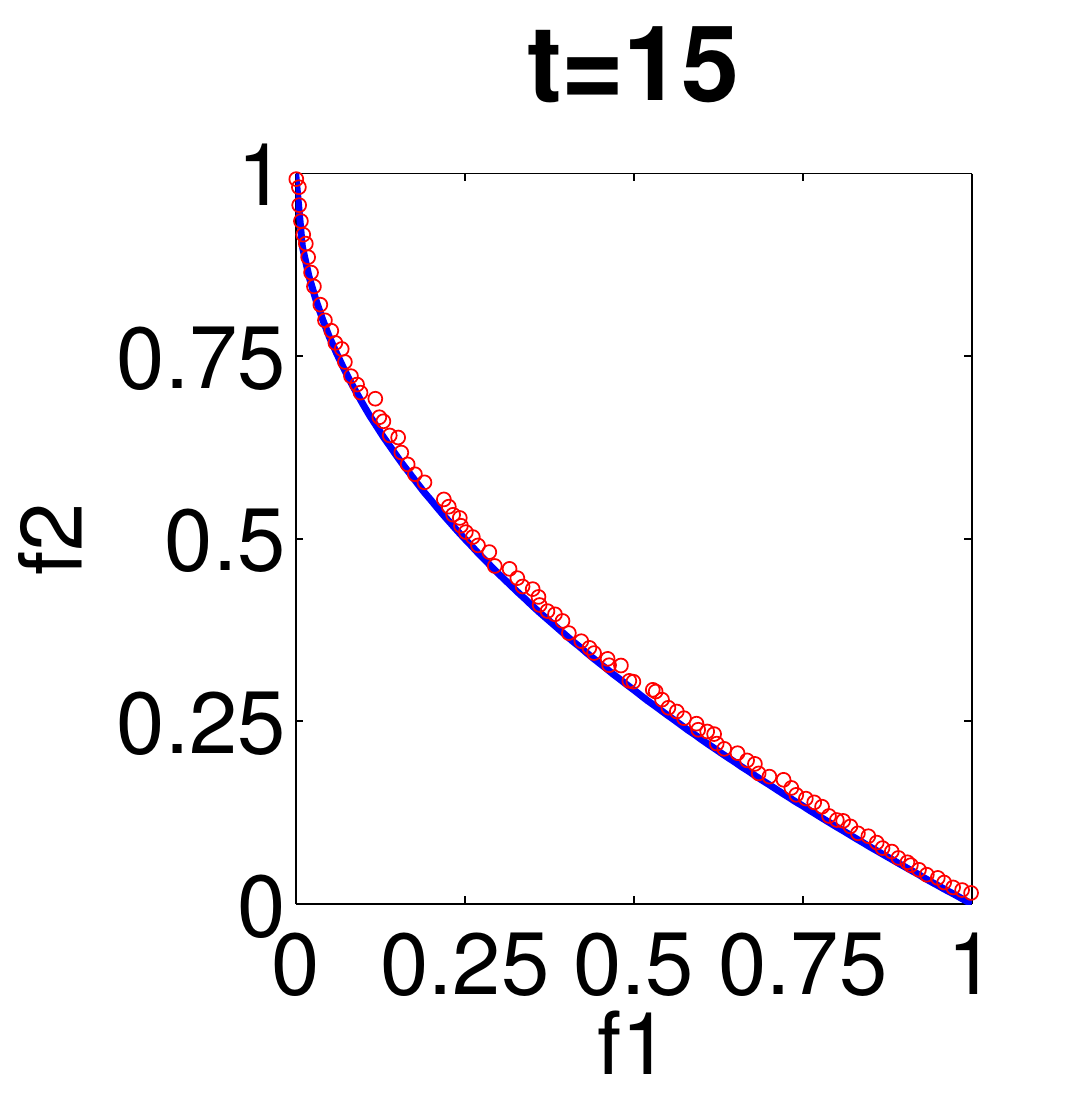}
 			\includegraphics[scale=0.2]{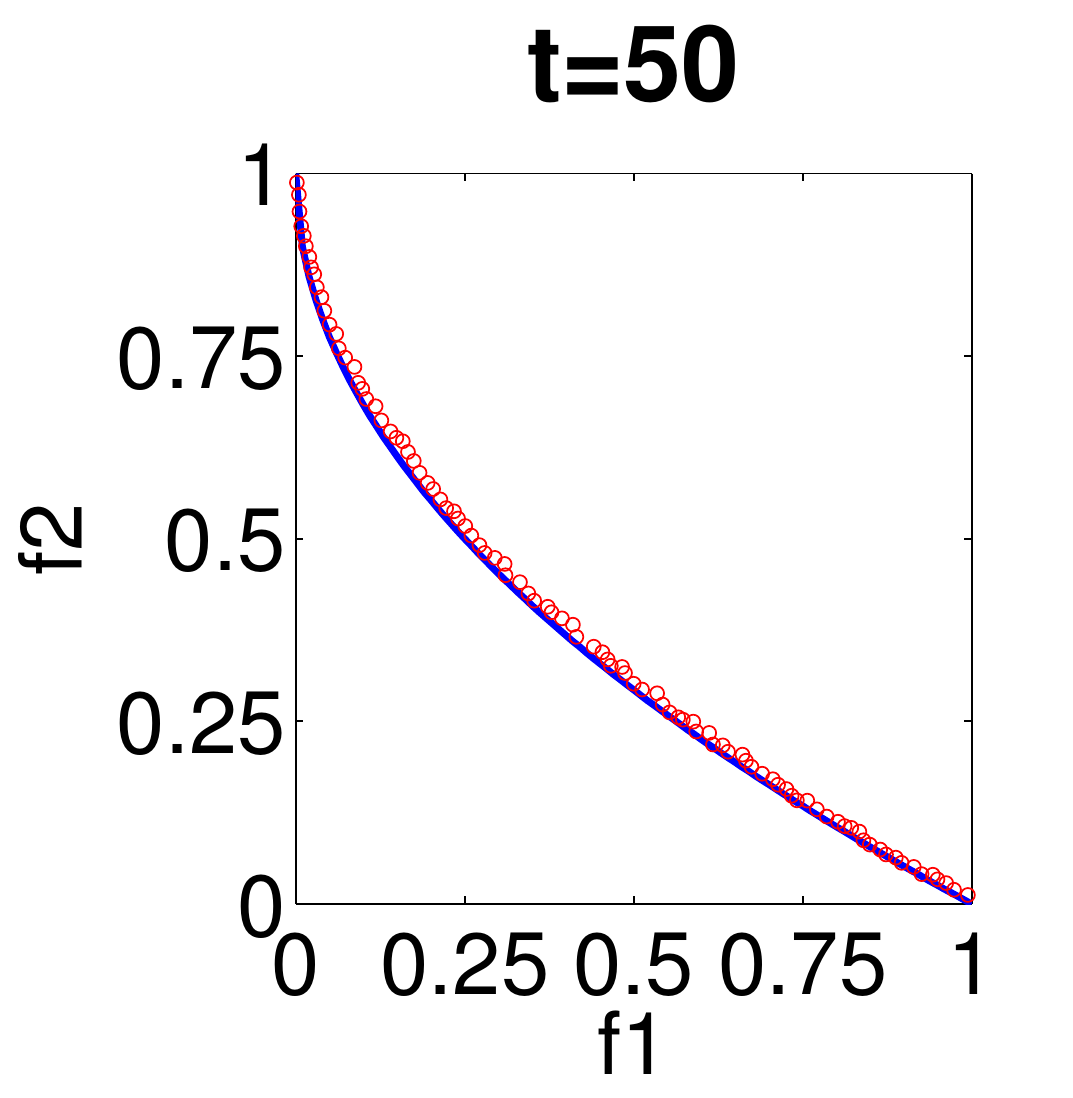}
 			\includegraphics[scale=0.2]{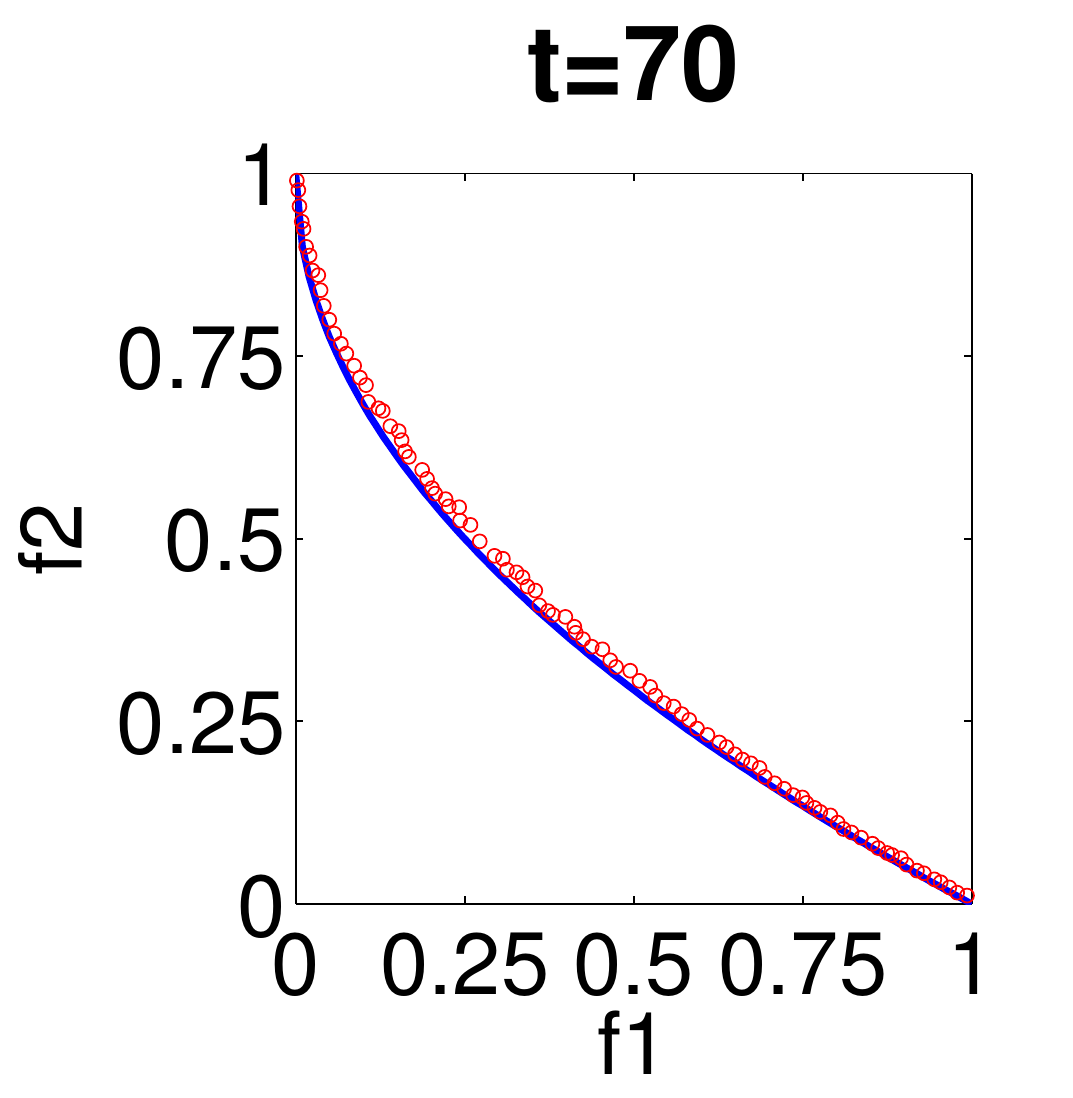}
 			\includegraphics[scale=0.2]{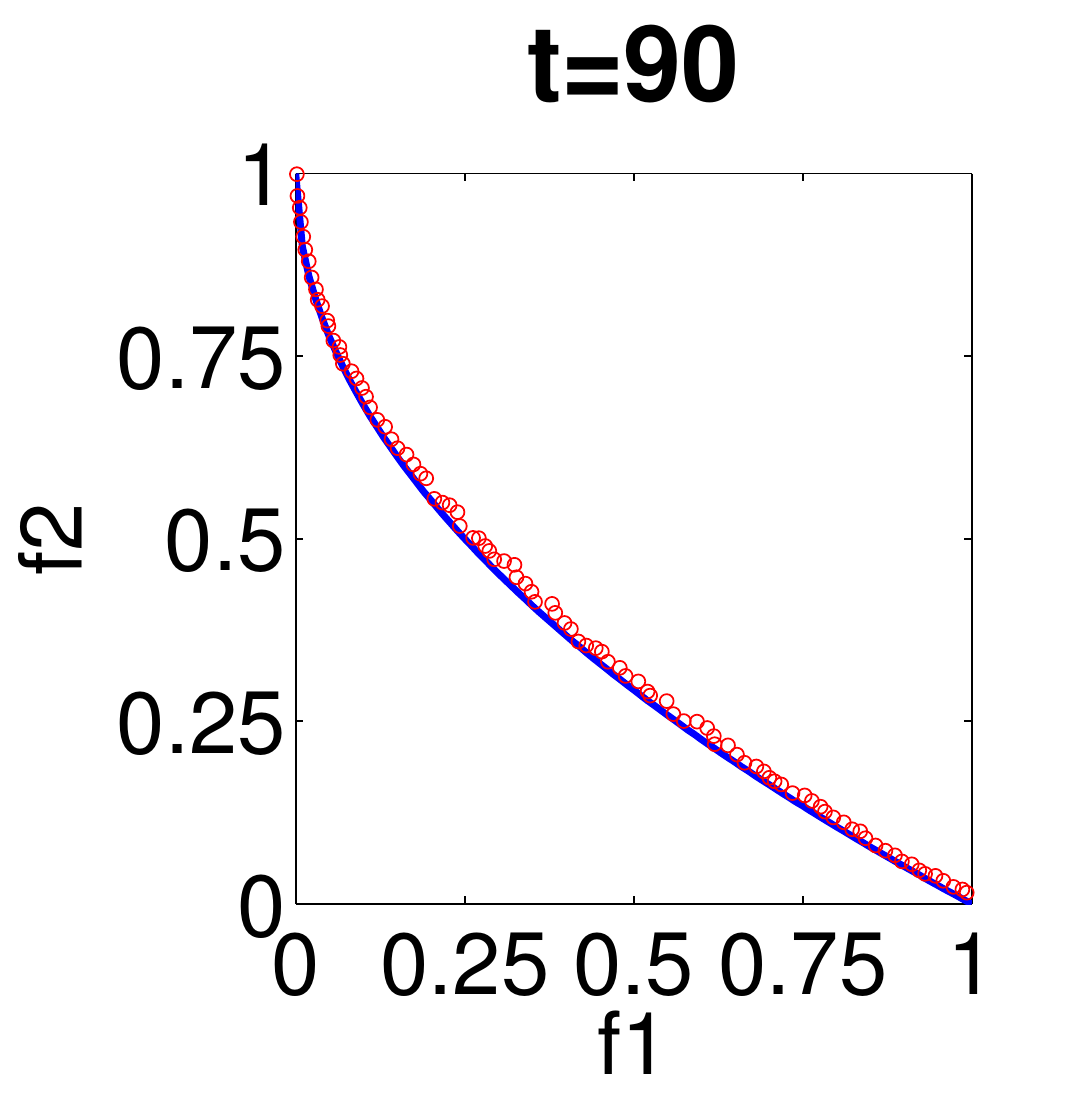}\\
 			\textbf{(e) FGERS-CPS}\\
 		\end{tabular}
 	\end{center}
 	\caption{Final population distribution of the five strategies at six time steps on FDA1.}
 	\label{fig:5}
 \end{figure*}
 
 \begin{figure}[htbp]
 	\centering
 	\subfigure[\textbf{ RIS}]{
 		\begin{minipage}[t]{0.2\linewidth}
 			\centering
 			\includegraphics[width=1.1in]{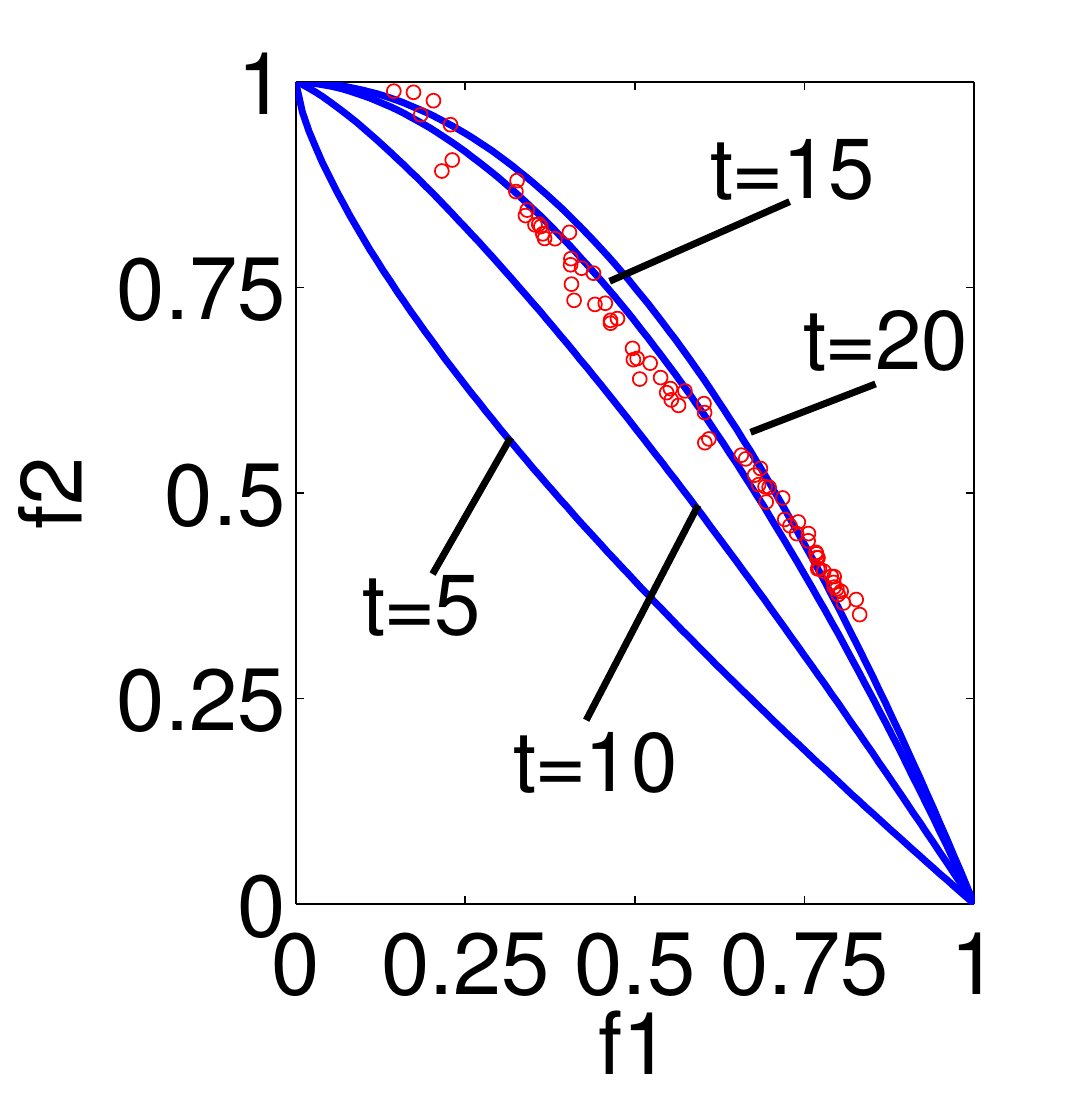}
 		\end{minipage}%
 	}%
 	\subfigure[ \textbf{FPS}]{
 		\begin{minipage}[t]{0.2\linewidth}
 			\centering
 			\includegraphics[width=1.1in]{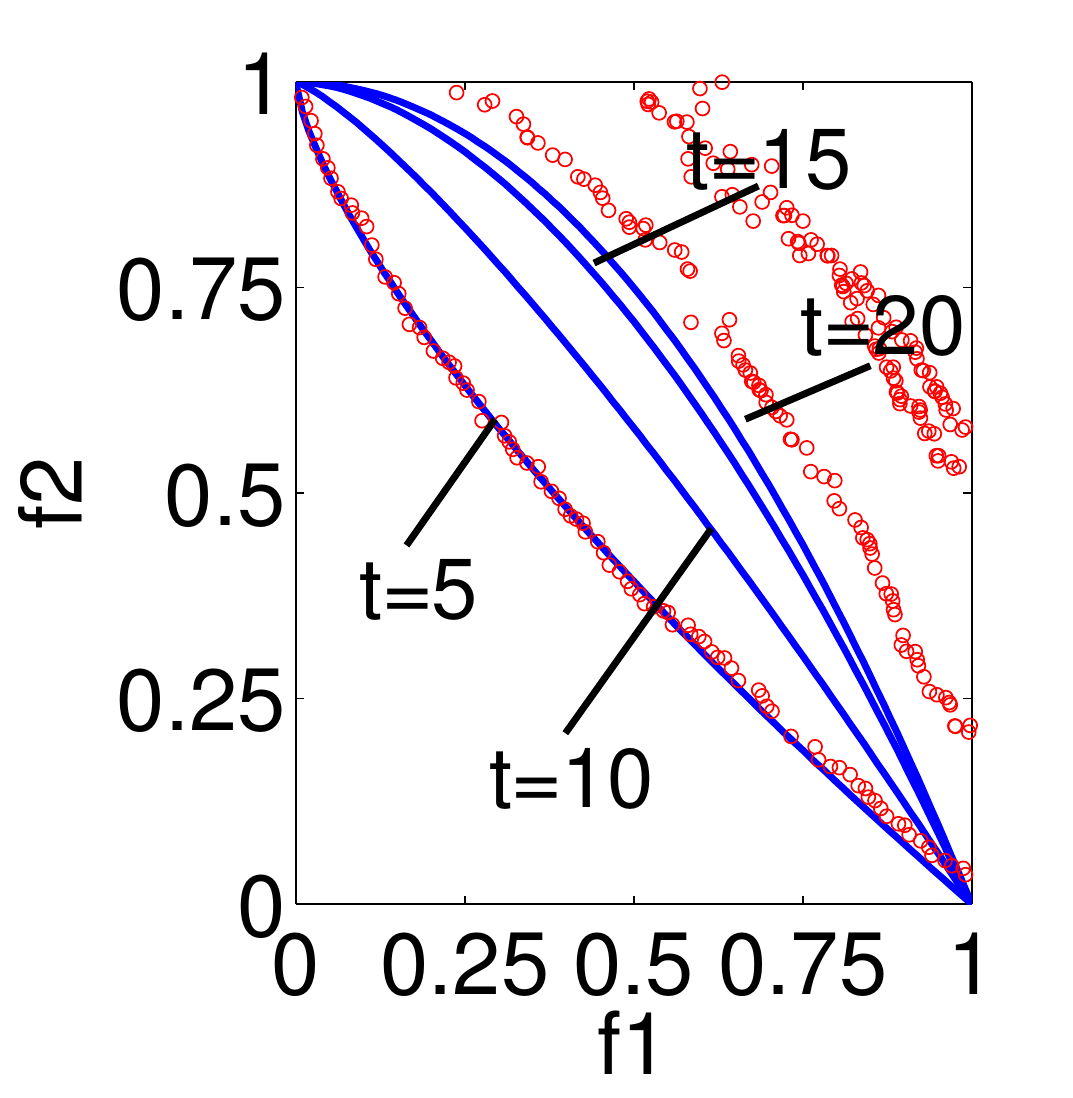}
 		\end{minipage}%
 	}%
 	\subfigure[\textbf{PPS}]{
 		\begin{minipage}[t]{0.2\linewidth}
 			\centering
 			\includegraphics[width=1.1in]{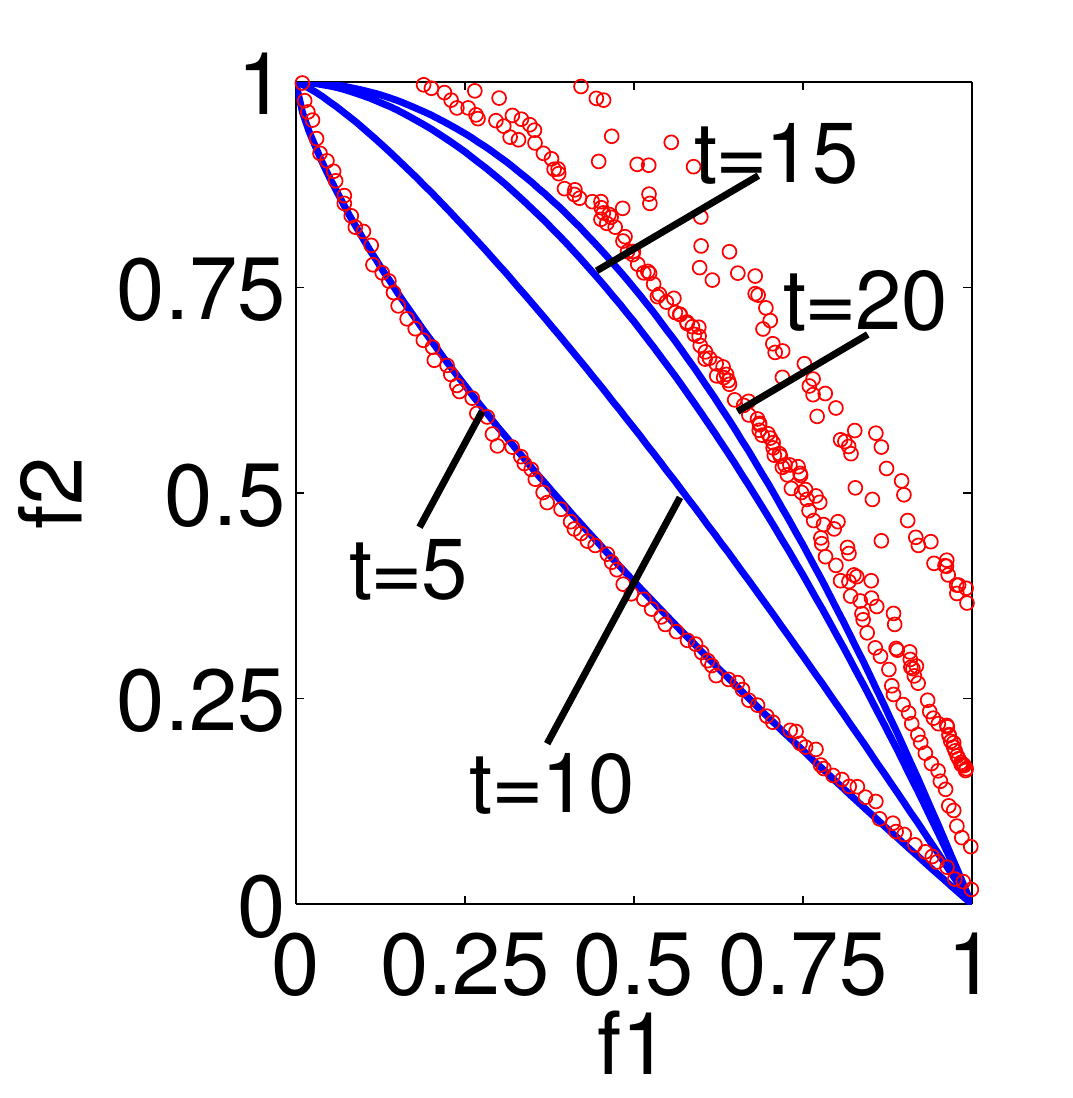}
 		\end{minipage}
 	}%
 	\subfigure[\textbf{SPPS}]{
 		\begin{minipage}[t]{0.2\linewidth}
 			\centering
 			\includegraphics[width=1.1in]{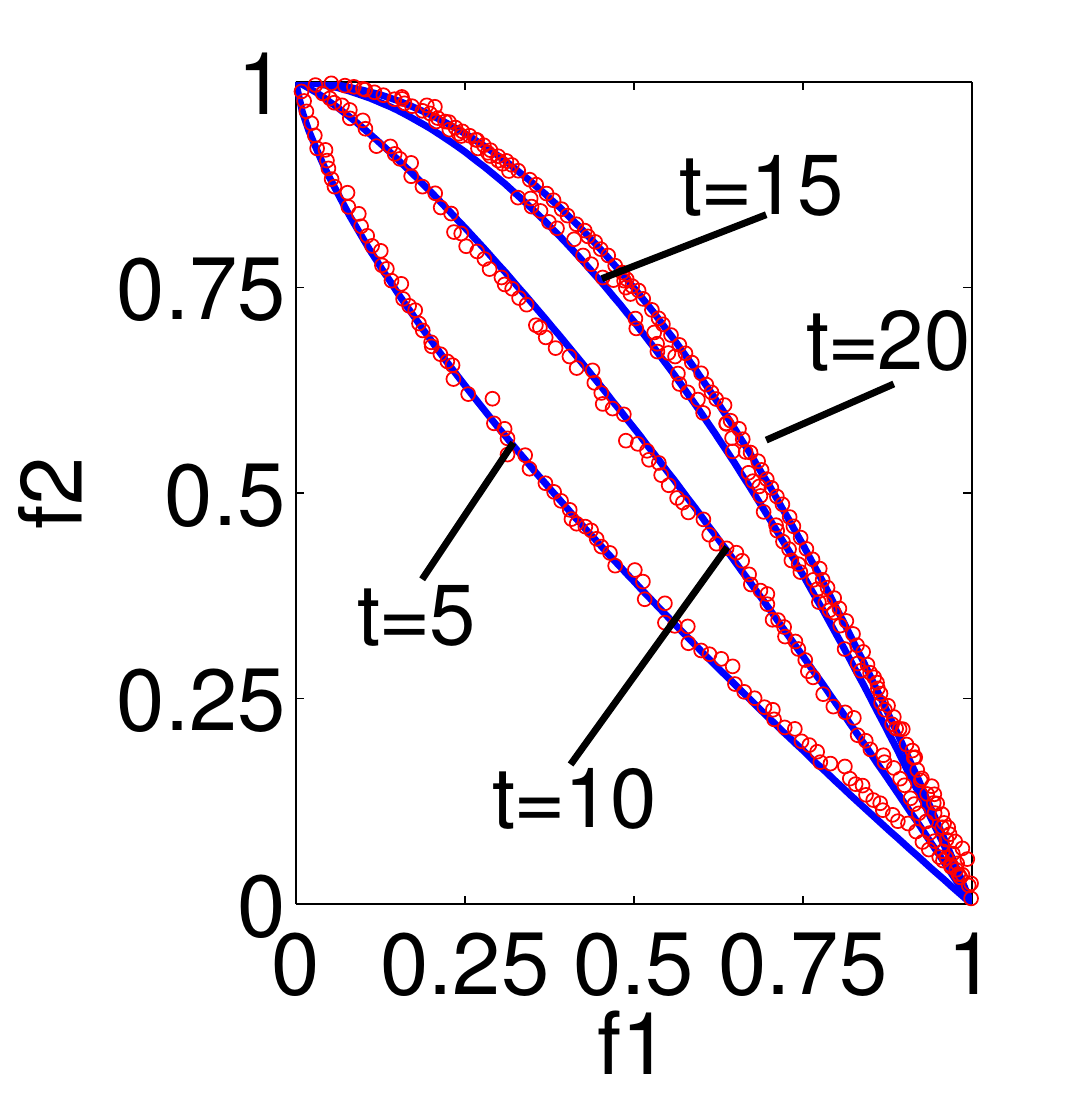}
 		\end{minipage}
 	}%
 	\subfigure[\textbf{FGERS-CPS}]{
 		\begin{minipage}[t]{0.2\linewidth}
 			\centering
 			\includegraphics[width=1.1in]{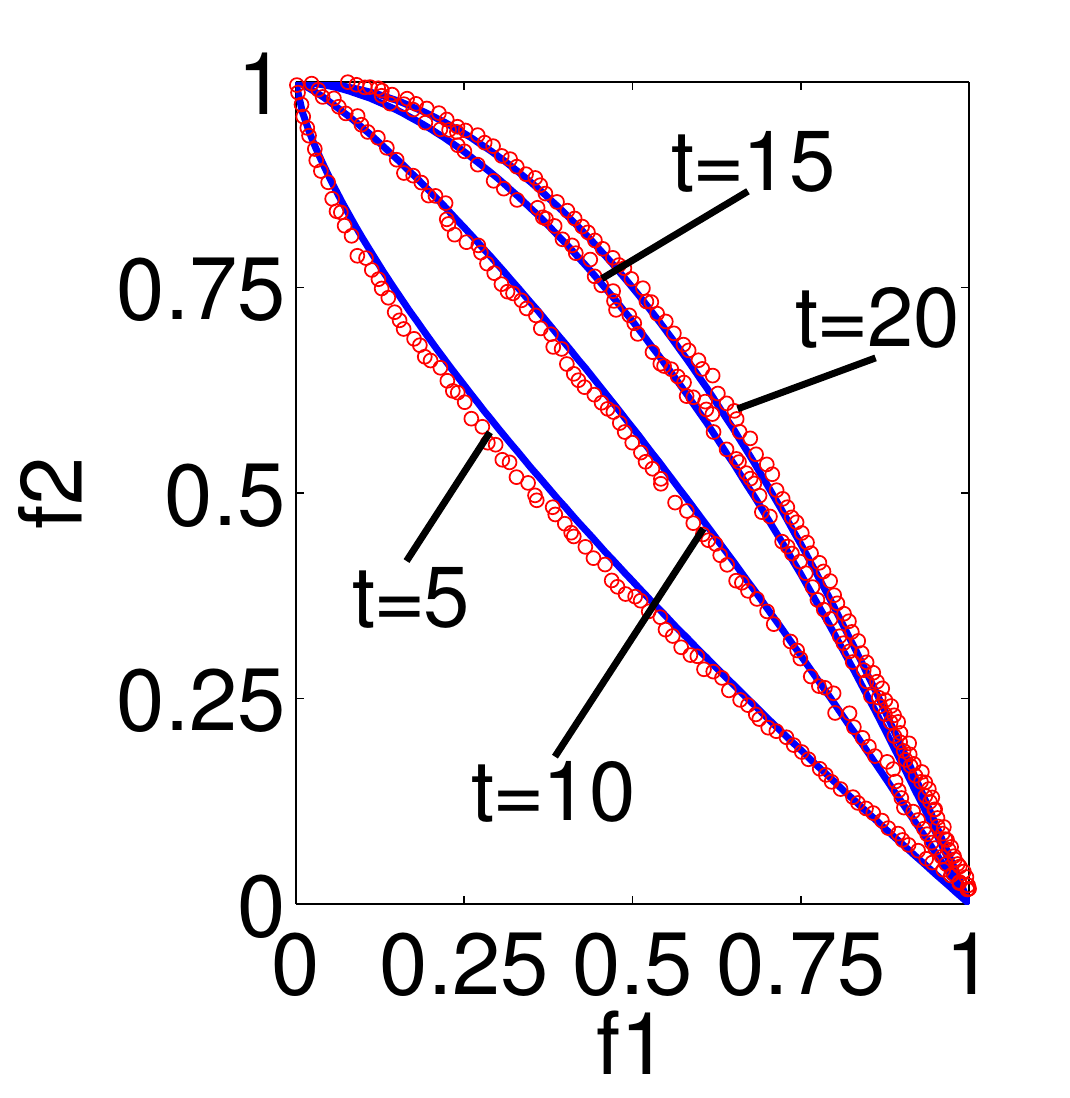}
 		\end{minipage}
 	}%
 	\\
 	\subfigure[\textbf{RIS}]{
 		\begin{minipage}[t]{0.2\linewidth}
 			\centering
 			\includegraphics[width=1.1in]{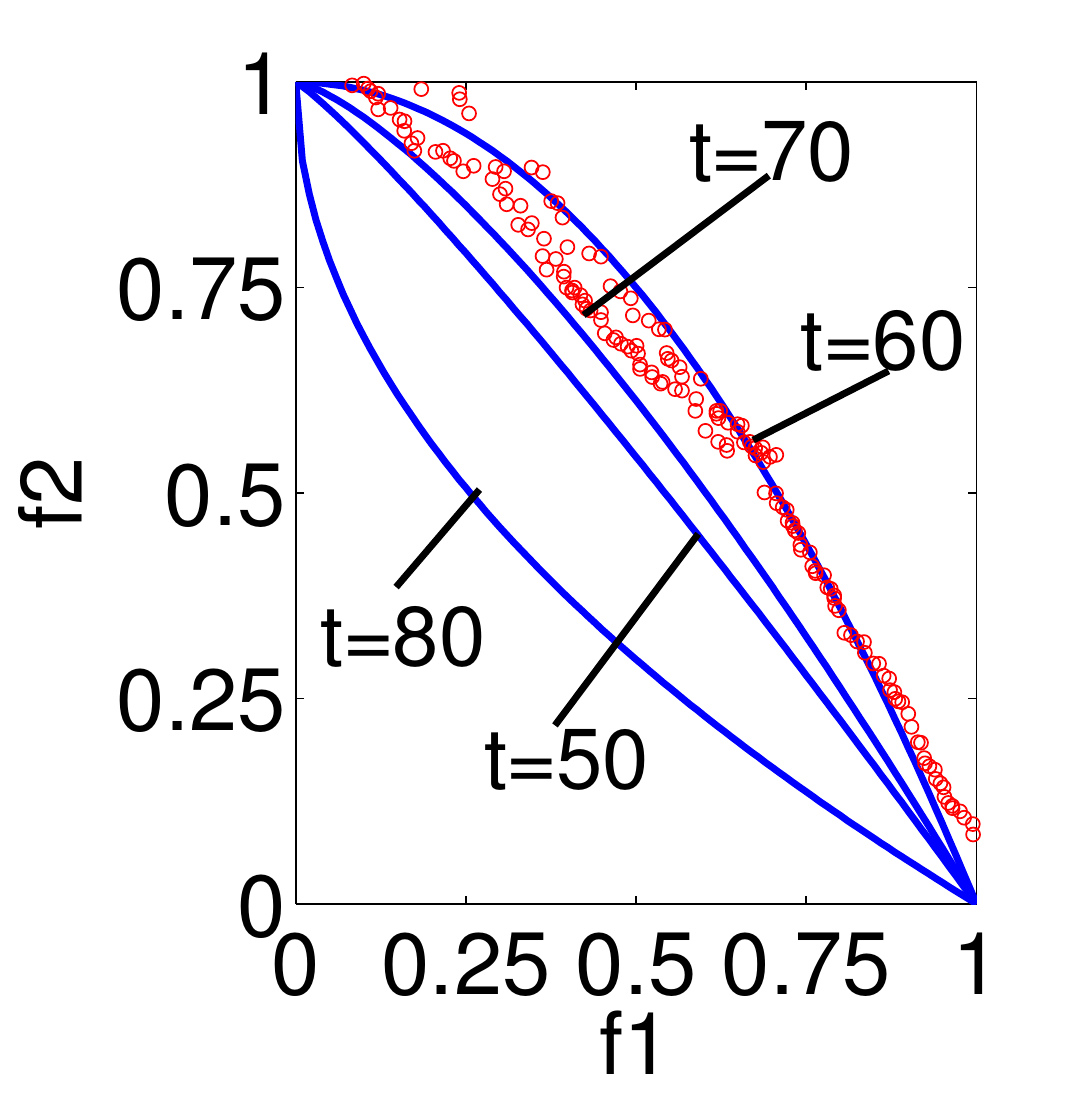}
 		\end{minipage}%
 	}%
 	\subfigure[\textbf{FPS}]{
 		\begin{minipage}[t]{0.2\linewidth}
 			\centering
 			\includegraphics[width=1.1in]{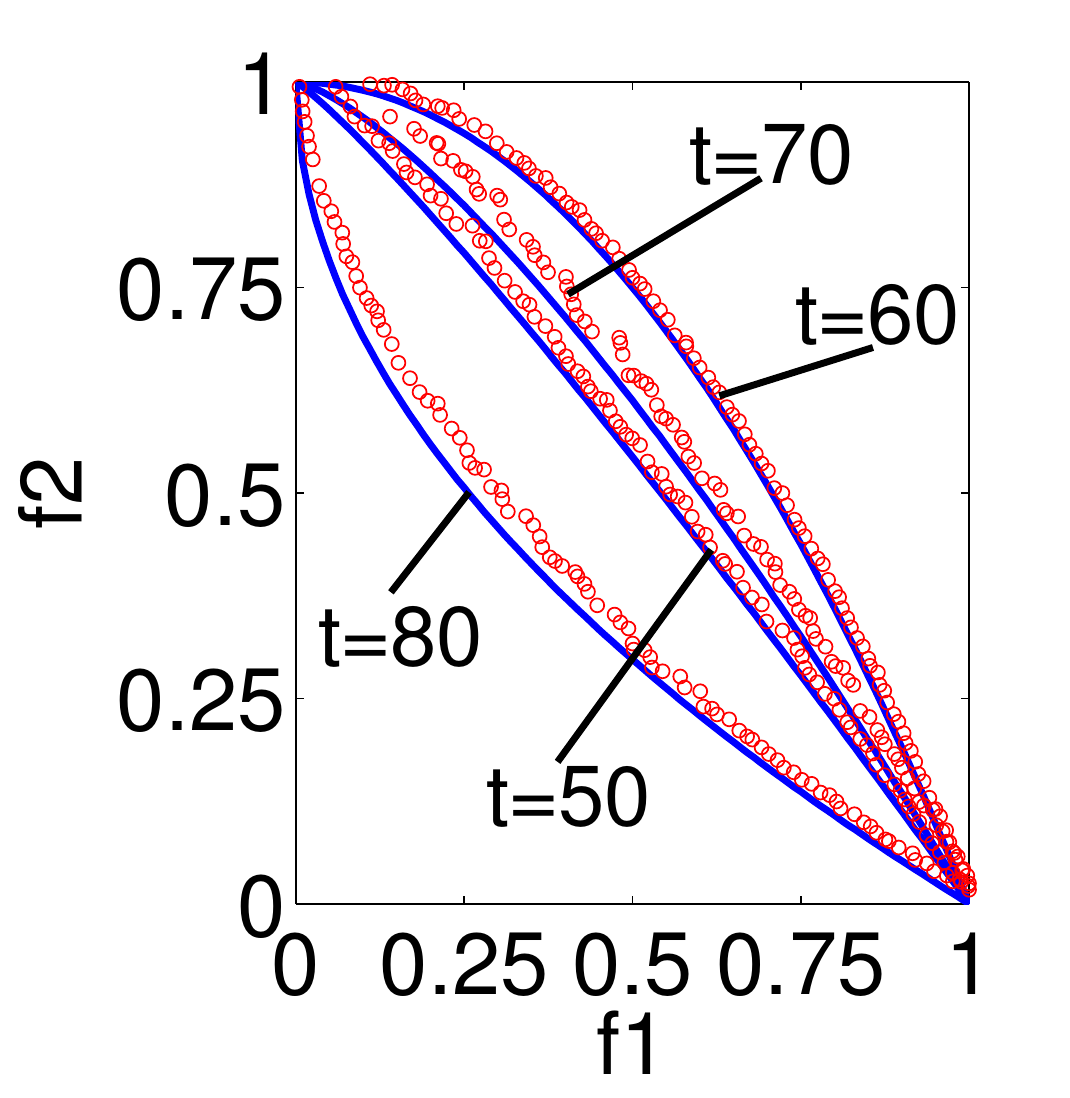}
 		\end{minipage}%
 	}%
 	\subfigure[\textbf{PPS}]{
 		\begin{minipage}[t]{0.2\linewidth}
 			\centering
 			\includegraphics[width=1.1in]{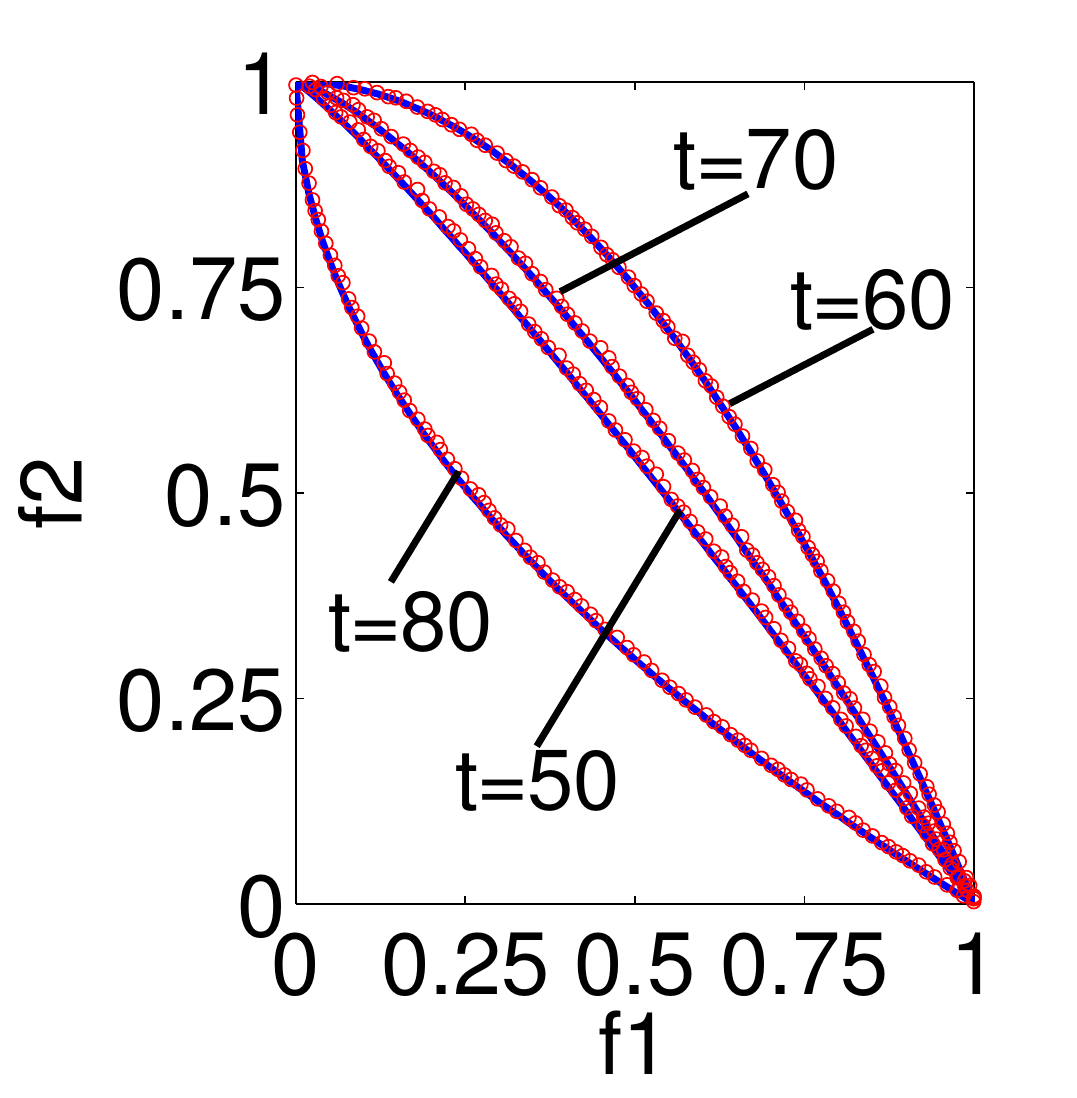}
 		\end{minipage}
 	}%
 	\subfigure[\textbf{SPPS}]{
 		\begin{minipage}[t]{0.2\linewidth}
 			\centering
 			\includegraphics[width=1.1in]{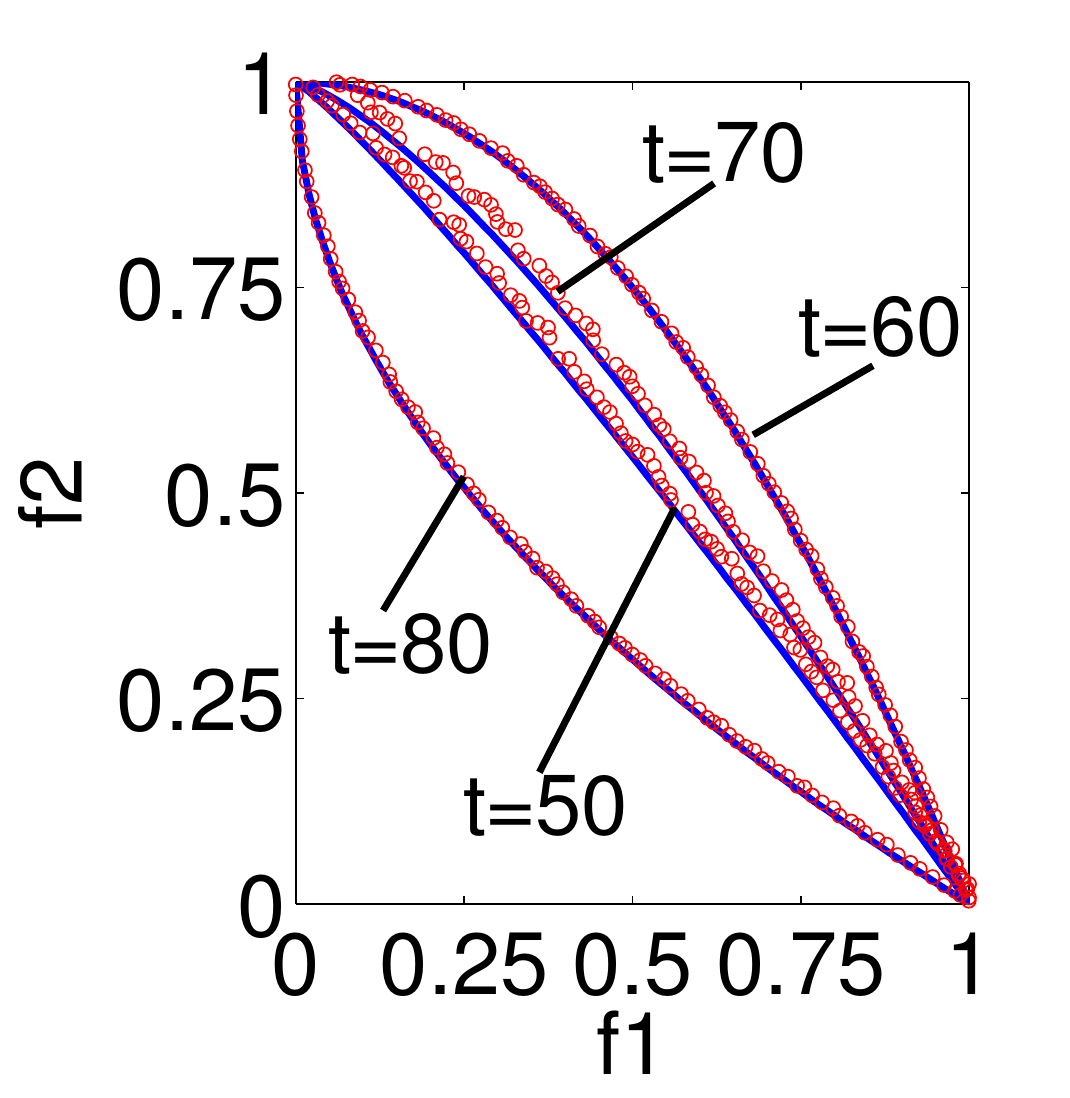}
 		\end{minipage}
 	}%
 	\subfigure[\textbf{FGERS-CPS}]{
 		\begin{minipage}[t]{0.2\linewidth}
 			\centering
 			\includegraphics[width=1.1in]{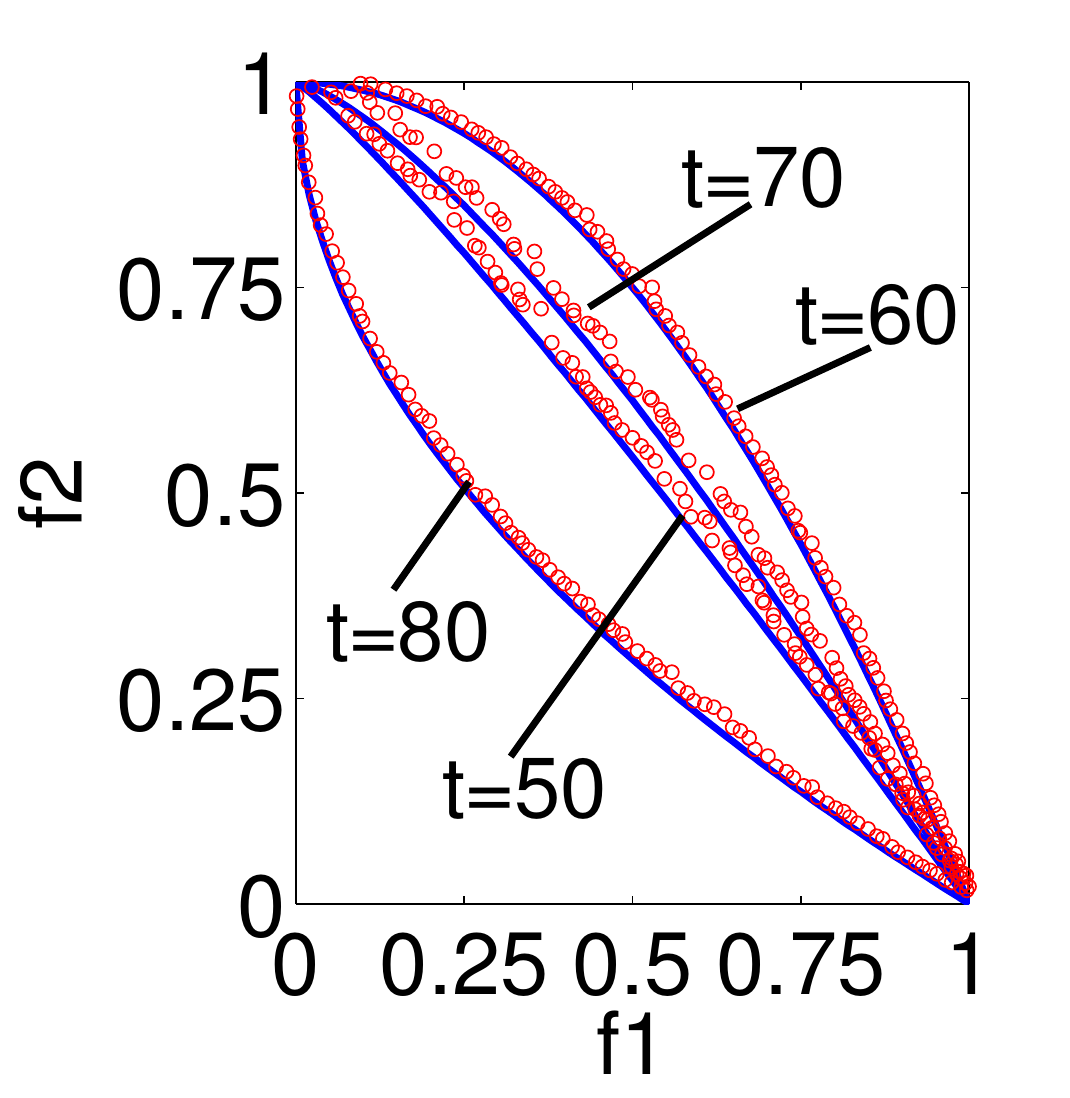}
 		\end{minipage}
 	}%
 	\centering
 	\caption{Final population distribution of the five strategies at eight time steps on dMOP2.}
 	\label{fig:6}
 \end{figure}

 \begin{figure}[htbp]
 	\centering
 	\subfigure[\textbf{ RIS}]{
 		\begin{minipage}[t]{0.2\linewidth}
 			\centering
 			\includegraphics[width=1.1in]{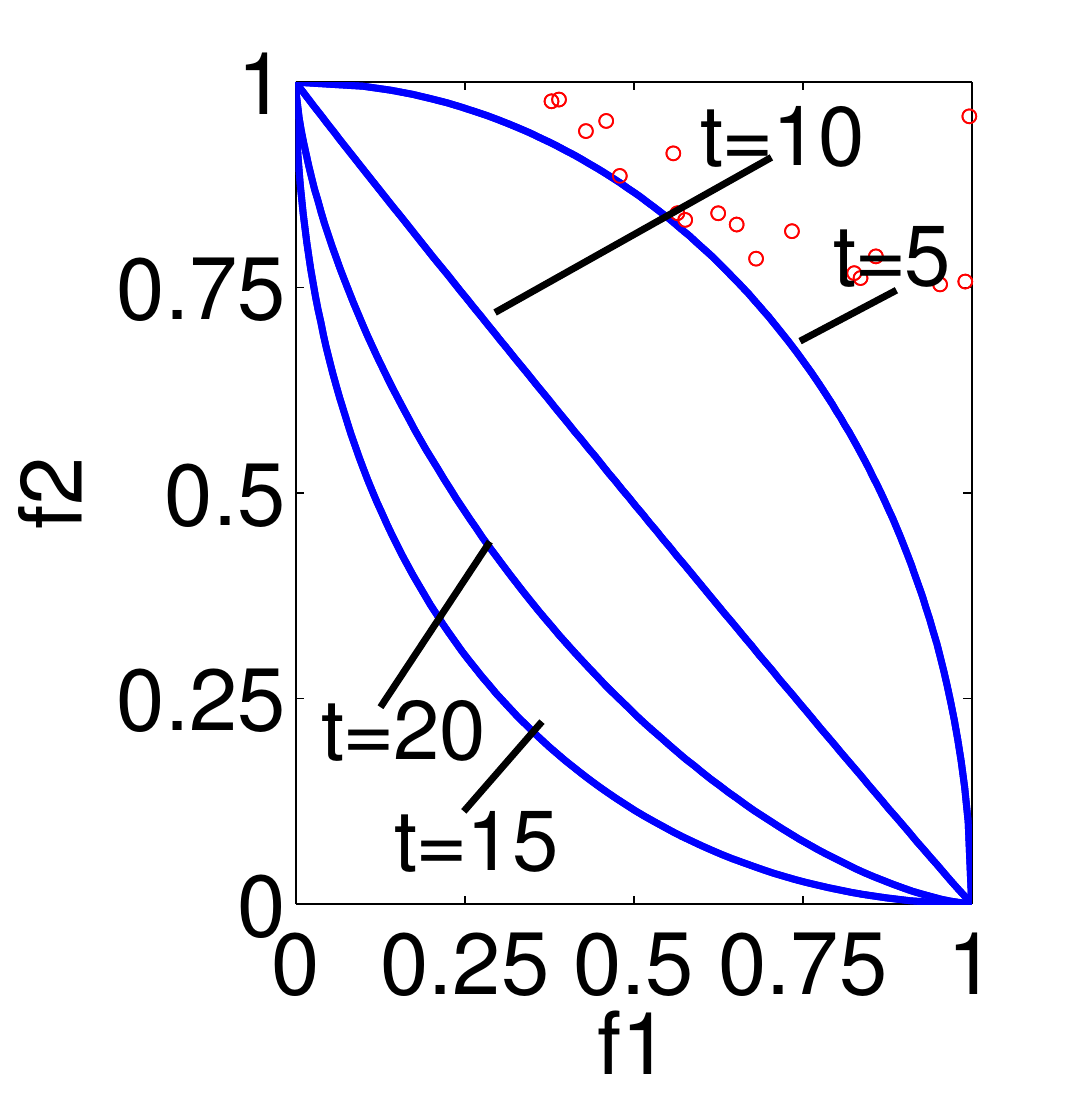}
 		\end{minipage}%
 	}%
 	\subfigure[ \textbf{FPS}]{
 		\begin{minipage}[t]{0.2\linewidth}
 			\centering
 			\includegraphics[width=1.1in]{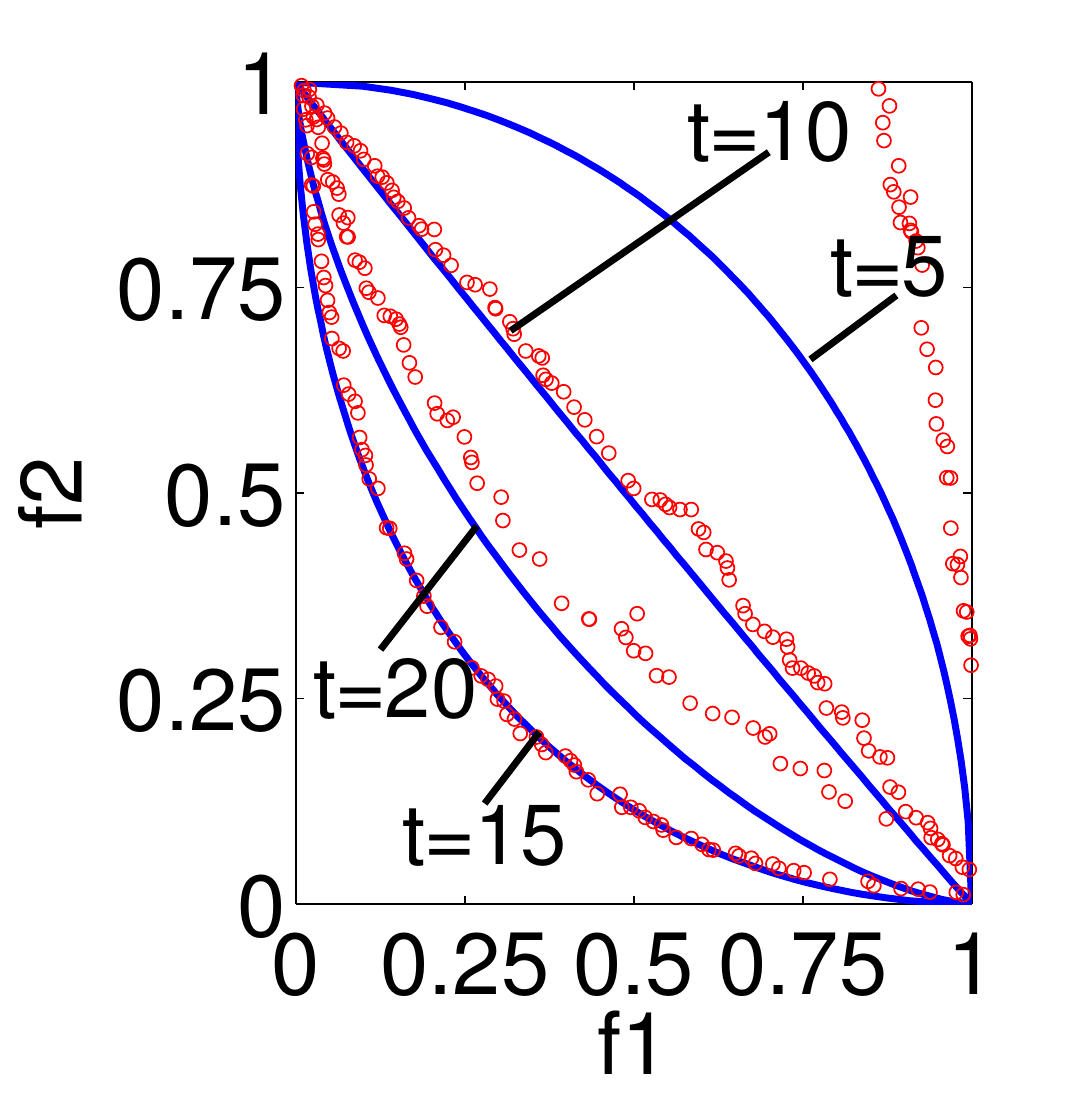}
 		\end{minipage}%
 	}%
 	\subfigure[\textbf{PPS}]{
 		\begin{minipage}[t]{0.2\linewidth}
 			\centering
 			\includegraphics[width=1.1in]{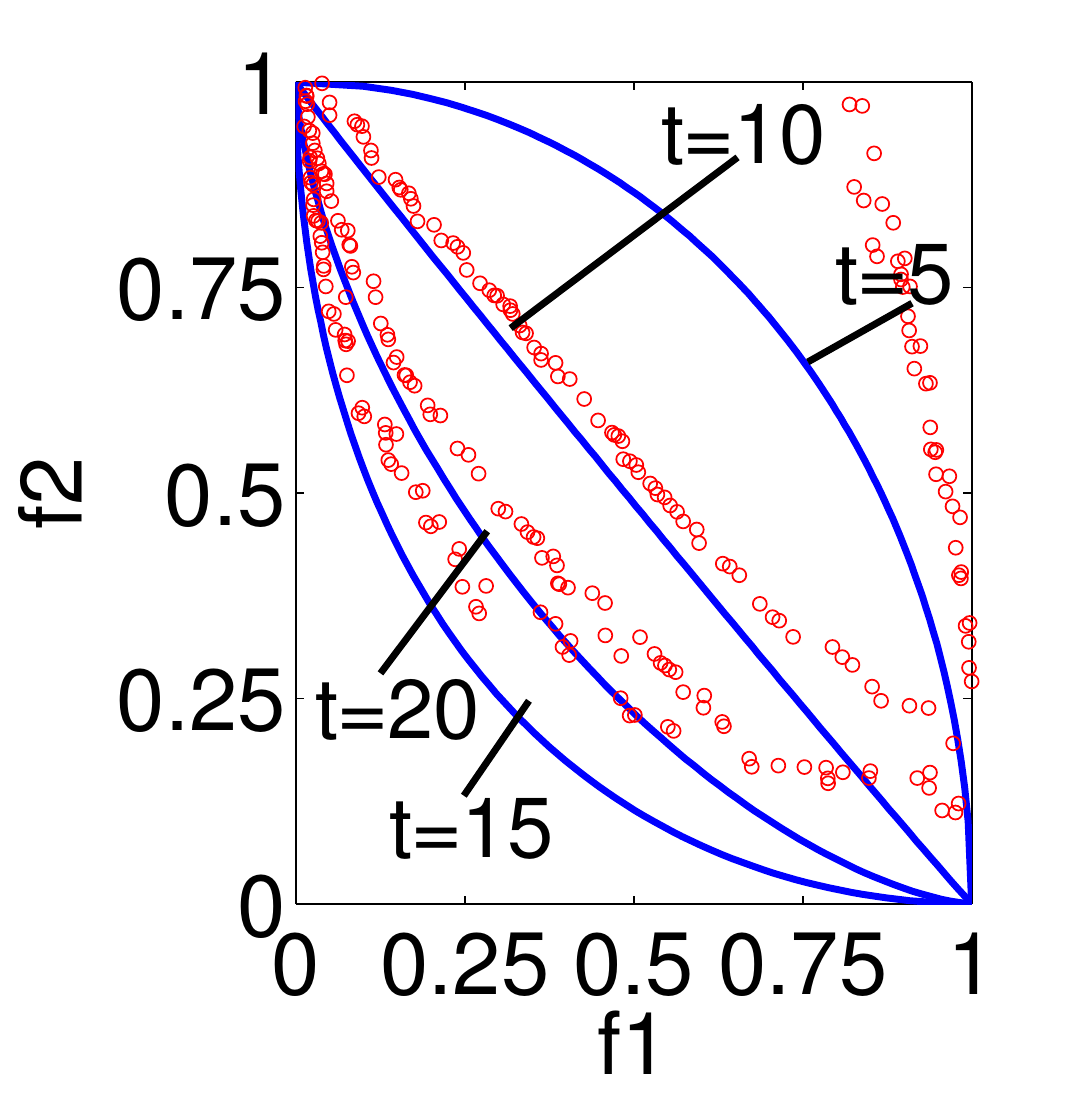}
 		\end{minipage}
 	}%
 	\subfigure[\textbf{SPPS}]{
 		\begin{minipage}[t]{0.2\linewidth}
 			\centering
 			\includegraphics[width=1.1in]{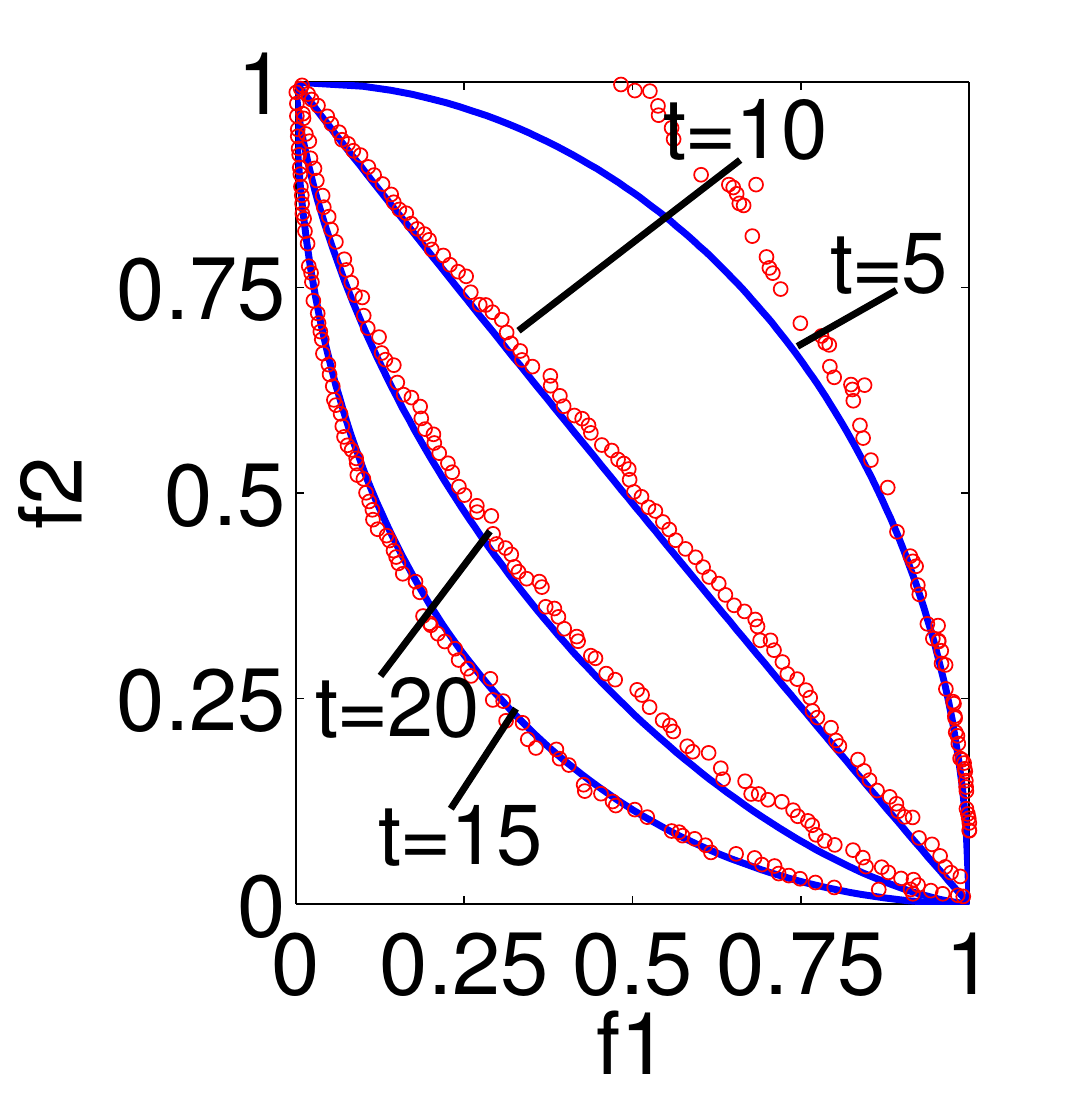}
 		\end{minipage}
 	}%
 	\subfigure[\textbf{FGERS-CPS}]{
 		\begin{minipage}[t]{0.2\linewidth}
 			\centering
 			\includegraphics[width=1.1in]{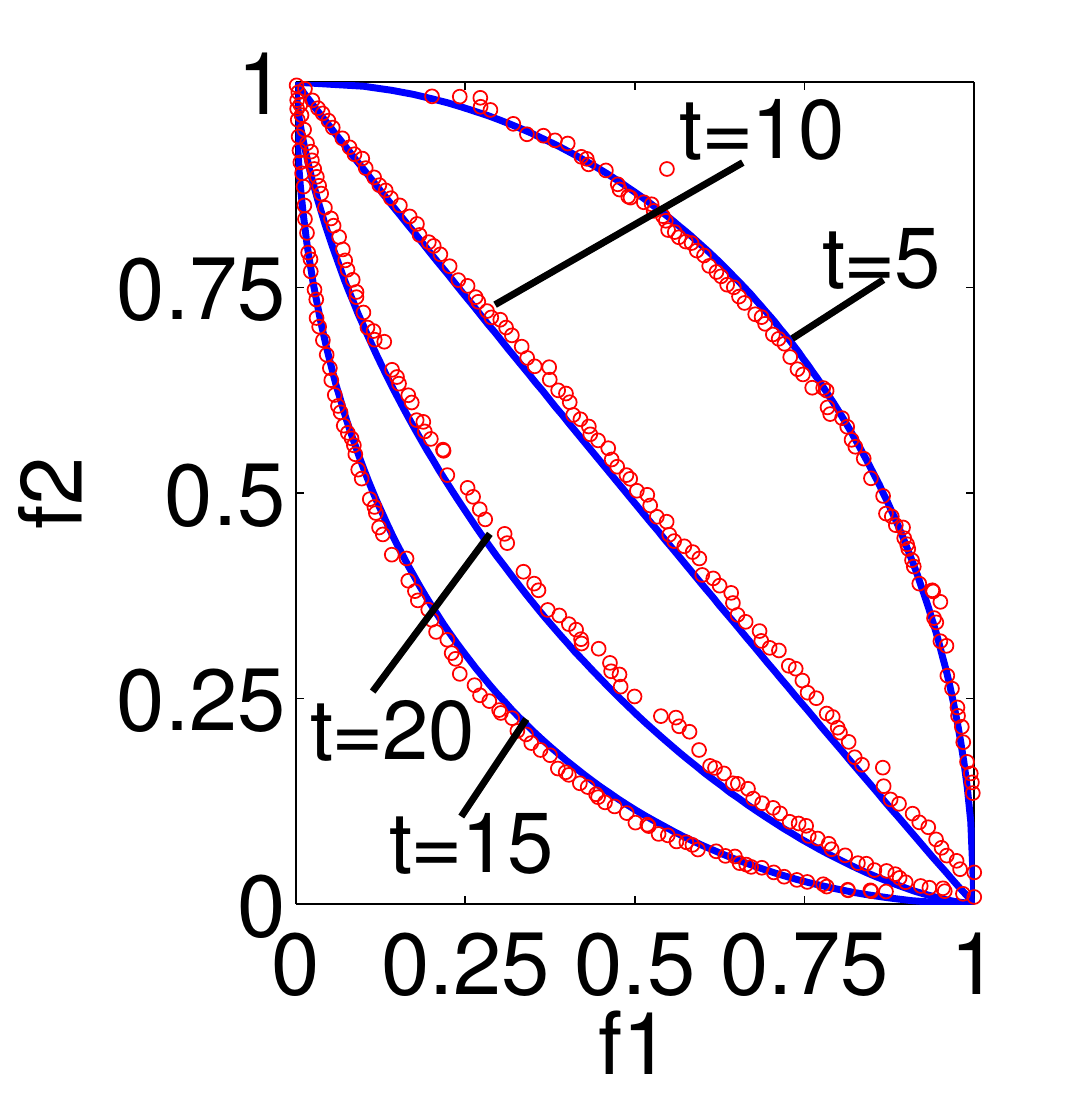}
 		\end{minipage}
 	}%
 	\\
 	\subfigure[\textbf{RIS}]{
 		\begin{minipage}[t]{0.2\linewidth}
 			\centering
 			\includegraphics[width=1.1in]{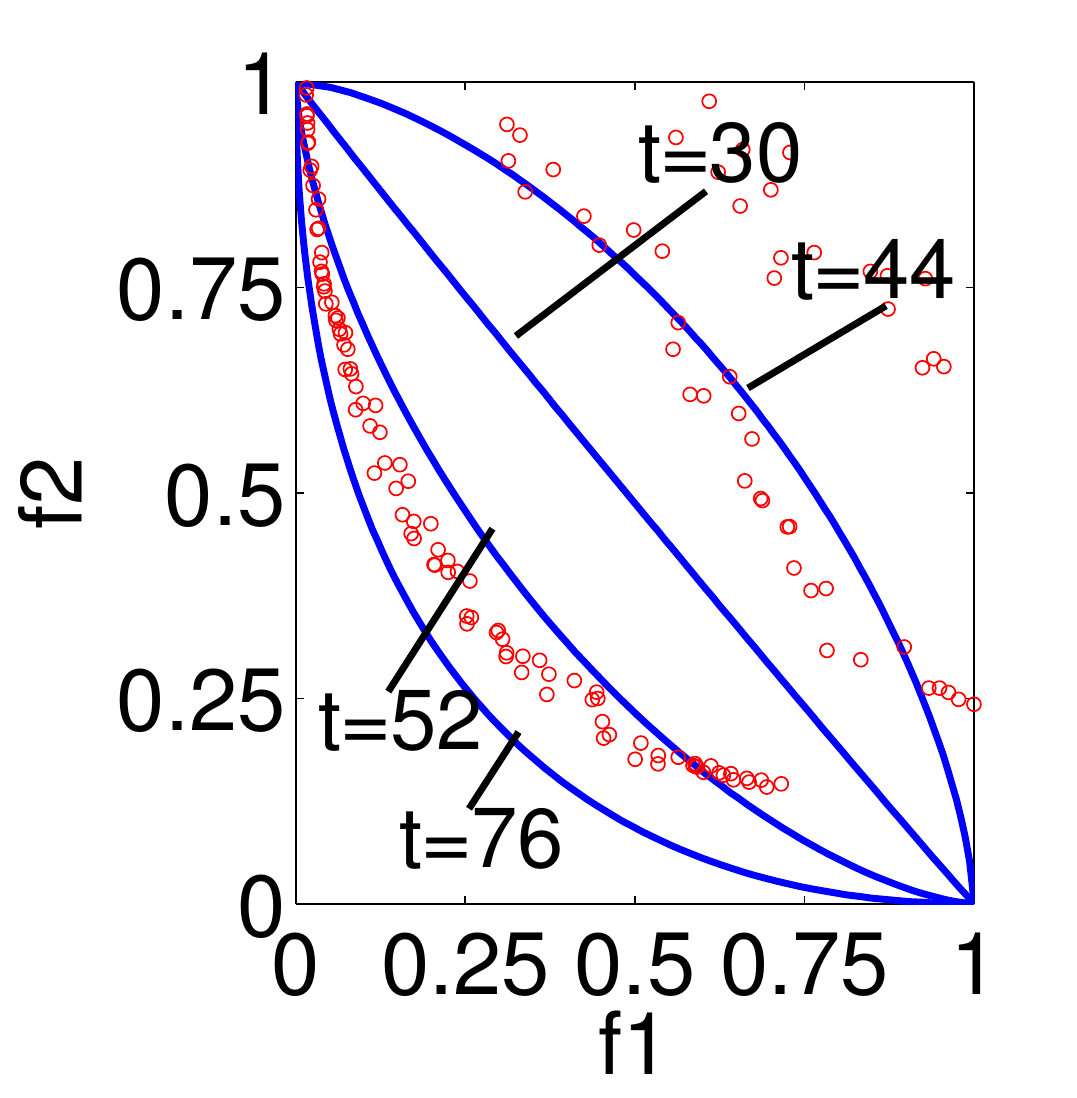}
 		\end{minipage}%
 	}%
 	\subfigure[\textbf{FPS}]{
 		\begin{minipage}[t]{0.2\linewidth}
 			\centering
 			\includegraphics[width=1.1in]{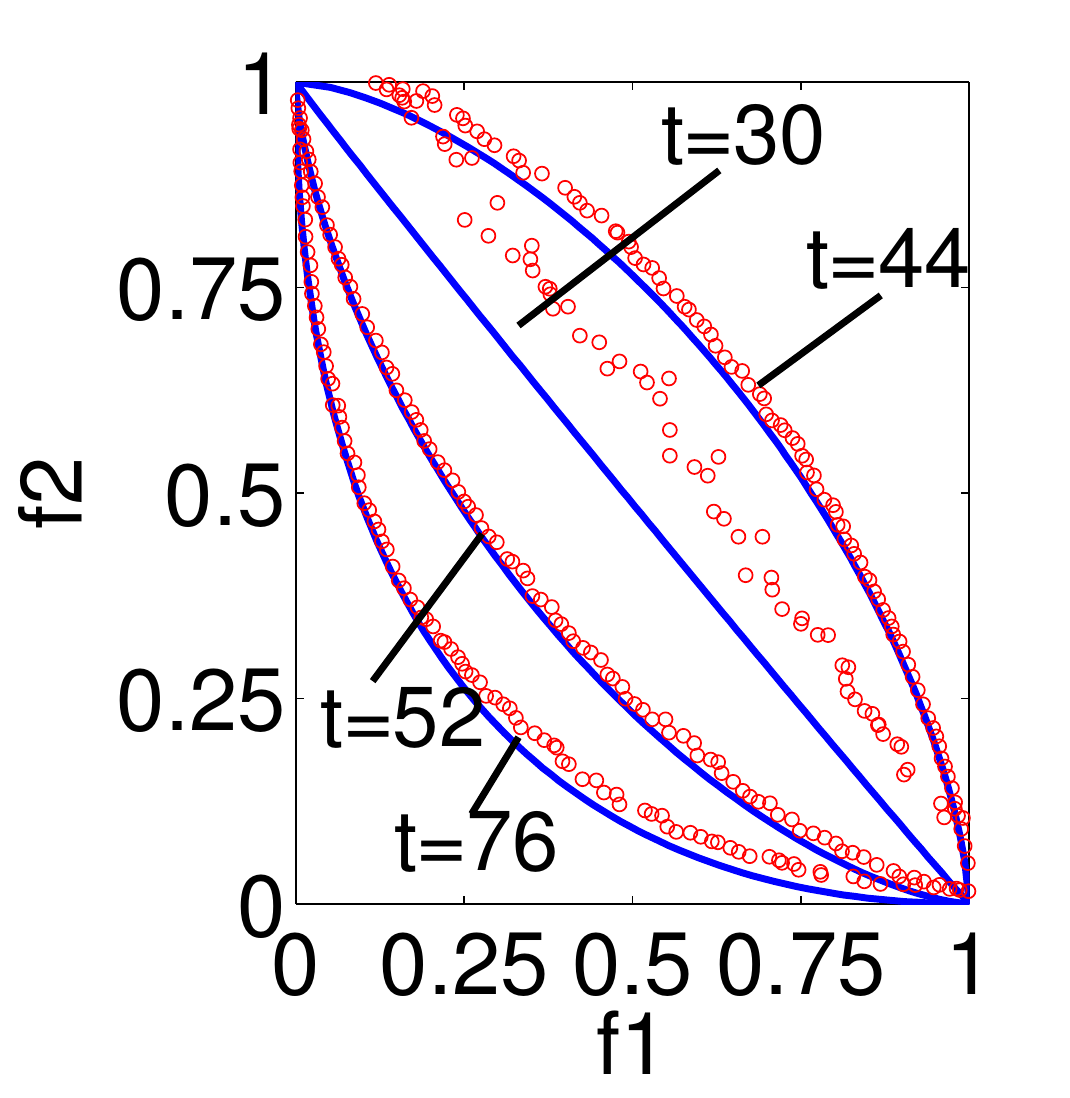}
 		\end{minipage}%
 	}%
 	\subfigure[\textbf{PPS}]{
 		\begin{minipage}[t]{0.2\linewidth}
 			\centering
 			\includegraphics[width=1.1in]{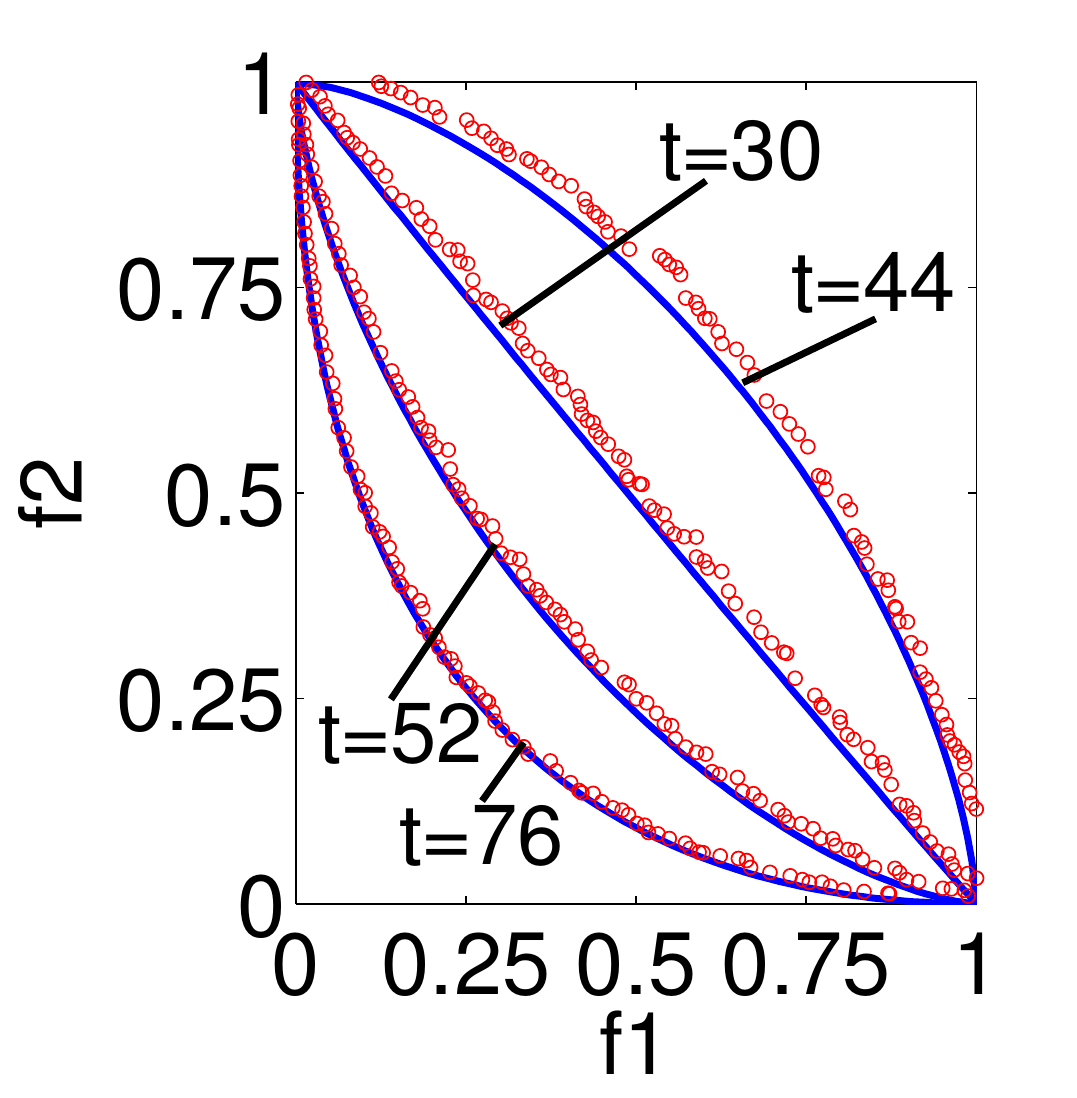}
 		\end{minipage}
 	}%
 	\subfigure[\textbf{SPPS}]{
 		\begin{minipage}[t]{0.2\linewidth}
 			\centering
 			\includegraphics[width=1.1in]{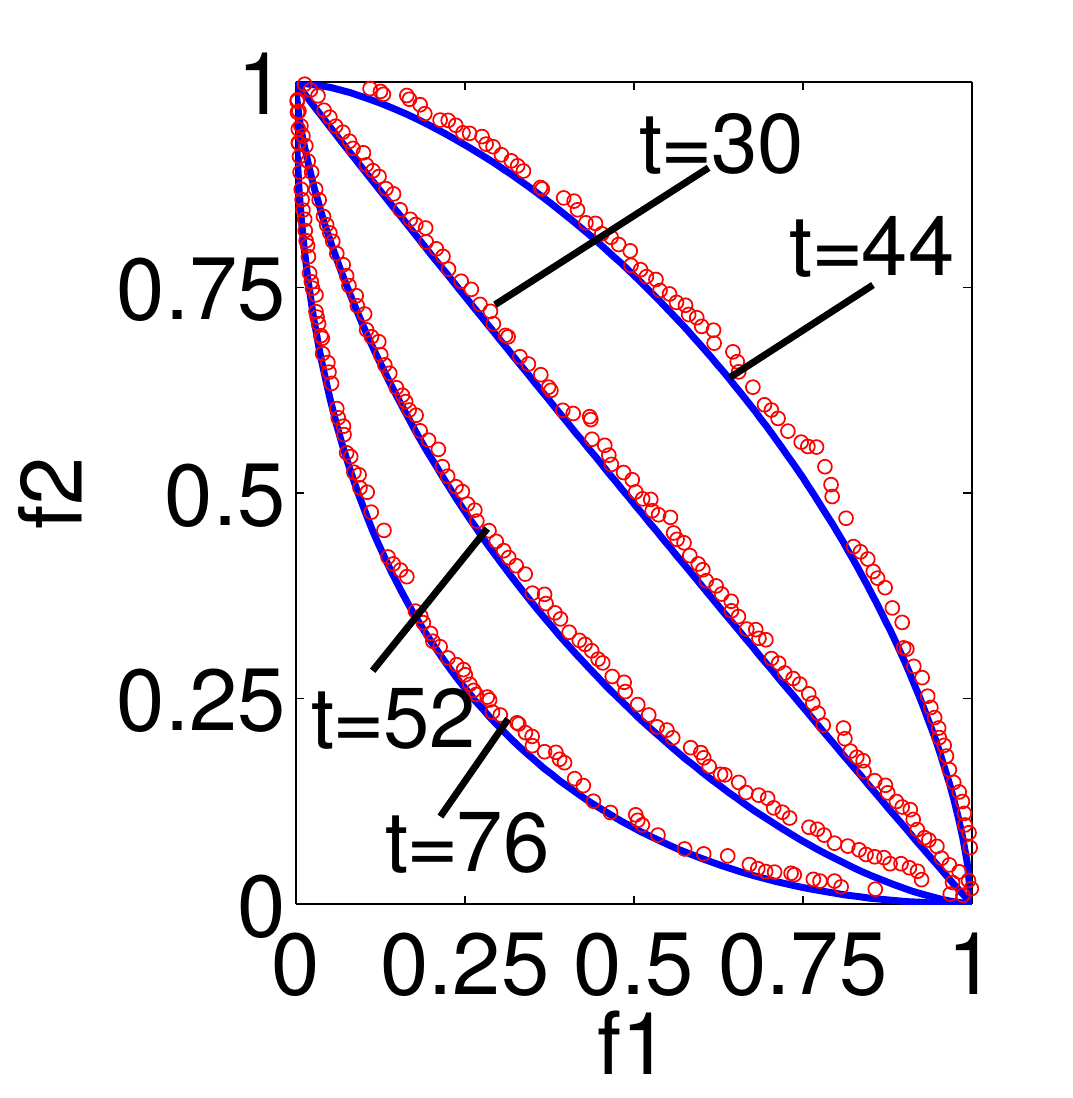}
 		\end{minipage}
 	}%
 	\subfigure[\textbf{FGERS-CPS}]{
 		\begin{minipage}[t]{0.2\linewidth}
 			\centering
 			\includegraphics[width=1.1in]{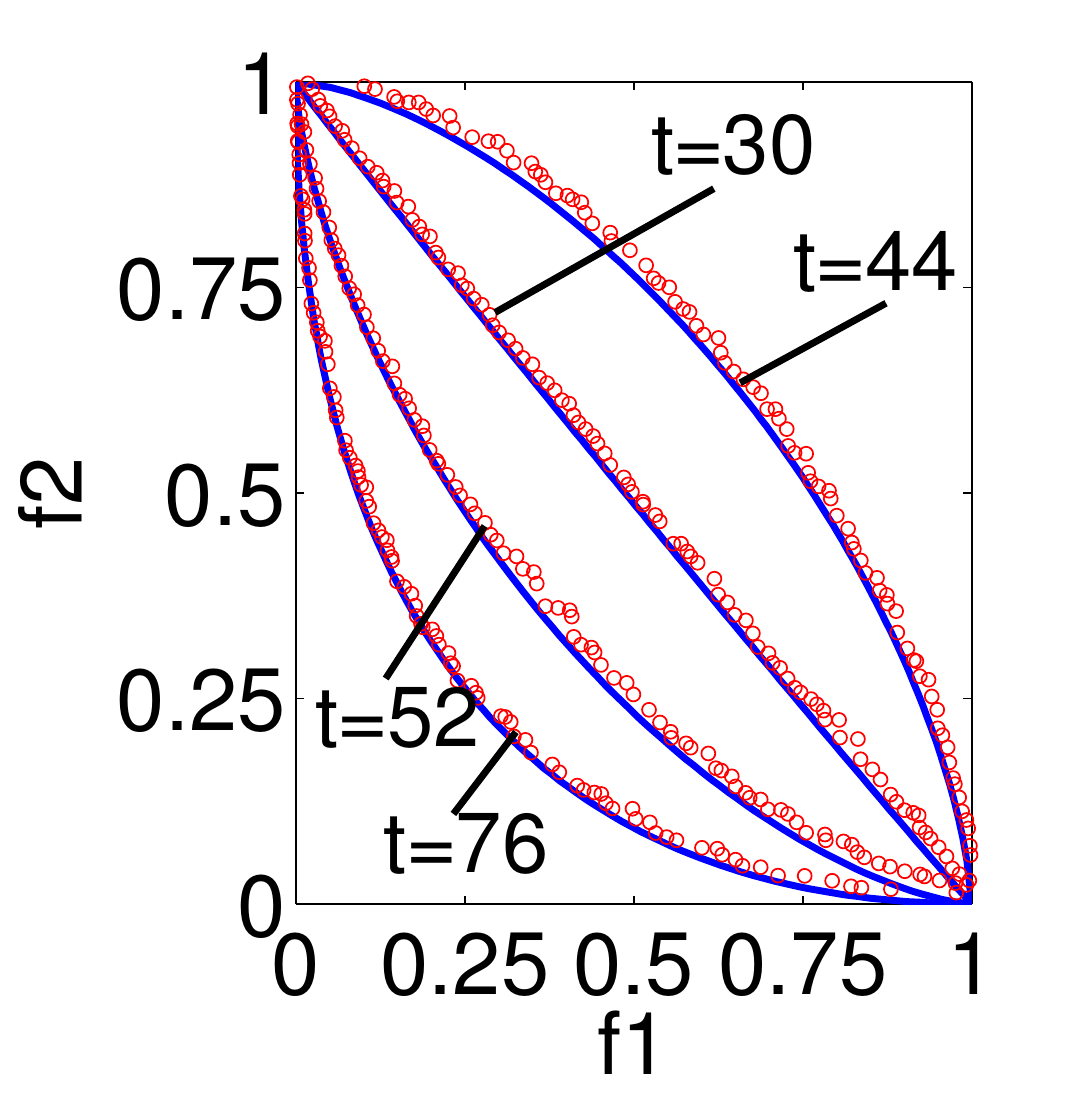}
 		\end{minipage}
 	}%
 	\centering
 	\caption{Final population distribution of the five strategies at eight time steps on F6.}
 	\label{fig:7}
 \end{figure}
\begin{figure}[htbp]
	\centering
	\subfigure[\textbf{ RIS}]{
		\begin{minipage}[t]{0.2\linewidth}
			\centering
			\includegraphics[width=1.1in]{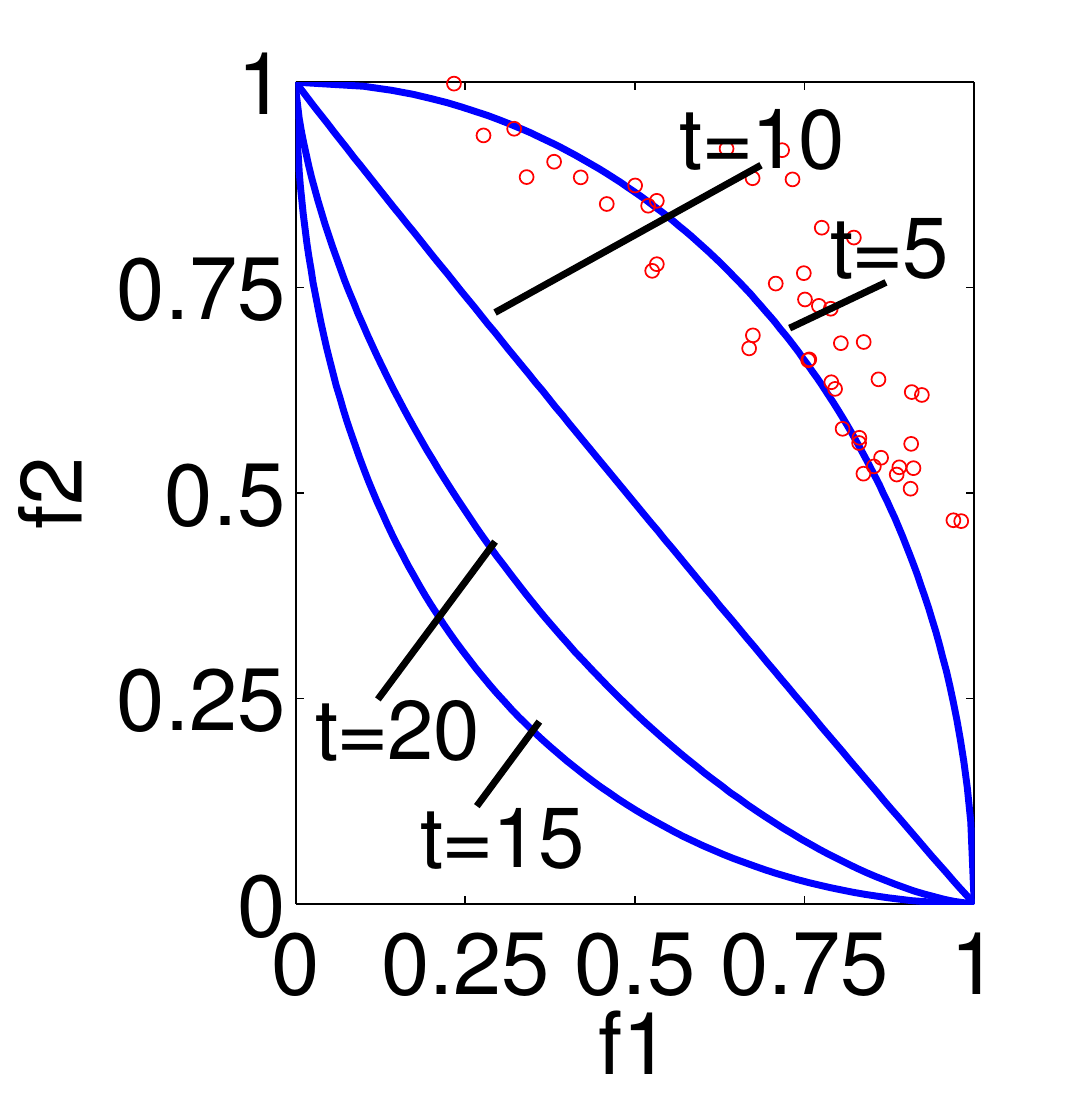}
		\end{minipage}%
	}%
	\subfigure[ \textbf{FPS}]{
		\begin{minipage}[t]{0.2\linewidth}
			\centering
			\includegraphics[width=1.1in]{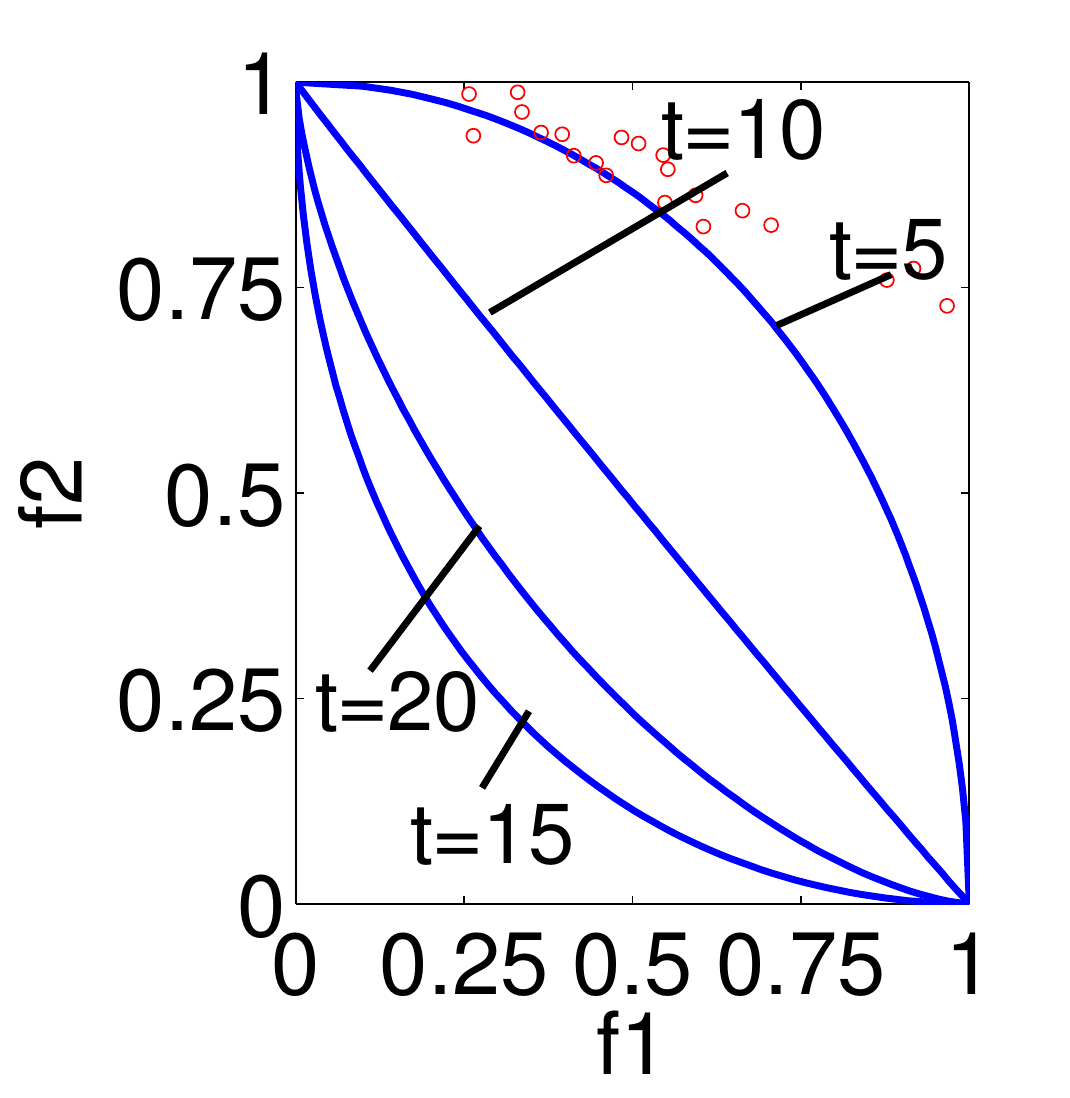}
		\end{minipage}%
	}%
	\subfigure[\textbf{PPS}]{
		\begin{minipage}[t]{0.2\linewidth}
			\centering
			\includegraphics[width=1.1in]{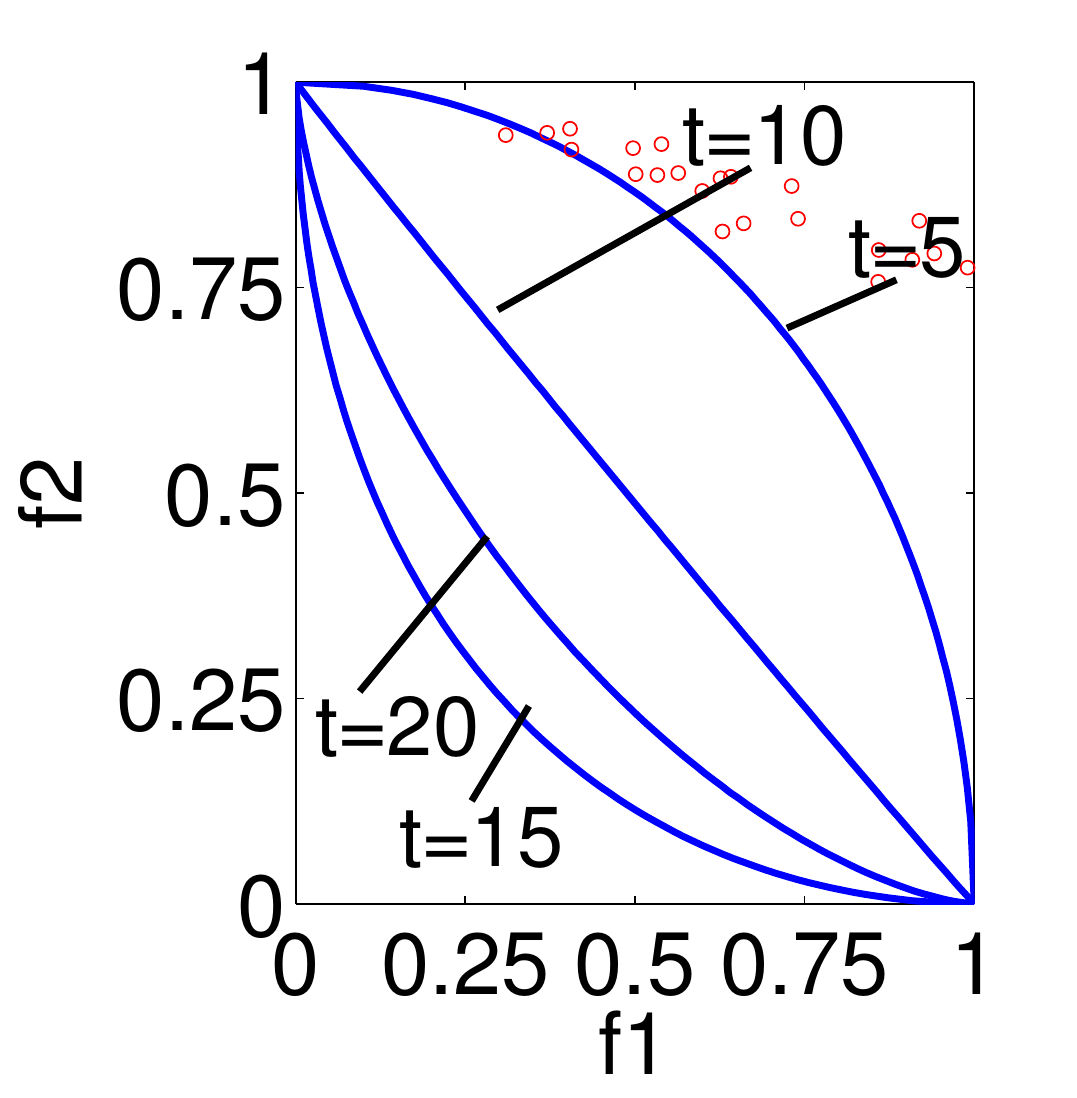}
		\end{minipage}
	}%
	\subfigure[\textbf{SPPS}]{
		\begin{minipage}[t]{0.2\linewidth}
			\centering
			\includegraphics[width=1.1in]{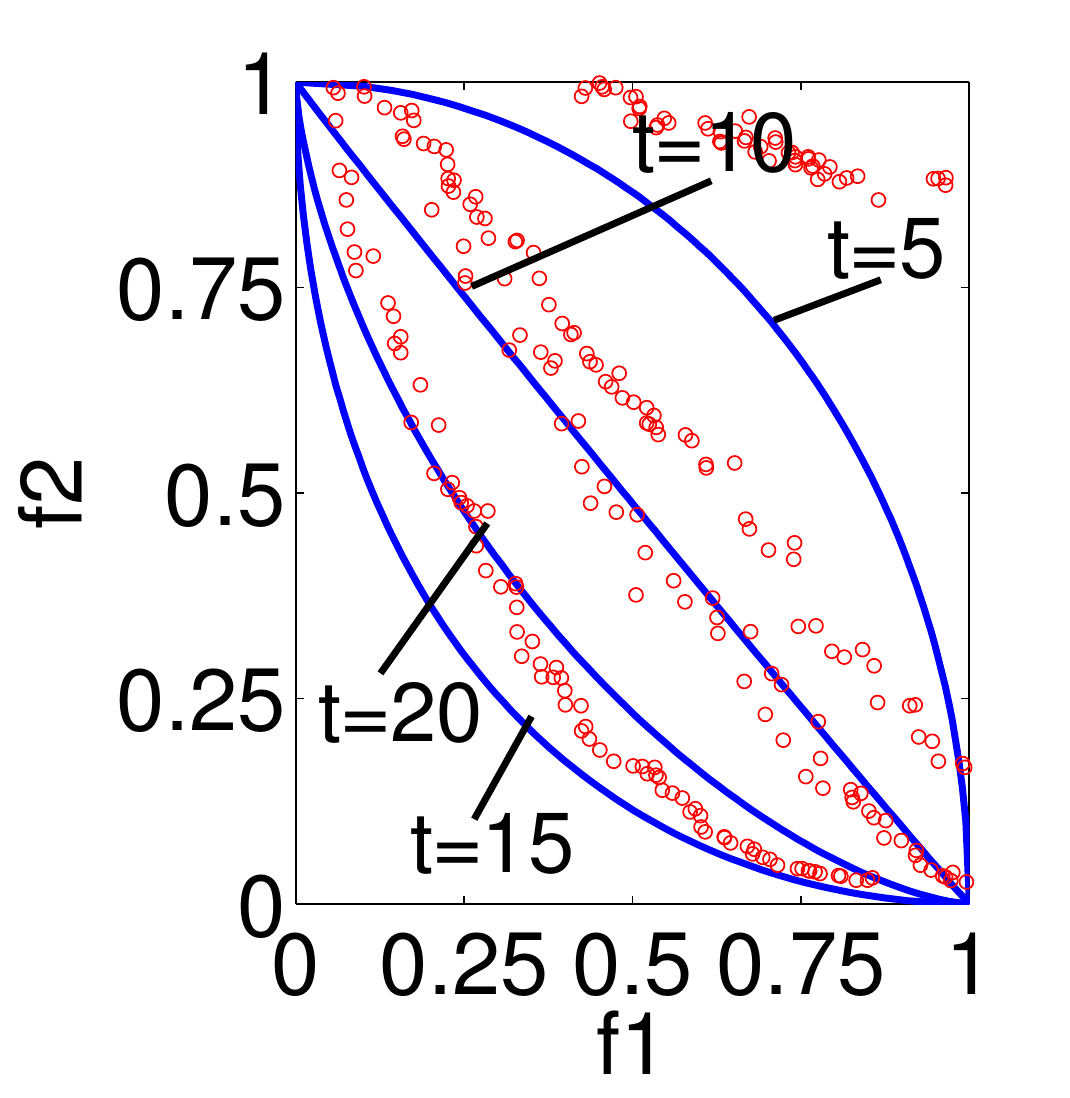}
		\end{minipage}
	}%
	\subfigure[\textbf{FGERS-CPS}]{
		\begin{minipage}[t]{0.2\linewidth}
			\centering
			\includegraphics[width=1.1in]{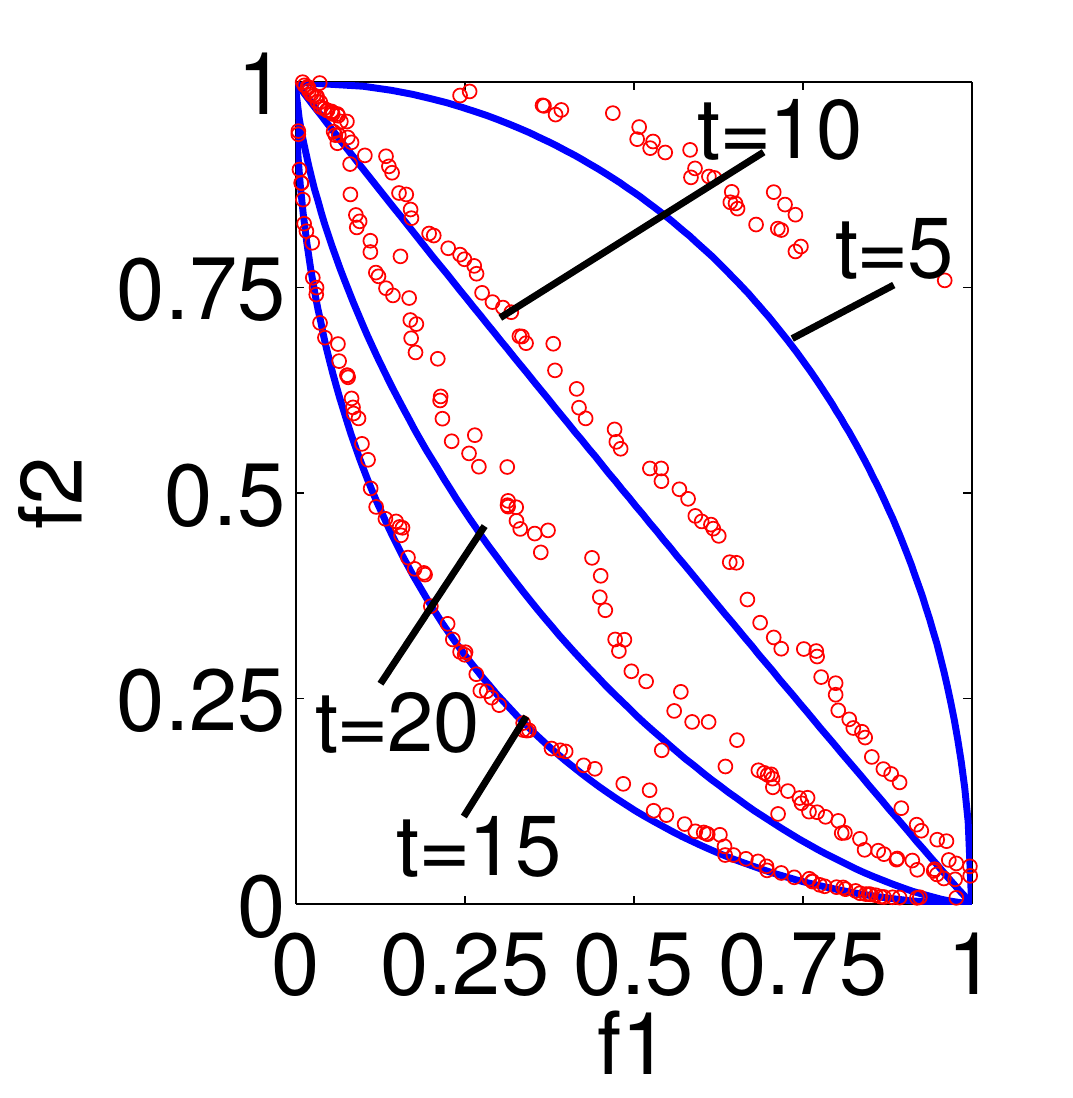}
		\end{minipage}
	}%
	\\
	\subfigure[\textbf{RIS}]{
		\begin{minipage}[t]{0.2\linewidth}
			\centering
			\includegraphics[width=1.1in]{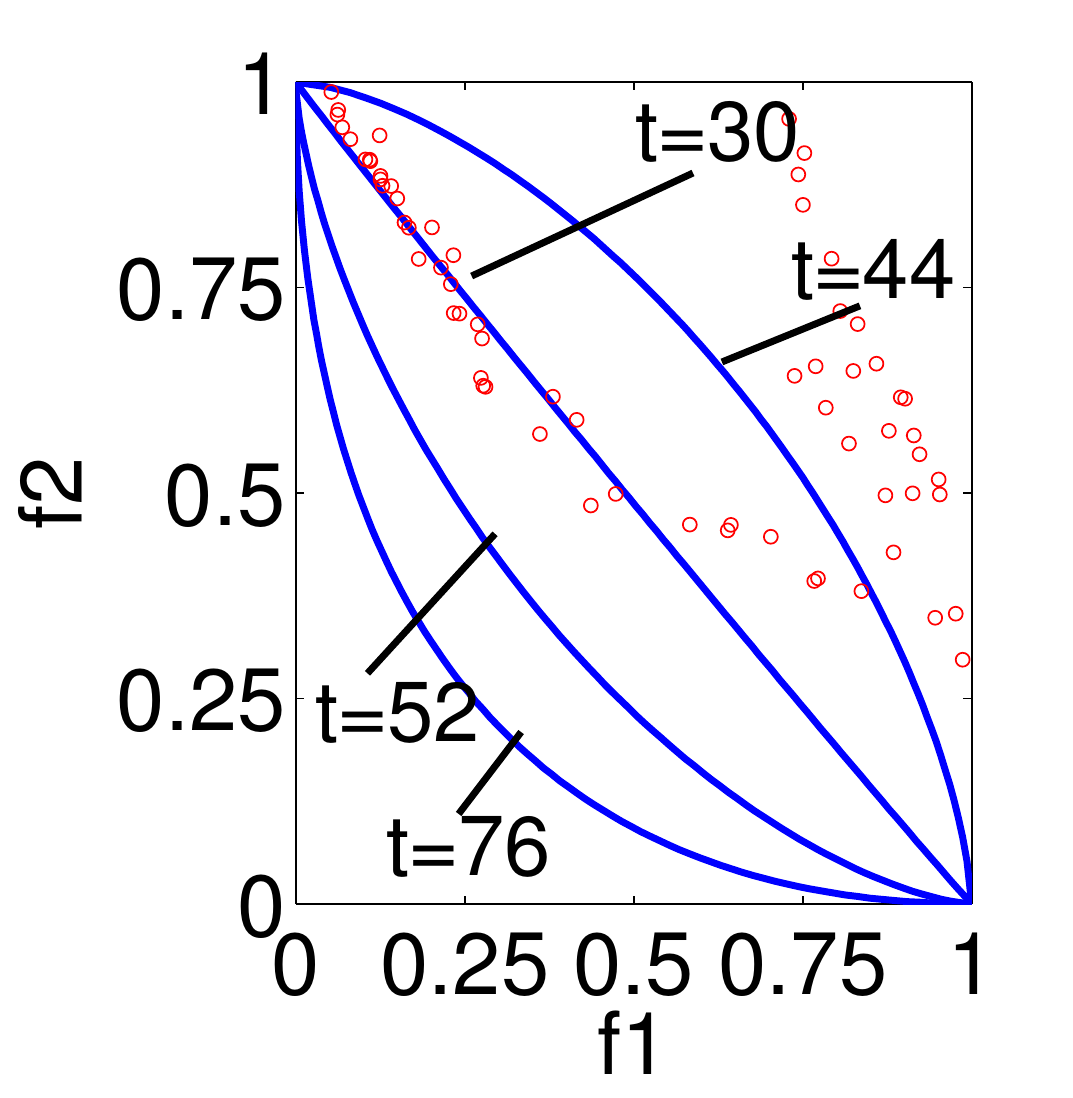}
		\end{minipage}%
	}%
	\subfigure[\textbf{FPS}]{
		\begin{minipage}[t]{0.2\linewidth}
			\centering
			\includegraphics[width=1.1in]{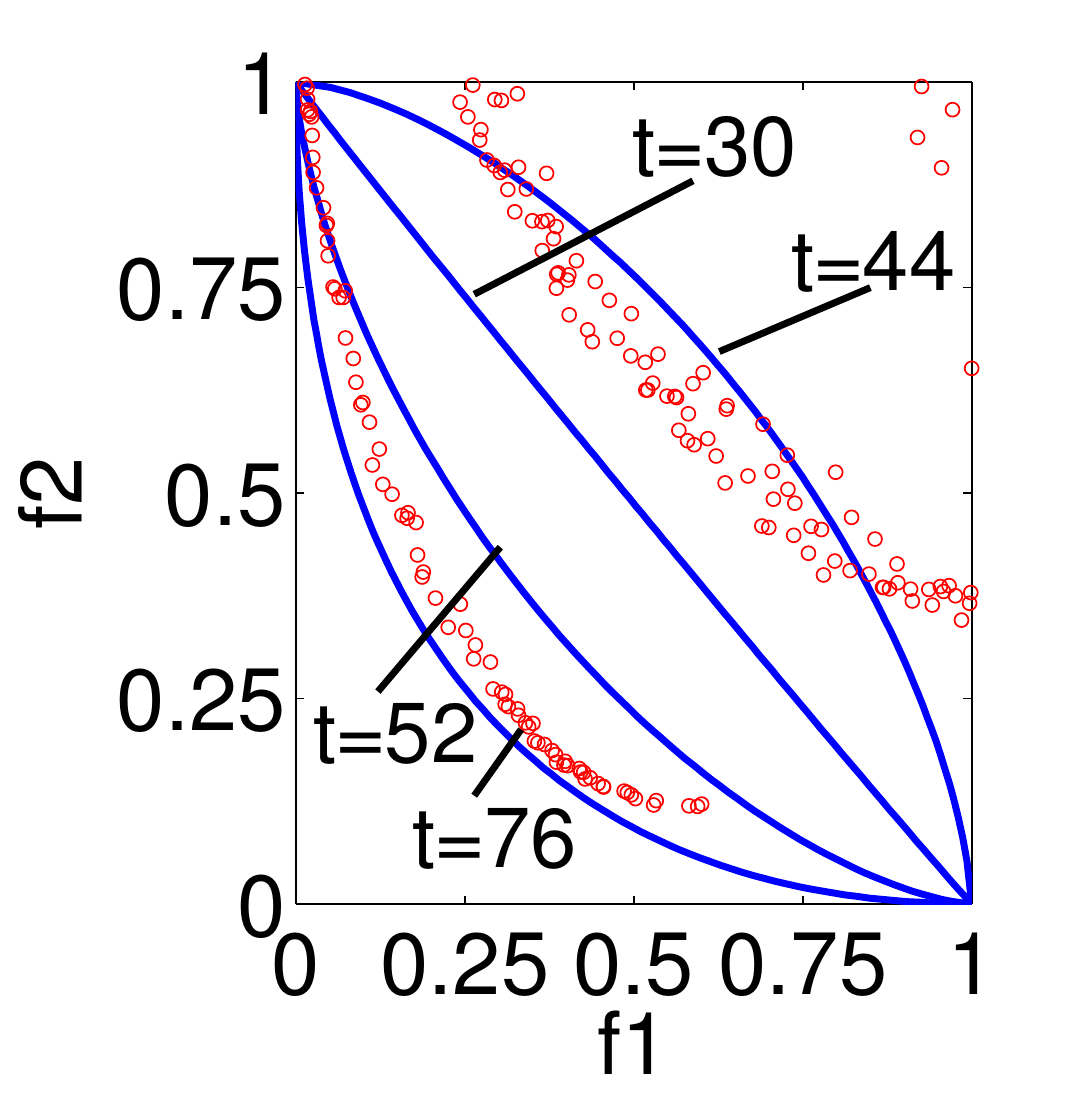}
		\end{minipage}%
	}%
	\subfigure[\textbf{PPS}]{
		\begin{minipage}[t]{0.2\linewidth}
			\centering
			\includegraphics[width=1.1in]{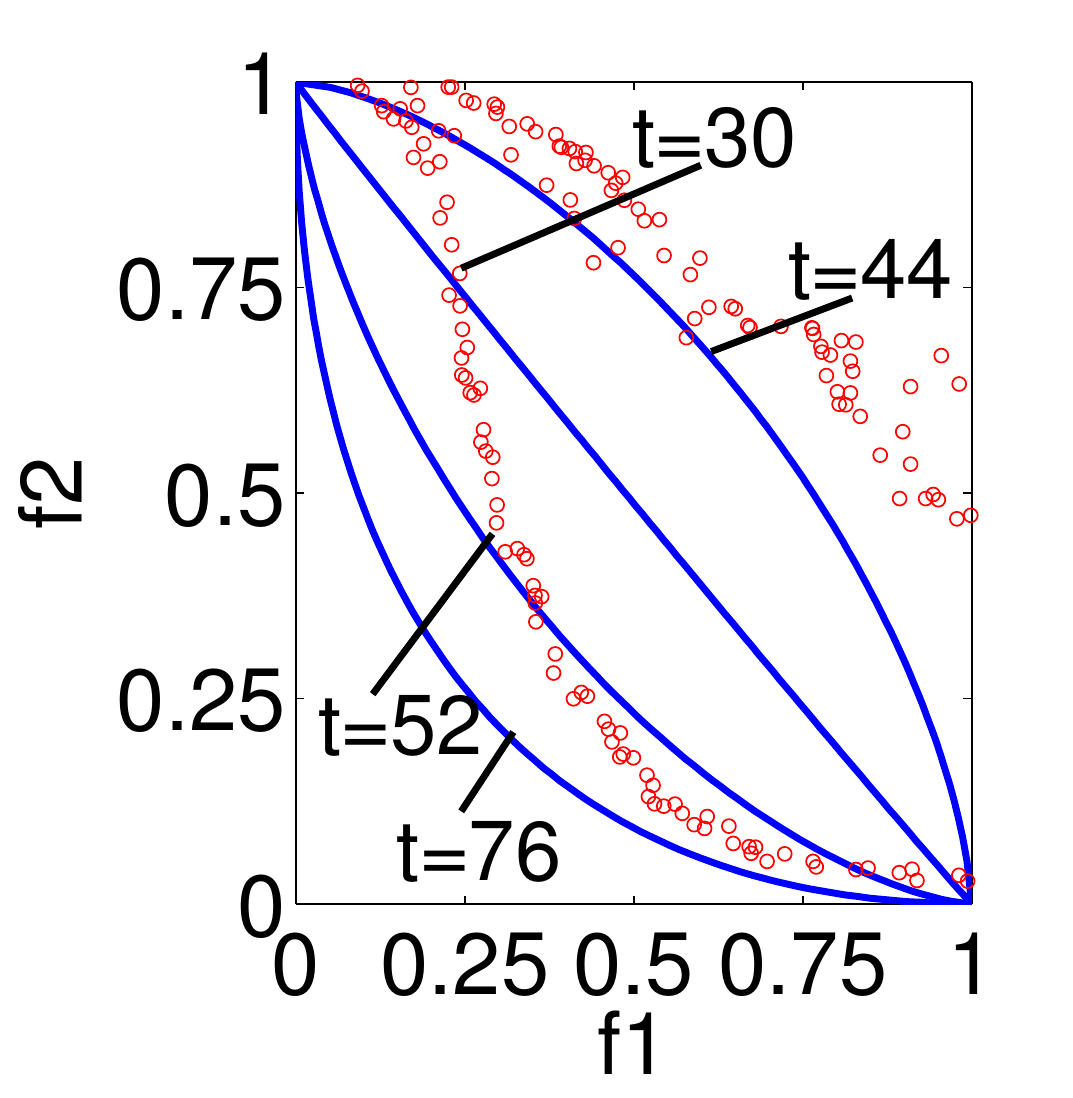}
		\end{minipage}
	}%
	\subfigure[\textbf{SPPS}]{
		\begin{minipage}[t]{0.2\linewidth}
			\centering
			\includegraphics[width=1.1in]{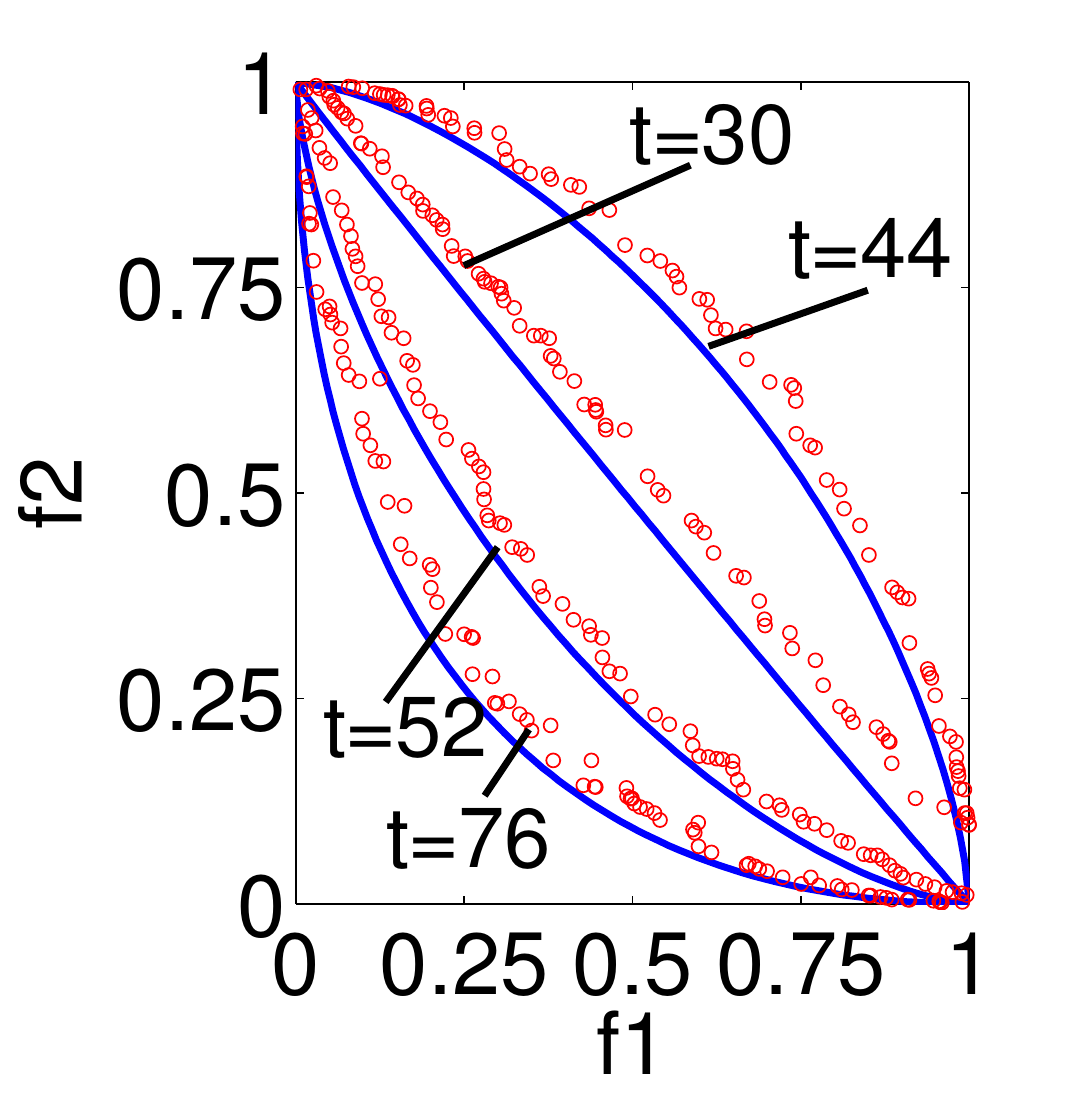}
		\end{minipage}
	}%
	\subfigure[\textbf{FGERS-CPS}]{
		\begin{minipage}[t]{0.2\linewidth}
			\centering
			\includegraphics[width=1.1in]{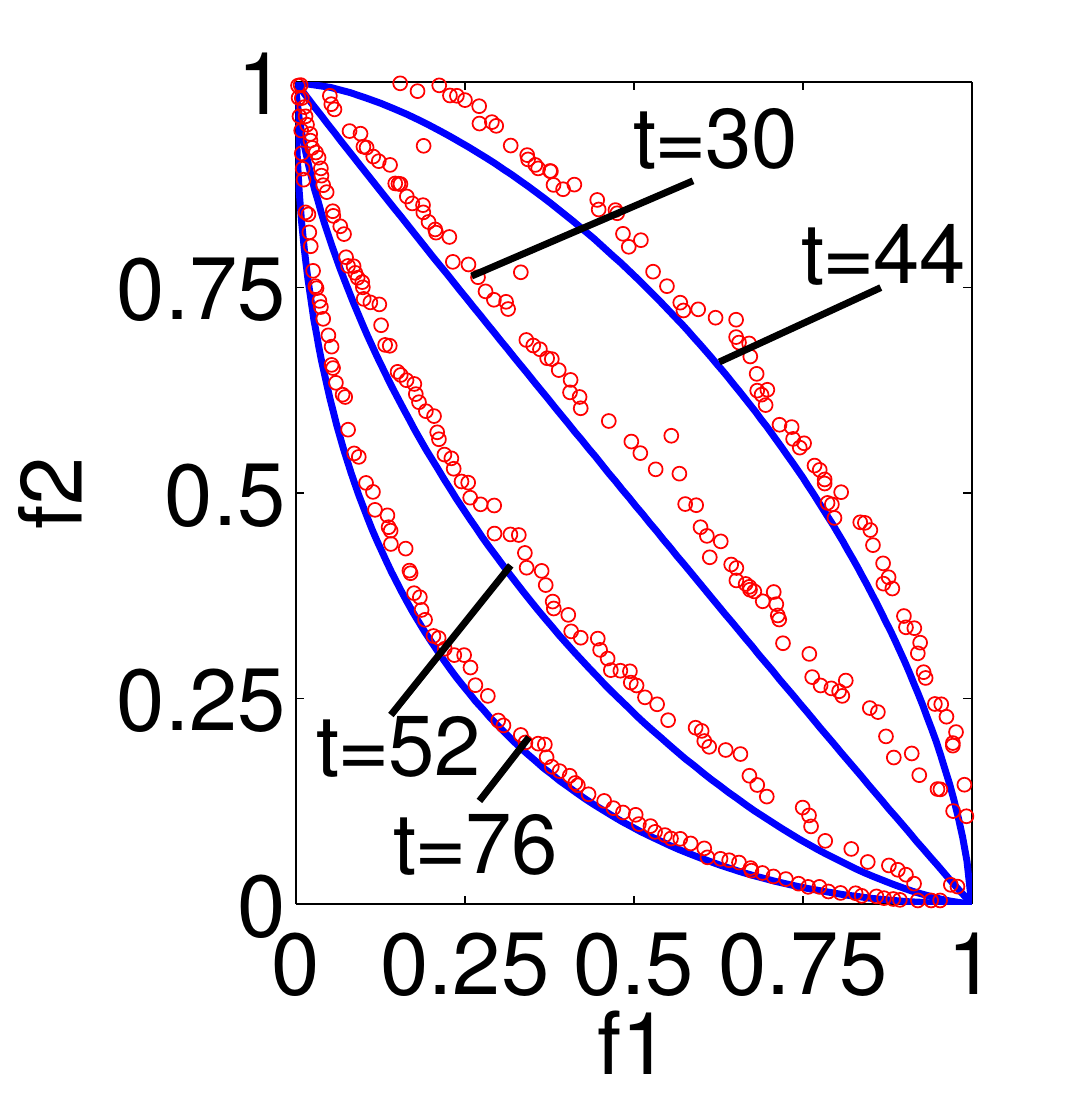}
		\end{minipage}
	}%
	\centering
	\caption{Final population distribution of the five strategies at eight time steps on F10.}
	\label{fig:8}
\end{figure}

 \subsection{Distribution diagram of final population}
 \label{sec:4.4}
 For a more intuitive comparison, four test problems with different characteristics, FDA1, dMOP2, F6, and F10, were selected to draw the final population distribution diagrams of five strategies. Figure \ref{fig:5} is Type 1 problem FDA1 with fixed PF, from which 6 moments were selected to observe the results. In the other diagrams, 8 moments were selected to observe the distribution results. In Figs. \ref{fig:5}-\ref{fig:8}, each red dot represents an individual in the population, and the blue line represents the true PF of the population. The closer the points are to the line, and the more uniform the distribution, the better the effect.
 
 It can be more clearly seen from Figs. \ref{fig:5} and \ref{fig:6} that the experimental results are basically consistent with Tables \ref{tab:2} and \ref{tab:3}, and the effect of FGERS-CPS is significantly better than RIS, FPS, and PPS. However, at these moments, the points obtained by SPPS and FGERS-CPS have almost converged to PF. At this time, it is difficult to find which strategy is better, only to be known by comparing data from Tables \ref{tab:2} and \ref{tab:3}. From Fig. \ref{fig:7}, the convergence and distribution of the five algorithms at eight moments can be shown. It is still obvious that the convergence effect of FGERS-CPS is better than other strategies on F6. For example, when $t$=5, the convergence effect of FGERS-CPS is better than SPPS. It can be seen from Fig \ref{fig:8}, the convergence effect of these strategies become worse than that in other figures, which shows F10 is a more complex problem than other problems. Nevertheless, FGERS-CPS still has a better convergence ability, whose effect is much better than RIS, FPS, and PPS, and better than SPPS slightly.



%
%


\section{More discussion}
\label{sec:5}
In order to further understand the role of FGERS, FGERS-CPS was compared with \textit{the feed-forward center point strategy in the traditional dynamic framework} (\textit{CPS}). In addition, FPS was also compared with combination of FPS and the generational response strategy of FGERS-CPS, which is to verify the convergence ability of the generational response strategy.

\subsection{Test instances and performance indicators}
\label{sec:5.1}
In this section, test problems were same as Section \ref{sec:4.1}, which are FDA test suite \cite{fda}, dMOP test suite \cite{dmop}, and F5-F10 \cite{pps}.

The performance indicator MIGD \cite{sgea} \cite{pps} was used to compare these two strategies in this section.  However, MIGD is the mean of IGD \cite{cao6} \cite{igd} values  at all time steps. To observe the convergence situation at every time step, IGD was also used. 
\subsection{Parameter settings}
\label{sec:5.2}
Let the frequency of environmental change be $\tau_t$ and the severity of environmental change be $n_t$. Three ($\tau_t$, $n_t$)s  were selected for testing:  (25, 10), (25, 5), (10, 10). When $\tau_t$ was set to 25 or 10, two figures represent high and low change frequency levels, respectively.   On the contrary, when $n_t$ is set to 10 and 5, two figures denote slight and severe environmental changes, respectively. When $\tau_t$ was 25, the number of generations was set to 2500.  While  $\tau_t$ was 10, the number of generations was set to 1000, so that the number of environmental change is 100. The other parameter settings were consistent with Section \ref{sec:4.2}. 
\begin{table*}
	\caption{Mean of MIGD values of CPS and FGERS-CPS on 13 benchmark DMOPs on three ($\tau_t$, $n_t$)s which are (25, 5), (25, 10) and (10, 10). The values in boldface denote to have the better effect between these two strategies.}
	\centering
	\label{tab10}
	\scriptsize
	\begin{tabular}{|C{1.5cm}|C{1.0cm}C{1.7cm}|C{1.0cm}C{1.7cm}|C{1.0cm}C{1.7cm}|}
		\hline\noalign{\smallskip}
		Problems	&\multicolumn{2}{c|}{(25,5)} 	&\multicolumn{2}{c|}{(25,10)} &\multicolumn{2}{c|}{(10,10)}\\
		{}      &CPS 	&FGERS-CPS	&CPS	&FGERS-CPS	&CPS	&FGERS-CPS\\
		\hline \noalign{\smallskip}
		FDA1	&0.0392	&\textbf{0.0123}	&0.0308	&\textbf{0.0112}	&0.1074	&\textbf{0.0446}\\
		FDA2	&0.0225	&\textbf{0.0219}	&0.0109	&\textbf{0.0095}	&0.0183	&\textbf{0.0141}\\
		FDA3	&1.0743	&\textbf{0.2664}	&0.0715	&\textbf{0.0135}	&0.3160	&\textbf{0.0564}\\
		FDA4	&0.1686	&\textbf{0.1562}	&\textbf{0.1169}	&0.1479	&\textbf{0.1710}	&0.1870\\
		dMOP1	&0.1984	&\textbf{0.1484}	&0.1538	&\textbf{0.1360}	&0.5152	&\textbf{0.4751}\\
		dMOP2	&0.1225	&\textbf{0.0893}	&0.0308	&\textbf{0.0133}	&0.1176	&\textbf{0.0557}\\
		dMOP3	&0.0388	&\textbf{0.0123}	&0.0284	&\textbf{0.0112}	&0.1104	&\textbf{0.0441}\\
		F5	&3.7479	&\textbf{0.2779}	&0.0716	&\textbf{0.0201}	&3.3434	&\textbf{0.1481}\\
		F6	&0.8530	&\textbf{0.1867}	&0.0202	&\textbf{0.0160}	&1.1106	&\textbf{0.0574}\\
		F7	&0.1978	&\textbf{0.1777}	&0.0229	&\textbf{0.0152}	&0.1147	&\textbf{0.0641}\\
		F8	&\textbf{0.1422}	&0.1785	&\textbf{0.1399}	&0.1737	&\textbf{0.2374}	&0.2554\\
		F9	&4.6058	&\textbf{0.5399}	&0.6193	&\textbf{0.1655}	&5.8486	&\textbf{1.1642}\\
		F10	&3.5809	&\textbf{0.2695}	&0.7556	&\textbf{0.0609}	&11.3837 &\textbf{1.0801}\\
		\noalign{\smallskip}\hline
	\end{tabular}
\end{table*}

\subsection{Comparison of CPS and FGERS-CPS}
\label{sec:5.3}
To be fair, the memory strategy and the adaptive diversity maintenance strategy were removed from FGERS-CPS. Only the feed-forward center point strategy was kept in FGERS-CPS, which means the only difference between FGERS-CPS and CPS is the generational response strategy in FGERS-CPS.

In this section,  CPS \cite{rong2019} \cite{ckps} and  FGERS-CPS  on 13 DMOPs with different characteristics were compared. Table \ref{tab10} lists the mean of MIGD values of CPS and FGERS-CPS on 13 benchmark DMOPs on three ($\tau_t$, $n_t$)s which are (25, 5), (25, 10), and (10, 10). The values in boldface denote to have a better effect between these two strategies.

As shown in Table \ref{tab10}, FGERS-CPS and CPS were compared on the FDA test suite \cite{fda}. Every figure in the table represents the mean of MIGD values. Decision variables in the FDA test suite are linearly related, so the FDA test suite is a relatively simple test suite.  As can be seen from Table \ref{tab10}, no matter whether ($\tau_t$, $n_t$) is (25, 5), (25, 10) or (10, 10) on FDA1, FDA2, and FDA3, MIGD values of FGERS-CPS are better than CPS. This shows that the generational response mechanism under the novel framework is rather effective.

On the three-dimensional problem FDA4, when ($\tau_t$, $n_t$) is (25, 5), the effect of FGERS-CPS is better than CPS. But when ($\tau_t$, $n_t$) is (25, 10) and (10, 10), FGERS-CPS is slightly worse than CPS. This may be related to the values of $n_t$. When $n_t$ is 5,  compared with 10, the severity of the environmental change was larger. At this time, it is difficult to converge for CPS with only an environmental response mechanism, but for FGERS-CPS, a generational response mechanism is added to make the effect better. This shows that FGERS-CPS can perform better in the severer environment. 

As shown in Table \ref{tab10}, FGERS-CPS and CPS were compared on the dMOP test suite \cite{dmop}. The dMOP test suite is also a test suite with linear-correlated decision variables, but it is more complicated than the FDA test suite, which means that the dMOP test suite is more difficult to converge for the algorithm. As can be seen from the experimental results, whether ($\tau_t$, $n_t$) is  (25, 5), (25, 10) or (10, 10), FGERS-CPS is better than CPS. This denotes that the generational response mechanism under the novel framework is working, because the only difference between FGERS-CPS and CPS is that FGERS-CPS has a generational response mechanism that CPS doesn't have.

As shown in Table \ref{tab10}, FGERS-CPS and CPS were compared and tested on the F5-F7 \cite{pps}. F5-F7 are test problems whose decision variables are nonlinear correlation. Compared with the FDA and the dMOP test suites, it is more difficult for the algorithm to converge. As can be seen from Table \ref{tab10}, no matter ($\tau_t$, $n_t$) is under (25, 5), (25, 10), and (10, 10), the MIGD values of FGERS-CPS are much better than that of CPS. For example, when ($\tau_t$, $n_t$) = (25, 5), on F5, the MIGD value of FGERS-CPS is 0.2779 and the MIGD value of CPS is 3.7479, which shows that the effect of FGERS-CPS is far better than CPS.

In Table \ref{tab10}, FGERS-CPS and CPS were compared and tested on the F8-F10 \cite{pps}. F8-F10 are also the test problems whose decision variables are nonlinear correlation. Among them, F8 is a test problem with three objectives. F9 and F10 are two more complex test problems than F5-F7. Among them, for F9, sometimes its environmental changes are slight, but sometimes, its PS  will jump from one area to another. For F10, the shapes of PFs in two consecutive environments are inconsistent. As shown in Table \ref{tab10}, FGERS-CPS is far better than CPS on complex problems F9 and F10. But on the three-dimensional problem F8, FGERS-CPS is not as effective as CPS.

\begin{figure}
	\centering	\includegraphics[width=0.8\textwidth]{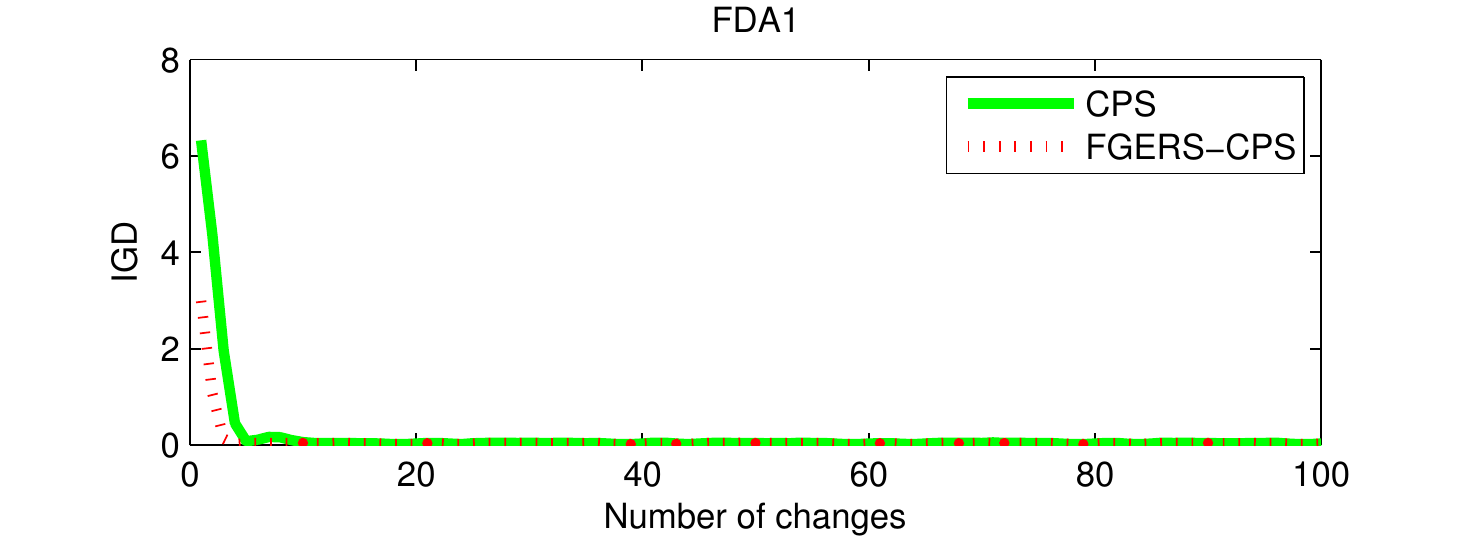}
	\caption{IGD trend comparison of CPS and FGERS-CPS over the number of changes for 20 runs when ($\tau_t$, $n_t$)=(10,10) on FDA1.} \label{fig5}
\end{figure}

\begin{figure}
	\centering	
	\includegraphics[width=0.8\textwidth]{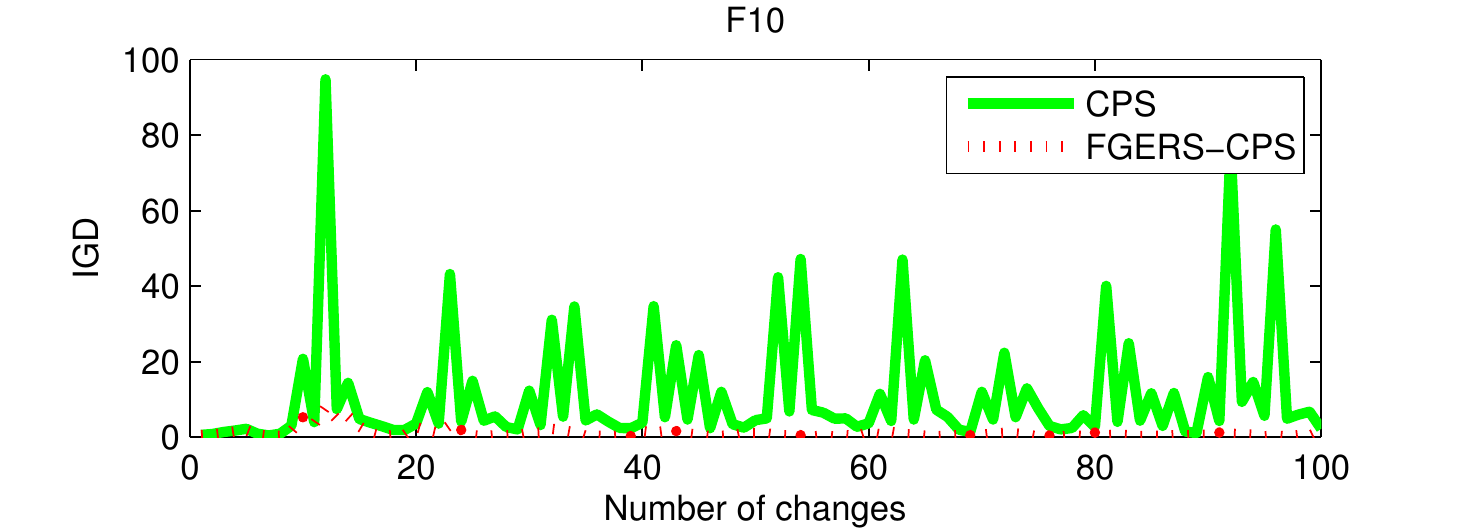}
	\caption{IGD trend comparison of CPS and FGERS-CPS over the number of changes for 20 runs when ($\tau_t$, $n_t$)=(10,10) on F10.} \label{fig6}
\end{figure}

In order to observe the convergence of FGERS-CPS and CPS at every time step, Figs \ref{fig5} and \ref{fig6} show the IGD trend graphs on FDA1 and F10 when ($\tau_t$, $n_t$) = (10, 10), respectively. FDA1 and F10 are the two most representative problems. Since FDA1 is one of the simplest test problems, and F10 is one of the most difficult test problems among used test instances in this paper. As can be seen in Fig. \ref{fig5}, in the beginning few environmental changes, the gap between CPS and FGERS-CPS is more obvious. Nevertheless, in the following some environmental changes, the gap between CPS and FGERS-CPS is not obvious. However, from Table \ref{tab10}, we can know that the effect of FGERS-CPS is better than CPS. As can be seen from Fig. \ref{fig6},  the IGD values of CPS always fluctuate greatly with the environmental changes on F10, while FGERS-CPS is relatively stable. Besides, FGERS-CPS is far lower than CPS. This shows that the feed-forward center point strategy in the novel framework  is greatly improved compared with that in the traditional framework.
\subsection{Comparison of FPS and FPS-GRS}
\label{fps-grs}
To further verify the power of the generational response strategy (GRS), FPS was compared with the combination of FPS and the generational response strategy (FPS-GRS). Here, the used generational response strategy was that in FGERS-CPS. 

From Table \ref{tab11}, it can be seen that when ($\tau_t,n_t$)s are (25,5) and (10,10), FPS-GRS is much better than FPS on all test problems, which means FPS-GRS is better than FPS in the environment with severe environmental change or low change frequency. Moreover, when ($\tau_t,n_t$) is (25,10), FPS-GRS is better than FPS on most test problems except for two 3-dimensional test problems. The reason why FPS is better than FPS-GRS on two 3-dimensional test problems when ($\tau_t,n_t$) is (25,10) will be one of our future research work.

All in all, GRS significantly improves the performance of FPS especially in the complex dynamic environment.
\begin{table*}
	\caption{Mean of MIGD values of FPS and FPS-GRS on 13 benchmark DMOPs on three ($\tau_t$, $n_t$)s which are (25, 5), (25, 10) and (10, 10). The values in boldface denote to have the better effect between these two strategies.}
	\centering
	\label{tab11}
	\scriptsize
	\begin{tabular}{|C{1.5cm}|C{1.0cm}C{1.7cm}|C{1.0cm}C{1.7cm}|C{1.0cm}C{1.7cm}|}
		\hline\noalign{\smallskip}
		Problems	&\multicolumn{2}{c|}{(25,5)} 	&\multicolumn{2}{c|}{(25,10)} &\multicolumn{2}{c|}{(10,10)}\\
			{}      &FPS 	&FPS-GRS	&FPS	&FPS-GRS	&FPS	&FPS-GRS\\
		\hline \noalign{\smallskip}
		FDA1		&0.4864 	&\textbf{0.0625} 	&0.0516 	&\textbf{0.0118} 	&0.0620 		&\textbf{0.0129} \\
		FDA2		&0.0129 	&\textbf{0.0096} 	&0.0085 	&\textbf{0.0075} 	&0.0221 		&\textbf{0.0213} \\
		FDA3		&0.4948 	&\textbf{0.0607} 	&0.0645 	&\textbf{0.0123} 	&0.4301 		&\textbf{0.2598} \\
		FDA4	    &0.2756 	&\textbf{0.2539} 	&\textbf{0.1414} 	&0.1469 	&0.1908 		&\textbf{0.1751} \\
		DMOP1		&0.0390 	&\textbf{0.0187} 	&0.0072 	&\textbf{0.0052} 	&0.0883 		&\textbf{0.0869} \\
		DMOP2		&0.7620 	&\textbf{0.0784} 	&0.0622 	&\textbf{0.0140}	&0.1767 		&\textbf{0.0898} \\
		DMOP3		&0.4589 	&\textbf{0.0662} 	&0.0523 	&\textbf{0.0117} 	&0.0685 		&\textbf{0.0128} \\
		F5		    &1.6632 	&\textbf{1.0333} 	&0.1852 	&\textbf{0.0435} 	&0.6599 		&\textbf{0.2713} \\
		F6		    &0.7360 	&\textbf{0.1430} 	&0.0548 	&\textbf{0.0245} 	&0.6275 		&\textbf{0.1852} \\
		F7		    &0.5555 	&\textbf{0.2659} 	&0.1273 	&\textbf{0.0247} 	&0.3752 		&\textbf{0.2019} \\
		F8	        &0.3079 	&\textbf{0.2723} 	&\textbf{0.1418}	&0.1486 	&0.1872 		&\textbf{0.1763} \\
		F9	 	    &1.4221 	&\textbf{0.8319} 	&0.3542 	&\textbf{0.0939} 	&0.7221 		&\textbf{0.3653} \\
		F10		    &1.1121 	&\textbf{0.6774} 	&0.4280 	&\textbf{0.0708} 	&0.8003 		&\textbf{0.2643} \\
		
		\noalign{\smallskip}\hline
	\end{tabular}
\end{table*}
\section{Conclusions and future work}
\label{sec:6}
This paper proposes a novel dynamic convergence-accelerated framework based on environmental and  generational response strategies. FGERS is improved based on the traditional dynamic multi-objective framework. A major unit of the traditional dynamic multi-objective framework is the environmental response mechanism. However, for the complex environments (such as the environment with  a low frequency of change or large severity of change), only the environmental response mechanism is not always satisfactory. A generational response mechanism  was added into the novel framework together with the environmental response mechanism. The generational response mechanism is to predict the convergence trend of the population to get the predicted population in the environmental static stage, and then use the predicted population and the current population for environmental selection. Finally, the selected population is used for the evolution of the next generation. In this paper, for convenience,  the feed-forward center point method was selected as the generational response strategy. At the same time, the feed-forward center point method was also selected as the prediction part in the environmental response strategy. By comparing FGERS-CPS with four classical response strategies in the traditional dynamic framework on 13 benchmark DMOPs with different characteristics, experimental results denote FGERS-CPS is  competitive for DMOPs, especially for complex DMOPs. 

Besides, to verify the effectiveness of FGERS, FGERS-CPS was compared with CPS. To be fair, the memory strategy and adaptive diversity maintenance strategy were removed from FGERS-CPS. Only the feed-forward center center strategy is kept in FGERS-CPS.
In the traditional dynamic framework, CPS is used as the environmental response mechanism.  FGERS-CPS not only uses the feed-forward center point method in the environmental response mechanism (this is the same as CPS) but also uses that in the generational response mechanism to predict the evolution trend of the population. By comparing FGERS-CPS with CPS on 13 DMOPs with different characteristics, different environmental change frequency levels, and environmental change severity levels, we found that FGERS-CPS is better than CPS on all two-objective test problems. Especially in complex problems, FGERS-CPS has a greater advantage. Nevertheless, in some cases, FGERS-CPS did not show its advantages on the three-objective test problems FDA4 and F8. However, when the severity of the environmental change is larger, FGERS-CPS will perform better, even better than CPS on three-objective test problems. How to achieve better results on DMOPs with three objectives or more objectives is one of our future research work.

In the end, to further verify the power of the generational response strategy, GRS was added to FPS to be FPS-GRS compared with FPS. The used GRS in GRS-FPS is that in FGERS-CPS, which is the feed-forward center point strategy. FPS-GRS was compared with FPS on 13 test problems with different characteristics in different environments (e.g., different change frequency and change severity). Experimental results show that FPS-GRS is much better than FPS especially in the complex environment (e.g. the environment with severe change or low change frequency), which denotes that GRS can significantly improve the performance of FPS.

For convenience, the step size of the generational response mechanism was set to 1. In fact, a suitable step size can help the generational response mechanism find the optimal solutions faster and greatly reduce the consumption of computing resources. Therefore, how to adjust the step size of the generational response strategy is also one of our future work.

In addition, it is worth noting that DMOPs we used is with deterministic changes. It is a kind of simple test problem. However, there are some more complex test problems, such as the problems with noise or less predictable environmental changes. How to detect and solve this kind of DMOPs is also one of our future research work.

Here, we just propose a novel framework. The feed-forward center point strategy used in the novel framework is only a simple example. We can try to use more complex methods to FGERS, such as methods in machine learning or deep learning. Therefore, how to design the better generational response mechanism and environmental response mechanism, is also the work we will study in the future.\\

\section*{Acknowledgements}
The authors wish to thank the support of Natural Science Foundation of Shandong province (ZR2018MF003), Key Research and Development Project of Shandong Province (2019JZZY010132, 2019-0101), National Natural Science Foundation of China (61601267), National Key Research and Development Project(2018YFE0119700), International S\&T Cooperation Program of Shandong Academy of Sciences (2019GHPY18).
\section*{References}

\bibliography{mybibfile}

\end{document}